\documentclass[cup6a]{cupbook}

\usepackage{amssymb,amsmath,subfigure,epsfig}

\title[67577 - Fall, 2008]{Introduction to Machine Learning}

\author{Amnon Shashua\\
School of Computer Science and Engineering\\
The Hebrew University of Jerusalem\\
Jerusalem, Israel}

\date{22nd April 2009}

\begin{document}

\pagenumbering{roman}
\maketitle
\tableofcontents
\cleardoublepage
\pagenumbering{arabic}

\newcommand{\figs}{./figs/}
\newcommand{\figsheight}{2.5cm}
\newcommand{\figssep}{0.4cm}
\newcommand{\figssepp}{0.15cm}
\newcommand{\figswhere}[1]{{\it #1:\ }}
\newcommand{\myfigs}{./pcafigs}
\def\eop {{\noindent\framebox[0.5em]{\rule[0.25ex]{0em}{0.75ex}}}}
\def\be {\begin{equation}}
\def\ee {\end{equation}} 
\def\bea {\begin{eqnarray}}
\def\eea {\end{eqnarray}} 
\def\beas {\begin{eqnarray*}}
\def\eeas {\end{eqnarray*}} 
\newtheorem{claim}{Claim}
\newtheorem{theorem}{Theorem}
\newtheorem{lemma}{Lemma}
\newtheorem{definition}{Definition}

\newcommand{\cond}{\ |\ }
\newcommand{\argmin}[1]{\underset{#1}{\mathrm{argmin}} \:}
\newcommand{\argmax}[1]{\underset{#1}{\mathrm{argmax}} \:}
\newcommand{\argminn}{\mbox{argmin}}
\newcommand{\argmaxx}{\mbox{argmax}}
\newcommand{\mbf}[1]{\mbox{\boldmath $#1$}}
\newcommand{\rank}{\mbox{rank}}
\newcommand{\diag}{\mbox{diag}}
\newcommand{\bfu}{\mbox{\bf u}}
\newcommand{\bfx}{\mbox{\bf x}}
\newcommand{\bfy}{\mbox{\bf y}}
\newcommand{\bfsig}{{\bf \Sigma}}
\newcommand{\bfw}{\mbox{\bf w}}
\newcommand{\bfv}{\mbox{\bf v}}
\newcommand{\bfg}{\mbox{\bf g}}
\newcommand{\bfh}{\mbox{\bf h}}
\newcommand{\bfa}{\mbox{\bf a}}
\newcommand{\bfz}{\mbox{\bf z}}
\newcommand{\bfq}{\mbox{\bf q}}
\newcommand{\bfp}{\mbox{\bf p}}
\newcommand{\bfc}{\mbox{\bf c}}
\newcommand{\bff}{\mbox{\bf f}}
\newcommand{\bfone}{\mbox{\bf 1}}
\newcommand{\bfsigma}{\boldsymbol{\sigma}}
\newcommand{\bflambda}{\boldsymbol{\lambda}}
\newcommand{\bfeps}{\boldsymbol{\epsilon}}
\newcommand{\bfmu}{\boldsymbol{\mu}}
\newcommand{\bfdelta}{\boldsymbol{\delta}}

\chapter{Bayesian Decision Theory}

During the next few lectures we will be looking at the inference from training data problem as a {\it random\/} process modeled by the joint probability distribution over input (measurements) and output (say class labels) variables.  In general, estimating the underlying distribution is a daunting and unwieldy task, but there are a number of constraints or "tricks of the trade" so to speak that under certain conditions make this task manageable and fairly effective.

To make things simple, we will assume a discrete world, i.e., that the values of our random variables take on a finite number of values. Consider for example two random variables $X$ taking on $k$ possible values $x_1,...,x_k$ and  $H$ taking on two values $h_1,h_2$. The values of $X$ could stand for a Body Mass Index (BMI) measurement $weight/height^2$ of a person and $H$ stands for the two possibilities $h_1$ standing for the "person being over-weight" and $h_2$ as the possibility "person of normal weight". Given a BMI measurement we would like to estimate the probability of the person being over-weight.

The joint probability $P(X,H)$ is a two dimensional array (2-way array) with $2k$ entries (cells). Each training example $(x_i,h_j)$ falls into one of those cells, therefore $P(X=x_i,H=h_j)=P(x_i,h_j)$  holds the ratio between the number of hits into cell $(i,j)$ and the total number of training examples (assuming the training data arrive i.i.d.). As a result $\sum_{ij} P(x_i,h_j) = 1$. 

The projections of the array onto its vertical and horizontal axes by summing over columns or over rows is called {\it marginalization\/} and produces $P(h_j) = \sum_i P(x_i,h_j)$ the sum over the j'th row is the probability $P(H=h_j)$, i.e., the probability of a person being over-weight (or not) before we see any measurement --- these are called {\it priors\/}. Likewise, $P(x_i)=\sum_j  P(x_i,h_j)$ is the probability $P(X=x_i)$ which is the probability of receiving such a BMI measurement to begin with --- this is often called {\it evidence\/}. Note that, by definition, $\sum_j P(h_j) = \sum_i P(x_i) =1$. In Fig.~\ref{fig:table} we have that $P(h_1)=14/22, P(h_2)=8/22$ that is there is a higher prior probability of a person being over-weight than being of normal weight. Also $P(x_3)=7/22$ is the highest meaning that we encounter ${\rm BMI}=x_3$ with the highest probability.

The {\it conditional probability\/} $P(h_j\ |\ x_i) = P(x_i,h_j)/P(x_i)$ is the ratio between the number of hits in cell $(i,j)$ and the number of hits in the i'th column, i.e., the probability that the outcome is $H=h_j$ given the measurement $X=x_i$. In Fig.~\ref{fig:table} we have $P(h_2\  |\ x_3)=3/7$. Note that 
$$\sum_j P(h_j\ |\ x_i)=\sum_j \frac{P(x_i,h_j)}{P(x_i)} = \frac{1}{P(x_i)}\sum_j P(x_i,h_j) = P(x_i)/P(x_i)=1.$$ 
Likewise, the conditional probability $P(x_i\ |\ h_j)=P(x_i,h_j)/P(h_j)$ is the number of hits in cell $(i,j)$ normalized by the number of hits in the j'th row and represents the probability of receiving ${\rm BMI}=x_i$ given the class label $H=h_j$ (over-weight or not) of the person. In Fig.~\ref{fig:table} we have $P(x_3\ |\ h_2)=3/8$ which is the probability of receiving ${\rm BMI}=x_3$ given that the person is known to be of normal weight. Note that $\sum_i P(x_i\ |\ h_j)=1$.
\begin{figure}
\begin{center}\begin{tabular}{c|c|c|c|c|cl}
\hline
$h_1$ & 2 & 5 & 4 & 2 & 1\\
\hline
$h_2$ & 0 & 0 & 3 & 3 & 2 \\
\hline
& $x_1$ & $x_2$ &$x_3$ &$x_4$ & $x_5$\\
\hline\end{tabular}\end{center}
\caption{\small Joint probability $P(X,H)$ where $X$ ranges over 5 discrete values and $H$ over two values. Each entry contains the number of hits for the cell $(x_i,h_j)$. The joint probability $P(x_i,h_j)$ is the number of hits divided by the total number of hits (22). See text for more details.}
\label{fig:table}
\end{figure}

The Bayes formula arises from:
$$P(x_i\ |\ h_j)P(h_j) = P(x_i,h_j) = P(h_j\ |\ x_i)P(x_i),$$
from which we get:
$$P(h_j\ |\ x_i) = \frac{P(x_i\ | \ h_j)P(h_j)}{P(x_i)}.$$
The left hand side $P(h_j\ |\ x_i)$ is called the {\it posterior\/} probability and $P(x_i\ | \ h_j)$ is called the {\it class conditional likelihood\/}. The Bayes formula provides a way to estimate the posterior probability from the prior, evidence and class likelihood. It is useful in cases where it is natural to compute (or collect data of) the class likelihood, yet it is not quite simple to compute directly the posterior. For example, given a measurement "12" we would like to estimate the probability that the measurement came from tossing a pair of dice or from spinning a roulette table. If $x=12$ is our measurement, and $h_1$ stands for "pair of dice" and $h_2$ for "roulette" then it is natural to compute the class conditional: $P("12"\ |\ "pair\  of\  dice")=1/36$ and $ P("12"\ |\ "roulette")=1/38$. Computing the posterior directly is much more difficult. As another example, consider medical diagnosis. Once it is known that a patient suffers from some disease $h_j$, it is natural to evaluate the probabilities $P(x_i\ |\ h_j)$ of the emerging symptoms $x_i$. As a result, in many inference problems it is natural to use the class conditionals as the basic building blocks and use the Bayes formula to invert those to obtain the posteriors.

The Bayes rule can often lead to unintuitive results --- the one in particular is known as "base rate fallacy" which shows how an nonuniform prior can influence the mapping from likelihoods to posteriors. On an intuitive basis, people tend to ignore priors and equate likelihoods to posteriors. The following example is typical:
 consider the "Cancer test kit" problem\footnote{This example is adopted from Yishai Mansour's class notes on Machine Learning.} which has the following features: given that the subject has Cancer "C", the probability of the test kit producing a positive decision "+" is $P(+\cond C)=0.98$  (which means that $P(-\cond C)=0.02$) and the probability of the kit producing a negative decision "-" given that the subject is healthy "H" is $P(-\cond H)=0.97$ (which means also that $P(+\cond H)=0.03$). The prior probability of Cancer in the population is $P(C)=0.01$. These numbers appear at first glance as quite reasonable, i.e, there is a probability of $98\%$ that the test kit will produce the correct indication given that the subject has Cancer. What we are actually interested in is the probability that the subject has Cancer given that the test kit generated a positive decision, i.e., $P(C\cond +)$. Using Bayes rule:
$$P(C\cond +) = \frac{P(+\cond C)P(C)}{P(+)}=\frac {P(+\cond C)P(C)}{P(+\cond C)P(C) +P(+\cond H)P(H) } = 0.266$$ 
which means that there is a $26.6\%$ chance that the subject has Cancer given that the test kit produced a positive response --- by all means a very poor performance. 

If we draw the posteriors $P(h_1\ | x)$ and $P(h_2\ |\ x)$ using the probability distribution array in Fig.~\ref{fig:table} we will see that $P( h_1\ | x) > P(h_2\ |\ x)$ for all values of $X$ smaller than a value which is in between $x_3$ and $x_4$. Therefore the decision which will minimize the probability of misclassification  would be to choose the class with the maximal posterior:
$$h^* = \argmax{j} P(h_j\ |\ x),$$
which is known as the Maximal A Posteriori (MAP) decision principle. Since $P(x)$ is simply a normalization factor, the MAP principle is equivalent to:
$$h^* = \argmax{j} P(x\ |\ h_j)P(h_j).$$
In the case where information about the prior $P(h)$ is not known or it is known that the prior is uniform, the we obtain the Maximum Likelihood (ML) principle:
$$h^* = \argmax{j} P(x\ |\ h_j).$$

The MAP principle is a particular case of a more general principle, known as "proper Bayes", where a {\it loss\/} is incorporated into the decision process. Let $l(h_i,h_j)$ be the loss incurred by deciding on class $h_i$ when in fact $h_j$ is the correct class. For example, the "0/1" loss function is:
$$l(h_i,h_j) = \left\{\begin{array}{cc} 1 & i\not = j\\ 0 & i=j\end{array}\right\}$$
The least-squares loss function is: $l(h_i,h_j)=\|h_i - h_j\|^2$ typically used when the outcomes are vectors in some high dimensional space rather than class labels.  We define the {\it expected risk\/}:
$$R(h_i\ |\ x) = \sum_j l(h_i,h_j)P(h_j\ |\ x).$$
The proper Bayes decision policy is to minimize the expected risk:
$$h^* =\argmin{j} R(h_j\ |\ x) .$$
The MAP policy arises in the case $l(h_i,h_j)$ is the 0/1 loss function:
$$R(h_i\ |\ x) = \sum_{j\not= i} P(h_j\ |\ x) = 1 - P(h_i\ |\ x),$$
Thus,
$$\argmin{j} R(h_j\ |\ x) = \argmax{j} P(h_j\ |\ x).$$

\section{Independence Constraints}

At this point we may pause and ask what have we obtained? well, not much. Clearly, the inference problem is captured by the joint probability distribution and we do not need all these formulas to see this. How do we obtain the necessary data to fill in the probability distribution array to begin with? Clearly without additional simplifying constraints the task is not practical as the size of these kind of arrays are exponential in the number of variables.
There are three families of simplifying constraints used in the literature:
\begin{itemize}
\item statistical independence constraints,
\item parametric form of the class likelihood $P(x_i\ |\ h_j)$ where the inference becomes a density estimation problem,
\item structural assumptions --- latent (hidden) variables, graphical models.
\end{itemize}
Today we will focus on the first of these simplifying constraints --- statistical independence properties.

Consider two random variables $X$ and $Y$. The variables are statistically independent $X\bot Y$ if $P(X\ |\ Y)=P(X)$ meaning that information about the value of $Y$ does not add anything about $X$. The independence condition is equivalent to the constraint: $P(X,Y)=P(X)P(Y)$. This can be easily proven: if $X\bot Y$ then $P(X,Y)=P(X\ |\ Y)P(Y)=P(X)P(Y)$. On the other hand, if $P(X,Y)=P(X)P(Y)$ then
$$P(X\ |\ Y) = \frac{P(X,Y)}{P(Y)} = \frac{P(X)P(Y)}{P(Y)} = P(X).$$
Let the values of $X$ range over $x_1,...,x_k$ and the values of $Y$ range over $y_1,...,y_l$. The associated $k\times l$ 2-way array, $P(X=x_i,Y=y_j)$ is represented by the outer product $P(x_i,y_j)=P(x_i)P(y_j)$ of two vectors $P(X)=(P(x_1),...,P(x_k))$ and $P(Y)=(P(y_1),...,P(y_l))$. In  other words, the 2-way array viewed as a {\it matrix\/} is of rank 1 and is determined by $k+l$ (minus 2 because the sum of each vector is 1) parameters rather than $kl$ (minus 1) parameters. 

Likewise, if $X_1\bot X_2\bot....\bot X_n$ are $n$ statistically independent random variables where $X_i$ ranges over $k_i$ discrete and distinct values, then the n-way array $P(X_1,...,X_n)=P(X_1)\cdot...\cdot P(X_n)$ is an outer-product of $n$ vectors and is therefore determined by $k_1+...+k_n$ (minus n) parameters instead of $k_1k_2...k_n$ (minus 1) parameters\footnote{I am a bit over simplifying things because we are ignoring here the fact that the entries of the array should be non-negative. This means that there are additional non-linear constraints which effectively reduce the number of parameters --- but nevertheless it stays exponential.}. Viewed as a tensor, the joint probability is a rank 1 tensor. The main point is that the statistical independence assumption reduced the representation of the multivariate joint distribution from {\it exponential to linear size}.

Since our variables are typically divided to measurement variables and an output/class variable $H$ (or in general $H_1,...,H_l$), it is useful to introduce another, weaker form, of independence known as {\it conditional independence\/}. Variables $X,Y$ are conditionally independent given $H$, denoted by $X\bot Y\ |\ H$, iff $P(X\ |\ Y,H)=P(X\ |\ H)$ meaning that given $H$, the value of $Y$ does not add any information about $X$. This is equivalent to the condition
$P(X,Y\ |\ H)= P(X\  |\ H)P(Y\ |\ H)$. The proof goes as follows:
\begin{itemize}
\item If $P(X\ |\ Y,H)=P(X\ |\ H)$, then

\beas
P(X,Y\cond H)&=&\frac{P(X,Y,H)}{P(H)}=\frac{P(X\cond Y,H)P(Y,H)}{P(H)}\\
&=&\frac{P(X\cond Y,H)P(Y\cond H)P(H)}{P(H)}
=P(X\  |\ H)P(Y\ |\ H)
\eeas

\item If $P(X,Y\ |\ H)= P(X\  |\ H)P(Y\ |\ H)$, then
$$P(X\ |\ Y,H)=\frac{P(X,Y,H)}{P(Y,H)}=\frac{P(X,Y\cond H)}{P(Y\cond H)}=P(X\cond H).$$
\end{itemize}

Consider as an example, Joe and Mo live on opposite sides of the city. Joe goes to work by train and Mo by car. Let $X$ be the event "Joe is late to work" and $Y$ be the event "Mo is late for work". Clearly $X$ and $Y$ are not independent because there could be other factors. For example, a train strike will cause Joe to be late, but because of the strike there would be extra traffic (people using their car instead of the train) thus causing Mo to be pate as well. Therefore, a third variable $H$ standing for the event "train strike" would decouple $X$ and $Y$.

From a computational standpoint, the conditional independence assumption has a similar effect to the unconditional independence. Let $X$ range over $k$ distinct value, $Y$ range over $r$ distinct values and $H$ range over $s$ distinct values. Then $P(X,Y,H)$ is a 3-way array of size $k\times r\times s$. Given that $X\bot Y\ |\ H$ means that $P(X,Y\ |\ H=h_i)$, a 2-way "slice" of the 3-way array along the H axis is represented by the outer-product of two vectors $P(X\ |\ H=h_i)P(Y\ |\ H=h_i)$. As a result the 3-way array is represented by $s(k+r-2)$ parameters instead of $skr-1$. Likewise, if $X_1\bot....\bot X_n\ |\ H$ then the n-way array $P(X_1,...,X_n\ |\ H=h_i)$ (which is a slice along the H axis of the $(n+1)$-array $P(X_1,...,X_n,H)$) is represented by an outer-product of $n$ vectors, i.e., by $k_1+..+k_n - n$ parameters. 

\subsection{Example: Coin Toss}

We will use the ML principle to estimate the bias of a coin. Let $X$ be a random variable taking the value $\{0,1\}$ and $H$ would be our hypothesis taking a real value in $[0,1]$ standing for the coin's bias. If the coin's bias is $q$ then $P(X=0\cond H=q)=q$ and $P(X=1\cond H=q)=1-q$. We receive $m$ i.i.d. examples $x_1,...,x_m$ where $x_i\in\{0,1\}$. We wish to determine the value of $q$. Given that $x_1\bot...\bot x_m\cond H$, the ML problem we must solve is:
$$q^*=\argmax{q} P(x_1,...,x_m\cond H=q) = \prod_{i=1}^m P(x_i\cond q)=\argmax{q} \sum_i \log P(x_i\cond q).$$
Let $0\le \lambda\le m$ stand for the number of '0' instances, i.e., $\lambda=|\{x_i=0\cond i=1,...,m\}|$.
Therefore our ML problem becomes:
$$q^*=\argmax{q} \left\{ \lambda \log q + (n-\lambda)\log (1-q)\right\}$$
Taking the partial derivative with respect to $q$ and setting it to zero:
$$\frac{\partial}{\partial q}[\lambda \log q + (n-\lambda)\log (1-q)] = \frac{\lambda}{q^*} - \frac{n-\lambda}{1-q^*} = 0,$$
produces the result: 
$$q^*=\frac{\lambda}{n}.$$

\subsection{Example: Gaussian  Density Estimation}

So far we considered constraints induced by conditional independent statements among the random variables as a means to reduce the space and time complexity of the multivariate distribution array. Another approach would be to assume some form of parametric form governing the entries of the array --- the most popular assumption is Gaussian distribution $P(X_1,...,X_n)\sim N(\mu, E)$ with mean vector $\mu$ and covariance matrix $E$. The parameters of the density function are denoted by $\theta=(\mu,E)$ and for every vector $\bfx\in R^n$ we have:
$$P(\bfx\cond \theta)=\frac{1}{(2\pi)^{n/2}|E|^{1/2}}\exp^{-\frac{1}{2}(\bfx-\mu)^\top E^{-1}(\bfx-\mu)}.$$
Assume we are given an i.i.d sample of $k$ points $S=\{\bfx_1,...,\bfx_k\}$, $\bfx_i\in R^n$, and we would like to find the Bayes optimal $\theta$:
$$\theta^* = \argmax{\theta} P(S\cond \theta),$$
by maximizing the likelihood (here we are assuming that the the priors $P(\theta)$ are equal, thus the maximum likelihood and the MAP would produce the same result). Because the sample was drawn i.i.d. we can assume that:
$$P(S\cond\theta)=\prod_{i=1}^k P(\bfx_i\cond\theta).$$
Let $L(\theta)=\log P(S\cond\theta)=\sum_i \log P(\bfx_i\cond \theta)$ and since Log is monotonously increasing we have that $\theta^*=\argmax{\theta} L(\theta)$. The parameter estimation would be recovered by taking derivatives with respect to $\theta$, i.e., $\nabla_\theta L=0$. We have:
\be
L(\theta)= -\frac{1}{2}\log |E| -\sum_{i=1}^k \frac{n}{2}\log(2\pi)  - \sum_i\frac{1}{2}(\bfx_i-\mu)^\top E^{-1}(\bfx_i-\mu).\label{eq:12-4}
\ee
We will start with a simple scenario where $E=\sigma^2I$, i.e., all the covariances are zero and all the variances are equal to $\sigma^2$. Thus, $E^{-1}=\sigma^{-2}I$ and $|E|=\sigma^{2n}$. After substitution (and removal of items which do not depend on $\theta$) we have:
$$L(\theta)= - nk\log\sigma - \frac{1}{2}\sum_i \frac{\| \bfx_i-\mu\|^2}{\sigma^2}.$$
The partial derivative with respect to $\mu$:
$$\frac{\partial L}{\partial\mu} = \sigma^{-2}\sum_i (\mu - \bfx_i) = 0$$
from which we obtain:
$$\mu = \frac{1}{k}\sum_{i=1}^k \bfx_i.$$
The partial derivative with respect to $\sigma$ is:
$$\frac{\partial L}{\partial\sigma} = \frac{nk}{\sigma} - \sigma^{-3}\sum_i \|\bfx_i - \mu\|^2=0,$$ 
from which we obtain:
$$\sigma^2 = \frac{1}{kn}\sum_{i=1}^k \|\bfx_i - \mu\|^2.$$
Note that the reason for dividing by $n$ is due to the fact that $\sigma_1^2=...=\sigma_n^2=\sigma^2$,  so that:
$$\frac{1}{k}\sum_{i=1}^k \|\bfx_i - \mu\|^2 = \sum_{j=1}^n \sigma_j^2 = n\sigma^2.$$
In the general case, $E$ is a full rank symmetric matrix, then the derivative of eqn.~(\ref{eq:12-4}) with respect to $\mu$ is:
$$\frac{\partial L}{\partial\mu} = E^{-1}\sum_i (\mu - \bfx_i) = 0,$$
and since $E^{-1}$ is full rank we obtain $\mu=(1/k)\sum_i \bfx_i$. For the derivative with respect to $E$ we note two auxiliary items:
$$\frac{\partial |E|}{\partial E} = |E| E^{-1},\ \ \ \ \ \ \ \   \frac{\partial }{\partial E} trace(AE^{-1}) = - (E^{-1}AE^{-1})^\top.$$
Using the fact that $\bfx^\top\bfy=trace(\bfx\bfy^\top)$ we can transform $\bfz^\top E^{-1}\bfz$ to $trace(\bfz\bfz^\top E^{-1})$ for any vector $\bfz$. Given that $E^{-1}$ is symmetric, then:
$$\frac{\partial}{\partial E} trace(\bfz\bfz^\top E^{-1}) = -E^{-1}\bfz\bfz^\top E^{-1}.$$
Substituting $\bfz=\bfx - \mu$ we obtain:
$$\frac{\partial L}{\partial E} = -kE^{-1} + E^{-1}\left(\sum_i (\bfx_i-\mu)(\bfx_i-\mu)^\top\right)E^{-1}=0,$$
from which we obtain:
$$E=\frac{1}{k}\sum_{i=1}^k (\bfx_i-\mu)(\bfx_i-\mu)^\top.$$

\section{Incremental Bayes Classifier}

Consider another application of conditional dependence which is the Bayes incremental rule. Suppose we have processed $n$ examples $X^{(n)}=\{X_1,...,X_n\}$ and computed somehow $P(H\ |\ X^{(n)})$. We are given a new measurement $X$ and wish to compute (update) the posterior $P(H\ |\ X^{(n)},X)$. We will use the chain rule\footnote{this is based on the rule $P(X_1,...,X_n)=P(X_1\ |\ X_2,...,X_n)P(X_2\ |\ X_3,...,X_n)\cdot\cdot\cdot P(X_{n-1}\ |\ X_n)P(X_n)$}:
$$P(X\ |\ Y,Z)=\frac{P(X,Y,Z)}{P(Y,Z})=\frac{P(Z\ |\ X,Y)P(X\ |\ Y)P(Y)}{P(Z\ |\ Y)P(Y)}=
\frac{P(Z\ |\ X,Y)P(X\ |\ Y)}{P(Z\ |\ Y)}$$
to obtain:
$$P(H\ |\ X^{(n)},X)=\frac{P(X\ |\ X^{(n)},H)P(H\ |\ X^{(n)})}{P(X\ |\ X^{(n)})}$$
from conditional independence, $P(X\ |\ X^{(n)},H) = P(X\ |\ H)$. The term $P(X\ |\ X^{(n)})$ can expanded as follows:

\begin{eqnarray*}
P(X\ |\ X^{(n)})&=&\sum_i \frac{P(X,X^{(n)}\ |\ H=h_i)P(H=h_i)}{P(X^{(n)})}\\
&=&\sum_i \frac{P(X\ |\ H=h_i)P(X^{(n)}\ |\ H=h_i)P(H=h_i)}{P(X^{(n)})}\\
&=&\sum_i P(X\ |\ H=h_i)P(H=h_i\ |\ X^{(n)})
\end{eqnarray*}

After substitution we obtain:
$$P(H=h_i\ |\ X^{(n)},X)=\frac{P(X\ |\ H=h_i)P(H=h_i\ |\ X^{(n)})}{\sum_j P(X\ |\ H=h_j)P(H=h_j\ |\ X^{(n)})}.$$
The old posterior $P(H\ |\ X^{(n)})$ is now the prior for the updated formula. Consider the following example\footnote{adopted from Ron Rivest's 1994 class notes.}: We have a coin which could be either fair or biased towards Head at a probability of $0.6$. Let $H=h_1$ be the event that the coin is fair, and $H=h_2$ that the coin is biased. We start with prior probabilities $P(h_1)=0.75$ and $P(h_2)=0.25$ (we have a higher initial belief that the coin is fair). Suppose our first coin toss is a Head, i.e., $X_1="0"$. Then,
$$P(h_1\ |\ x_1)=\frac{P(x_1\ |\ h_1)P(h_1)}{P(x_1)}=\frac{0.5*0.75}{0.5*0.75+0.6*0.25}=0.714$$
and $P(h_2\ |\ x_1)=0.286$. Our posterior belief that the coin is fair has gone down after a Head toss. Assume we have another measurement $X_2="0"$, then:
$$P(h_1\ |\ x_1,x_2)=\frac{P(x_2\ |\ h_1)P(h_1\ |\ x_1)}{normalization}=\frac{0.5*0.714}{0.5*0.714 + 0.6*0.286}=0.675,$$
and $P(h_2\ |\ x_1,x_2)=0.325$, thus our belief that the coin is fair continues to go down after Head tosses.

\section{Bayes Classifier for 2-class Normal Distributions}

For the last topic in this lecture consider  the 2-class inference problem. We will encountered this problem  in this course in the context of  SVM and LDA. In the Bayes framework, if $H=\{h_1,h_2\}$ denotes the "class member" variable with two possible outcomes, then the MAP decision policy calls for making the decision based on data $\bfx$:
$$h^* = \argmax{h_1,h_2} \left\{ P(h_1\cond\bfx), P(h_2\cond\bfx)\right\},$$
or in other words the class $h_1$ would be chosen if $P(h_1\cond\bfx) > P(h_2\cond\bfx)$. The {\it decision surface\/} (as a function of $\bfx$) is therefore  described by:
$$ P(h_1\cond\bfx) - P(h_2\cond\bfx)=0.$$

The questions we ask here is what would the Bayes optimal decision surface be like if we assume that the two classes are normally distributed with different means and the same covariance matrix? What we will see is that under the condition of equal priors $P(h_1)=P(h_2)$ the decision surface is a hyperplane --- and not only that, it is the same hyperplane produced by LDA.
\begin{claim}
If $P(h_1)=P(h_2)$ and $P(\bfx\cond h_1)\sim N(\mu_1,E)$ and $P(\bfx\cond h_1)\sim N(\mu_2,E)$, the the Bayes optimal decision surface is a hyperplane $\bfw^\top(\bfx-\mu)=0$ where $\mu=(\mu_1+\mu_2)/2$ and $\bfw=E^{-1}(\mu_1-\mu_2)$. In other words, the decision surface is described by:
\be \bfx^\top E^{-1}(\mu_1-\mu_2) - \frac{1}{2}(\mu_1+\mu_2)E^{-1}(\mu_1-\mu_2)=0.\label{eq:12-5}
\ee
\end{claim}
{\bf Proof:\ }
The decision surface is described by $P(h_1\cond\bfx) - P(h_2\cond\bfx)=0$ which is equivalent to the statement that the ratio of the posteriors is 1, or equivalently that the log of the ratio is zero, and using Bayes formula we obtain:
$$0=\log \frac{P(\bfx\cond h_1)P(h_1)}{P(\bfx\cond h_2)P(h_2)} = \log \frac{P(\bfx\cond h_1)}{P(\bfx\cond h_2)}.$$
In other words, the decision surface is described by 
$$\log P(\bfx\cond h_1) - \log P(\bfx\cond h_2)= -\frac{1}{2}(\bfx-\mu_1)^\top E^{-1}(\bfx-\mu_1) + \frac{1}{2}(\bfx-\mu_2)^\top E^{-1}(\bfx-\mu_2)=0.$$
After expanding the two terms we obtain eqn.~(\ref{eq:12-5}). \eop

\chapter{Maximum Likelihood/ Maximum Entropy Duality}
\label{chap:3}

In the previous lecture we defined the principle of Maximum Likelihood (ML): suppose we have random variables $X_1,...,X_n$ form  a random sample from a discrete distribution whose joint probability distribution is $P(\bfx\cond \phi)$ where $\bfx=(x_1,...,x_n)$ is a vector in the sample and $\phi$ is a parameter from some parameter space (which could be a discrete set of values --- say class membership). When $P(\bfx\cond \phi)$ is considered as a function of $\phi$ it is called the {\it likelihood function\/}. The ML principle is to select the value of $\phi$ that maximizes the likelihood function over the observations (training set) $\bfx_1,...,\bfx_m$. If the observations are sampled i.i.d. (a common, not always valid, assumption), then the ML principle is to maximize:
$$\phi^* = \argmax{\phi} \prod_{i=1}^m P(\bfx_i\cond\phi) = \argmaxx \log \prod_{i=1}^m P(\bfx_i\cond\phi) = \argmaxx \sum_{i=1}^m \log P(\bfx_i\cond\phi)$$
which due to the product nature of the problem it becomes more convenient to maximize the log likelihood. We will take a closer look today at the ML principle by introducing a key element known as the {\it relative entropy\/} measure between distributions.

\section{ML and Empirical Distribution}

The ML principle states that the empirical distribution of an i.i.d. sequence of examples is the closest possible (in terms of relative entropy which would be defined later) to the true distribution. To make this statement clear let $\cal X$ be a set of symbols $\{ a_1,...,a_n\}$ and let $P(a\cond \theta)$ be the probability (belonging to a parametric family with parameter $\theta$) of drawing a symbol $a\in{\cal X}$.  Let $x_1,...,x_m$ be a sequence of symbols drawn i.i.d. according to $P$.  The {\it occurrence frequency\/} $f(a)$ measures the number of draws of the symbol $a$:
$$ f(a) = | \{i\ : \ x_i = a\}|,$$
and let the {\it empirical distribution\/} be defined by 
$$\hat{P}(a) = \frac{1}{\sum_{\alpha\in{\cal X}}f(\alpha)} f(a) = \frac{1}{\| f\|_1} f(a)= (1/m) f(a).$$ 
The joint probability $P(x_1,...,x_m\cond\phi)$ is equal to the product $\prod_i P(x_i\cond \phi)$ which according to the definitions above is equal to:
$$P(x_1,...,x_m\cond\phi) = \prod_{i=1}^m p(x_i\cond\theta)=\prod_{a \in{\cal X}} P(a\cond\phi)^{f(a)}.$$
The ML principle is therefore equivalent to the optimization problem:
\be
\max_{P\in Q} \prod_{a \in{\cal X}} P(a\cond\phi)^{f(a)}\label{eq:ML1}
\ee
where $Q = \{\bfq\in R^n: \bfq\ge 0,\ \sum_i q_i =1\}$ denote the set of $n$-dimensional probability vectors ("probability simplex"). Let $p_i$ stand for $P(a_i\cond\phi)$ and $f_i$ stand for $f(a_i)$. Since $\argmaxx_x z(x) = \argmaxx_x \ln z(x)$ and given that $\ln \prod_i p_i^{f_i} = \sum_i f_i\ln p_i$ the solution to this problem can be found by setting the partial derivative of the Lagrangian to zero:
$$L(\bfp,\lambda,\mu)= \sum_{i=1}^n f_i\ln p_i - \lambda(\sum_i p_i -1) - \sum_i \mu_i p_i,$$
where $\lambda$ is the Lagrange multiplier associated with the equality constraint $\sum_i p_i -1 =0$ and $\mu_i\ge 0$ are the Lagrange multipliers associated with the inequality constraints $p_i\ge 0$. We also have the complementary slackness condition that sets $\mu_i=0$ if $p_i > 0$.

After setting the partial derivative with respect to $p_i$ to zero we get:
$$p_i = \frac{1}{\lambda+\mu_i} f_i.$$
Assume for now that $f_i>0$ for $i=1,...,n$. Then from complementary slackness we must have $\mu_i=0$ (because $p_i>0$). We are left therefore with the result $p_i=(1/\lambda)f_i$. 
Following the constraint $\sum_i p_1 =1$ we obtain $\lambda = \sum_i f_i$. As a result we obtain: $P(a\cond\phi) = \hat{P}(a)$. 
In case $f_i=0$ we could use the convention $0\ln 0=0$ and from continuity arrive to $p_i=0$.

We have arrived to the following theorem:
\begin{theorem}
\label{thm:1}
The empirical distribution estimate $\hat P$ is the unique Maximum Likelihood estimate of the probability model $Q$ on the occurrence frequency $f()$.
\end{theorem}
This seems like an obvious result but it actually runs deep because the result holds for a very particular (and non-intuitive at first glance) distance measure between non-negative vectors. Let $dist(\bff,\bfp)$ be some distance measure between the two vectors. The result above states that:
\be
\hat{P} = \argmin{\bfp} dist(\bff,\bfp)\ \ s.t.\ \ \bfp\ge 0,\ \sum_i p_i=1,\label{eq:11ml}
\ee
for some (family?) of distance measures $dist()$. It turns out that there is only one\footnote{not exactly ---  the picture is a bit more complex. Csiszar's 1972 measures: $dist(\bfp,\bff)=\sum_i f_i\phi(p_i/f_i)$ will satisfy eqn.~\ref{eq:11ml} provided that $\phi'^{-1}$ is an exponential. However, $dist(\bff,\bfp)$ (parameters positions are switched) will not do it, whereas the relative entropy will satisfy eqn.~\ref{eq:11ml} regardless of the order of the parameters $\bfp,\bff$.}
 such distance measure, known as the relative-entropy,  which satisfies the ML result stated above.

\section{Relative Entropy}

The relative-entropy (RE) measure  $D(\bfx || \bfy)$ between two non-negative vectors $\bfx, \bfy\in R^n$ is defined as:
$$D(\bfx || \bfy) = \sum_{i=1}^n x_i\ln\frac{x_i}{y_i} - \sum_i x_i + \sum_i y_i.$$
In the definition we use the convention that $0\ln\frac{0}{0}=0$ and based on continuity that $0\ln\frac{0}{y}=0$ and $x\ln\frac{x}{0}=\infty$.
When $\bfx,\bfy$ are also probability vectors, i.e., belong to $Q$, then $D(\bfx || \bfy) = \sum_i x_i\ln\frac{x_i}{y_i}$ is also known  as the Kullback-Leibler divergence. 
The RE measure is not a distance metric as it is not symmetric, $D(\bfx || \bfy) \not= D(\bfy || \bfx)$, and does not satisfy the triangle inequality. Nevertheless, it has several interesting properties which make it a fundamental measure in statistical inference.

The relative entropy is always non-negative and is zero if and only if $\bfx=\bfy$. This comes about from the log-sum inequality:
$$\sum_i x_i\ln\frac{x_i}{y_i} \ge (\sum_i x_i)\ln\frac{\sum_i x_i}{\sum_i y_i}$$
Thus,
$$D(\bfx || \bfy) \ge (\sum_i x_i)\ln\frac{\sum_i x_i}{\sum_i y_i} - \sum_i x_i + \sum_i y_i= \bar x \ln\frac{\bar x}{\bar y} - \bar x + \bar y$$
But $a\ln (a/b) \ge a - b$ for $a,b\ge 0$ iff $\ln(a/b) \ge 1 - (b/a)$ which follows from the inequality $\ln (x+1) > x/(x+1)$ (which holds for $x > -1$ and $x\not=0$). We can state the following theorem:
\begin{theorem}
\label{thm:2}
Let $\bff\ge 0$ be the occurrence frequency on a training sample. $\hat{P}\in Q$ is a ML estimate iff
$$\hat{P} = \argmin{\bfp} D(\bff || \bfp)\ \ s.t.\ \ \bfp\ge 0,\ \sum_i p_i=1.$$
\end{theorem}
{\bf Proof:\ }
$$D(\bff || \bfp) = - \sum_i f_i\ln p_i + \sum_i f_i\ln f_i -\sum_i f_i + 1,$$
and 
$$\argmin{\bfp}D(\bff || \bfp) = \argmax{\bfp} \sum_i f_i\ln p_i = \argmax{\bfp}\ln \prod_i p_i^{f_i}.$$
\eop

\begin{figure}
\begin{center}
\includegraphics[height=7cm]{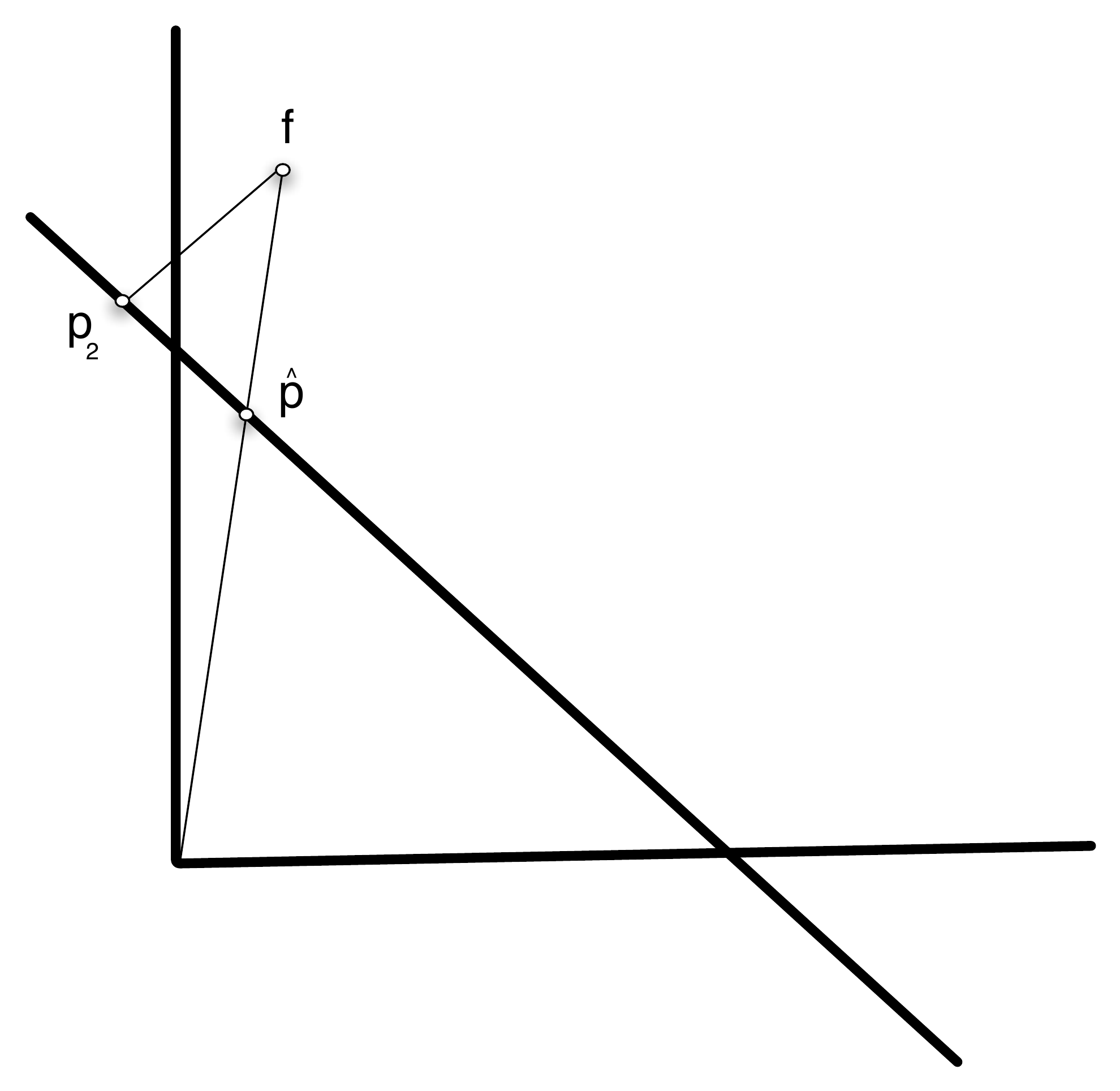} 
\end{center}
\caption{\protect\small Projection of a non-neagtaive vector $\bff$ onto the hyperplane $\sum_i x_i - 1 = 0$. Under relative-entropy the projection $\hat P$ is a scaling of $\bff$ (and thus lives in the probability simplex). Under least-squares the projection $p_2$ lives outside of the probability simplex, i.e., could have negative coordinates.} 
\label{fig:KL-L2}
\end{figure}

There are two (related) interesting points to make here. First, from the proof of Thm.~\ref{thm:1} we observe that the non-negativity constraint $\bfp\ge 0$ need not be enforced - as long as $\bff\ge 0$ (which holds by definition) the closest $\bfp$ to $\bff$ under the constraint $\sum_i p_i =1$ {\it must come out non-negative}. Second, the fact that the closest point $\bfp$ to $\bff$ comes out as a scaling of $\bff$ (which is by definition the empirical distribution $\hat P$) arises because of the relative-entropy measure. For example, if we had used a least-squares distance measure $\| \bff - \bfp\|^2$ the result would not be a scaling of $\bff$. In other words, we are looking for a projection of the vector $\bff$ onto the probability simplex, i.e., the intersection of the hyperplane $\bfx^\top\bfone = 1$ and the non-negative orthant $\bfx\ge 0$. Under relative-entropy the projection is simply a scaling of $\bff$ (and this is why we do not need to enforce non-negativity). Under least-sqaures, a projection onto the hyper-plane $\bfx^\top\bfone = 1$ could take us out of the non-negative orthant (see Fig.~\ref{fig:KL-L2} for illustration). So, relative-entropy is special in that regard --- it not only provides the ML estimate, but also simplifies the optimization process\footnote{The fact that non-negativity "comes for free" does not apply for all class (distribution) models. This point would be refined in the next lecture.} (something which would be more noticeable when we handle a latent class model next lecture).

\section{Maximum Entropy and Duality ML/MaxEnt}

The relative-entropy measure is not symmetric thus we expect different outcomes of the optimization $\min_x D(x || y)$ compared to $\min_y D(x || y)$. The latter of the two, i.e., $\min_{P\in {\cal Q}} D(P_0 || P)$, where $P_0$ is some empirical evidence and $\cal Q$ is some model, provides the ML estimation. For example, in the next lecture we will consider $\cal Q$ the set of low-rank joint distributions (called latent class model) and see how the ML (via relative-entropy minimization) solution can be found. 

Let $H(\bfp) = -\sum_i p_i\ln p_i$ denote the {\it entropy\/} function. With regard to $\min_x D(x || y)$ we can state the following observation:
\begin{claim}
$$\argmin{\bfp\in {\cal Q}} D(\bfp || \frac{1}{n}\bfone) = \argmax{\bfp\in {\cal Q}} H(\bfp).$$
\end{claim}
{\bf Proof: } 
$$D(\bfp || \frac{1}{n}\bfone) = \sum_i p_i\ln p_i + (\sum_i p_i)\ln(n) = \ln(n) - H(\bfp),$$
which follows from the condition $\sum_i p_i = 1$. \eop

In other words, the closest distribution to uniform is achieved by maximizing the entropy. To make this interesting we need to add constraints. Consider a linear constraint on $\bfp$ such as $\sum_i \alpha_i p_i = \beta$. To be concrete, consider a die with six faces thrown many times and we wish to estimate the probabilities $p_1,...,p_6$ given only the {\it average\/} $\sum_i ip_i$. Say, the average is $3.5$ which is what one would expect from an unbiased die. The {\it Laplace's principle of insufficient reasoning\/}  calls for assuming uniformity unless there is additional information (a controversial assumption in some cases). In other words, if we have no information except that each $p_i\ge 0$ and that $\sum_i p_i=1$ we should choose the uniform distribution since we have no reason to choose any other distribution. Thus, employing Laplace's principle we would say that if the average is $3.5$ then the most "likely" distribution is the uniform. What if $\beta = 4.2$? This kind of problem can be stated as an optimization problem:
$$\max_{\bfp} H(\bfp)\ \ s.t., \sum_i p_i=1,\ \sum_i \alpha_i p_i = \beta,$$
where $\alpha_i = i$ and $\beta=4.2$.  We have now two constraints and with the aid of Lagrange multipliers  we can arrive to the result:
$$ p_i = \exp ^{-(1- \lambda)}\exp^{\mu \alpha_i}.$$
Note that because of the exponential $p_i\ge 0$ and again "non-negativity comes for free"\footnote{
Any measure of the class $dist(\bfp,\bfp_0)=\sum_i {p_0}_i\phi(p_i/{p_0}_i)$ minimized under linear constraints will satisfy the result of $p_i\ge 0$ provided that $\phi'^{-1}$ is an exponential.}.
Following the constraint $\sum_i p_i =1$ we get $ \exp ^{-(1- \lambda)} = 1/\sum_i \exp^{\mu \alpha_i}$ from which obtain:
$$ p_i = \frac{1}{Z} \exp^{\mu \alpha_i},$$
where $Z$ (a function of $\mu$) is a normalization factor and $\mu$ needs to be set by using $\beta$ (see later). There is nothing special about the uniform distribution, thus we could be seeking a probability vector $\bfp$ as close as possible to some prior probability ${\bfp}_0$ under the constraints above:
$$\min_{\bfp} D(\bfp || {\bfp}_0)\ \ s.t., \sum_i p_i=1,\ \sum_i \alpha_i p_i = \beta,$$
with the result:
$$p_i = \frac{1}{Z} {p_0}_i\exp^{\mu \alpha_i}.$$
We could also consider adding more linear constraints on $\bfp$ of the form: $\sum_i f_{ij} p_i = b_j$, $j=1,...,k$. The result would be:
$$p_i = \frac{1}{Z} {p_0}_i\exp^{\sum_{j=1}^k \mu_j f_{ij}}.$$
Probability distributions of this form are called {\it Gibbs Distributions}. In practical applications the linear constraints on $\bfp$ could arise from {\it average\/} information about the system such as temperature of a fluid (where $p_i$ are the probabilities of the particles moving at various velocities), rainfall data or general environmental data (where $p_i$ represent the probability of finding animal colonies at discrete locations in a 3D map). A constraint of the form $\sum_i f_{ij} p_i = b_j$ states that the expectation $E_p[f_j]$ should be equal to the empirical distribution $\beta = E_{\hat P}[f_j]$ where $\hat P$ is either uniform or given as input. Let 
$${\cal P} = \{\bfp\in R^n\ :\ \bfp\ge 0,\ \sum_ip_i=1,\ E_p[f_j]=E_{\hat p}[f_j], j=1,...,k\},$$
and
$${\cal Q} = \{\bfq\in R^n\ ;\ \bfq\  {\rm is\  a\  Gibbs\  distribution}\}$$
We could therefore consider looking for the ML solution for the parameters $\mu_1,...,\mu_k$ of the Gibbs distribution:
$$\min_{\bfq\in {\cal Q}} D(\hat{\bfp} || \bfq),$$
where if $\hat{\bfp}$ is uniform then $\min D(\hat{\bfp} || \bfq)$ can be replaced by $\max \sum_i\ln q_i$ (because $D((1/n)\bfone || \bfx) = -\ln(n) - \sum_i \ln x_i$). 

As it turns out, the MaxEnt and ML are duals of each other and the intersection of the two sets ${\cal P} \cap {\cal Q}$ contains only a single point which {\it solves both problems}.
\begin{theorem}
The following are equivalent:
\begin{itemize}
\item MaxEnt: ${\bfq}^* = \argminn_{\bfp\in{\cal P}} D(\bfp || {\bfp}_0)$
\item ML: ${\bfq}^* = \argminn_{\bfq\in {\cal Q}} D(\hat{\bfp} || \bfq)$
\item ${\bfq}^* \in {\cal P} \cap {\cal Q}$
\end{itemize}
\end{theorem}
In practice, the duality theorem is used to recover the parameters of the Gibbs distribution using the ML route (second line in the theorem above) --- the algorithm for doing so is known as the  {\it iterative scaling algorithm\/} (which we will not get into).

\chapter{EM Algorithm: ML over Mixture of Distributions}
\label{chap:45}

 In Lecture~\ref{chap:3} we saw that the Maximum Likelihood (ML) principle over i.i.d. data is achieved  by minimizing the relative entropy between a model $\cal Q$ and the occurrence-frequency of the training data. Specifically, let $\bfx_1,..,\bfx_m$ be i.i.d. where each $\bfx_i \in {\cal X}^d$ is a $d$-tupple of symbols taken from an alphabet $\cal X$ having $n$ different letters $\{a_1,...,a_n\}$. Let $\hat P$ be the empirical joint distribution, i.e., an array with $d$ dimensions where each axis has $n$ entries, i.e., 
each entry ${\hat P}_{i_1,...,i_d}$, where $i_j=1,...,n$,  represents the (normalized) co-occurrence of the $d$-tupe $a_{i_1},...,a_{i_d}$ in the training set $\bfx_1,...,\bfx_m$. We wish to find a joint distribution $P^*$ (also a $d$-array) which belongs to some model family of distributions $\cal Q$ closest as possible to $\hat P$ in relative-entropy:
$$P^* = \argmin{P\in {\cal Q}} D({\hat P} || P).$$
In this lecture we will focus on a model of distributions $\cal Q$ which represents {\it mixtures\/} of simple distributions $\cal H$--- known as {\it latent class models\/}. A latent class model arises when the joint probability $P(X_1,...,X_d)$ we observe (i.e., from which $\hat P$ is generated by observing samples $\bfx_1,...,\bfx_m$) is in fact a marginal of $P(X_1,...,X_d,Y)$ where $Y$ is a "hidden" (or "latent") random variable which has $k$ different discrete values $\alpha_1,..,\alpha_k$. Then,
$$P(X_1,...,X_d) = \sum_{j=1}^k P(X_1,...,X_d\cond Y=\alpha_j)P(Y=\alpha_j).$$
The idea is that given the value of the hidden variable $H$ the problem of recovering the model $P(X_1,...,X_d\cond Y=\alpha_j)$, which belongs to some family of joint distributions $\cal H$, is a relatively simple problem. To make this idea clearer we consider the following example:
Assume we have two coins. The first coin has a probability of heads ("0") equal to $p$ and the second coin has a probability of heads equal to $q$. At each trial we choose to toss coin 1 with probability $\lambda$ and coin 2 with probability $1-\lambda$. Once a coin has been chosen it is tossed 3 times, producing an observation $\bfx\in\{0,1\}^3$. We are given a set of such observations $D=\{\bfx_1,...,\bfx_m\}$ where each observation $\bfx_i$ is a triplet of coin tosses (the same coin). 
Given $D$, we can construct the empirical distribution $\hat P$ which is a $2\times 2 \times 2$ array defined as: 
$${\hat P}_{i_1,i_2,i_3} = \frac{1}{m}| \{\bfx_i = \{i_1,i_2,i_3\},\ i=1,...,m\}|.$$
Let $y_i\in\{1,2\}$ be a random variable associated with the observation $\bfx_i$ such that $y_i=1$ if $\bfx_i$ was generated by coin 1 and $y_i=2$ if $\bfx_i$ was generated by coin 2. If we knew the values of $y_i$ then our task would be simply to estimate two separate Bernoulli distributions by separating the triplets generated from coin 1 from those generated by coin 2. 
Since $y_i$ is not known, we have the marginal:
\bea
P(\bfx=(x_1,x_2,x_3)) &=& P(\bfx=(x_1,x_2,x_3)\cond y=1)P(y=1)\nonumber\\
& +&  P(\bfx=(x_1,x_2,x_3)\cond y=2)P(y=2)\nonumber\\
 &=& \lambda p^{n_i}(1-p)^{(3-n_i)} + (1-\lambda)q^{n_i}(1-q)^{(3-n_i)},\label{eq:1}
\eea
where $(x_1,x_2,x_3)\in\{0,1\}^3$ is a triplet coin toss and 
$0\le n_i\le 3$ is the number of heads ("0") in the triplet of tosses. In other words, the likelihood $P(\bfx)$ of  triplet of tosses $\bfx=(x_1,x_2,x_3)$ is a linear combination ("mixture") of two Bernoulli distributions. 
Let $\cal H$ stand for Bernoulli distributions:
$${\cal H}=\{ \bfu^{\otimes d}\ :\ \bfu\ge 0,\ \sum_{i=1}^n u_i = 1\}$$
where $\bfu^{\otimes d}$ stands for the outer-product of $\bfu\in R^n$ with itself $d$ times, i.e., an n- way array indexed by $i_1,...,i_d$, where  $i_j\in \{1,...,n\}$,  and whose value there is equal to $u_{i_1}\cdot\cdot\cdot u_{i_d}$.
The model family $\cal Q$ is a mixture of Bernoulli distributions: 
$${\cal Q} = \{ \sum_{j=1}^k \lambda_j P_j\ :\ \bflambda\ge 0,\ \sum_j\lambda_j=1,\ P_j\in{\cal H}\},$$
where specifically for our coin-toss example becomes:
$${\cal Q}=\{\lambda \left(\begin{array}{c} p\\ 1-p \end{array}\right)^{\otimes 3} +  (1-\lambda) \left(\begin{array}{c} q\\ 1-q \end{array}\right)^{\otimes 3}\ :\  \lambda,p,q \in [0,1]\}$$
We see therefore that the eight entries of $P^*\in{\cal Q}$ which minimizes $D(\hat P|| P)$ over the set $\cal Q$ is determined by three parameters $\lambda,p,q$. For the coin-toss example this looks like:
\begin{eqnarray*}
&&\argmin{0 \le \lambda,p,q \le 1} D\left(\hat P\  ||\  \lambda \left(\begin{array}{c} p\\ 1-p \end{array}\right)^{\otimes 3} +  (1-\lambda) \left(\begin{array}{c} q\\ 1-q \end{array}\right)^{\otimes 3} \right)\\
&&=\argmax{0 \le \lambda,p,q \le 1} \sum_{i_1=0}^1 \sum_{i_2=0}^1 \sum_{i_3=0}^1 \hat P_{i_1i_2i_3} \log\left(\lambda p^{n_{i_{123}}}(1-p)^{(3-n_{i_{123}})} + (1-\lambda)q^{n_{i_{123}}}(1-q)^{(3-n_{i_{123}})} \right)
\end{eqnarray*}
where $n_{i_{123}}=i_1+i_2+i_3$. Trying to work out an algorithm for minimizing the unknown parameters $\lambda,p,q$ would be somewhat "unpleasant" (and even more so for other families of distributions $\cal H$) because of the log-over-a-sum present in the optimization function --- if we could somehow turn this into a sum-over-log our task  would be much easier. We would then be able to turn the problem into a succession of problems over  $\cal H$ rather than a single problem over ${\cal Q}=\sum_j\lambda_j{\cal H}$. Another point worth attention is the non-negativity of the output variables --- simply minimizing the relative-entropy measure under the constraints of the class model $\cal Q$ would not guarantee a non-negative solution. As we shall see, breaking down the problem into a successions of problems over $\cal H$ would give us the "non-negativity for free" feature.

The technique for turning the log-over-sum into a sum-over-log as part of finding the ML solution for a mixture model is known as the Expectation-Maximization (EM) algorithm introduced by Dempster, Laird and Rubin in 1977. It is based on two ideas: (i) introduce auxiliary variables, and (ii) use of Jensen's inequality. 

\section{The EM Algorithm: General}

Let  $D=\{\bfx_1,...,\bfx_m\}$ represent the training data where $\bfx_i\in{\cal X}$ is taken from some instance space ${\cal X}$ which we leave unspecified. For now, we leave matters to be as general as possible and specifically we do not make independence assumptions  on the data generation process. 

The ML problem is to find a setting of parameters $\theta$ which maximizes the likelihood $P(\bfx_1,...,\bfx_m\cond \theta)$, namely, we wish to maximize $P(D\cond\theta)$ over parameters $\theta$, which is equivalent to maximizing the log-likelihood:
$$\theta^*=\argmax{\theta} \log P(D\cond\theta) = \log \left(\sum_{\bfy} P(D,\bfy \cond\theta)\right),$$
where $\bfy$ represents the hidden variables. We will denote $L(\theta)=\log P(D\cond\theta)$. Let $q(\bfy\cond D,\theta)$ be some (arbitrary) distribution of the hidden variables $\bfy$ conditioned on the parameters $\theta$ and the input sample $D$, i.e., $\sum_{\bfy} q(\bfy\cond D,\theta)=1$. We define a {\it lower bound\/} on $L(\theta)$ as follows:
\bea
L(\theta)&=&  \log \left(\sum_{\bfy} P(D,\bfy \cond\theta)\right)\\
&=&  \log \left(\sum_{\bfy} q(\bfy\cond D,\theta) \frac{P(D,\bfy \cond\theta)}{q(\bfy\cond D,\theta)}\right)\\
&\ge& \sum_{\bfy} q(\bfy\cond D,\theta) \log \frac{P(D,\bfy \cond\theta)}{q(\bfy\cond D,\theta)}\\
&=& Q(q,\theta).
\eea
The inequality comes from Jensen's inequality $\log \sum_j\alpha_ja_j \ge \sum_j \alpha_j\log a_j$  when $\sum_j\alpha_j=1$. What we have obtained is an "auxiliary" function $Q(q,\theta)$ satisfying
$$L(\theta) \ge Q(q,\theta),$$
for all distributions $q(\bfy\cond D,\theta)$.  
The maximization of $Q(q,\theta)$ proceeds by interleaving the variables $q$ and $\theta$ as we separately ascend on each set of variables.  At the $(t+1)$ iteration we fix the current value of $\theta$ to be $\theta^{(t)}$ of the $t$'th iteration and maximize $Q(q,\theta^{(t)})$ over $q$, and then maximize $Q(q^{(t+1)},\theta)$ over $\theta$:
\bea
q^{(t+1)} &=& \argmax{q} Q(q,\theta^{(t)})\\
\theta^{(t+1)} &=&\argmax{\theta} Q(q^{(t+1)},\theta).
\eea
The strategy of the EM algorithm is to maximize the lower bound $Q(q,\theta)$ with the hope that if we ascend on the lower bound function we will also ascend with respect to $L(\theta)$. The claim below guarantees that an ascend on $Q$ will also generate an ascend on $L$:
\begin{claim}[Jordan-Bishop]
The optimal $q(\bfy\cond D,\theta^{(t)})$ at each step is $P(\bfy\cond D,\theta^{(t)})$.
\end{claim}
{\bf Proof:\ }
We will show that $Q(P(\bfy\cond D,\theta^{(t)}),\theta^{(t)}) = L(\theta^{(t)})$ which proves the claim since $L(\theta) \ge Q(q,\theta)$ for all $q,\theta$, thus the best $q$-distribution we can hope to find is one that makes the lower-bound meet $L(\theta)$ at $\theta=\theta^{(t)}$.
\beas
Q(P(\bfy\cond D,\theta^{(t)}),\theta^{(t)})  &=& \sum_{\bfy} P(\bfy\cond D,\theta^{(t)}) \log \frac{P(D,\bfy \cond\theta^{(t)})}{P(\bfy\cond D,\theta^{(t)})}\\
&=& \sum_{\bfy} P(\bfy\cond D,\theta^{(t)}) \log \frac{P(\bfy\cond D,\theta^{(t)})P(D\cond\theta^{(t)})}{P(\bfy\cond D,\theta^{(t)})}\\
&=& \log P(D\cond\theta^{(t)}) \sum_{\bfy} P(\bfy\cond D,\theta^{(t)})\\
&=& L(\theta^{(t)})
\eeas
\eop

The proof provides also the validity for the approach of ascending along the lower bound $Q(q,\theta)$ because at the point $\theta^{(t)}$ the two functions coincide, i.e., the lower bound function at $\theta=\theta^{(t)}$ is equal to $L(\theta^{(t)})$ therefore if we continue and {\it ascend\/} along $Q(\cdot)$ we are {\it guaranteed\/} to ascend along $L(\theta)$ as well\footnote{this manner of deriving EM was adapted from  Jordan and Bishop's book notes, 2001.} --- therefore, convergence is guaranteed. It can also be shown (but omitted here) that the point of convergence is a stationary point of $L(\theta)$ (was shown originally by C.F. Jeff Wu in 1983 years after EM was introduced in 1977) under fairly general conditions.
The second step of maximizing over $\theta$ then becomes:
\be
\theta^{(t+1)} = \argmax{\theta} \sum_{\bfy} P(\bfy\cond D,\theta^{(t)})\log P(D,\bfy\cond \theta).
\label{eq:em1}
\ee
This defines the EM algorithm. Often the "Expectation" step is described as taking the expectation of:
$$E_{\bfy\sim P(\bfy\cond D,\theta^{(t)})}\left[ \log P(D,\bfy\cond \theta)\right],$$
followed by a Maximization step of finding $\theta$ that maximizes the expectation --- hence the term EM for this algorithm. 

Eqn.~\ref{eq:em1} describes a principle but not an algorithm because in general, without making assumptions on the statistical relationship between the data points and the hidden variable the problem presented in eqn.~\ref{eq:em1} is unwieldy. We will reduce eqn.~\ref{eq:em1} to something more manageable  by making the i.i.d. assumption. This is detailed in the following section.

\section{EM with i.i.d. Data}

The EM optimization presented in eqn.~\ref{eq:em1} can be simplified if we assume the data points (and the hidden variable values) are i.i.d.
$$P(D\cond\theta)=\prod_{i=1}^n P(\bfx_i\cond\theta),\ \ \ \ \ \ \ P(D,\bfy\cond\theta)=\prod_{i=1}^n P(\bfx_i,y_i\cond\theta),$$
and
$$P(\bfy\cond D,\theta)=\prod_{i=1}^n P(y_i\cond \bfx_i,\theta).$$
For any $\alpha(y_i)$ we have:
\beas
\sum_{\bfy} \alpha(y_i) P(\bfy\cond D,\theta) &=& \sum_{y_1}\cdot\cdot\cdot\sum_{y_n}\alpha(y_i) P(y_1\cond \bfx_1,\theta)\cdot\cdot\cdot P(y_n\cond \bfx_n,\theta)\\
&=& \sum_{y_i}\alpha(y_i)P(y_i\cond \bfx_i,\theta)
\eeas
this is because $\sum_{y_j}P(y_j\cond \bfx_j,\theta)=1$. Substituting the simplifications above into eqn.~\ref{eq:em1} we obtain:
\be
\theta^{(t+1)} = \argmax{\theta} \sum_{j=1}^k \sum_{i=1}^m P(y_i=\alpha_j\cond \bfx_i,\theta^{(t)})\log P(\bfx_i,y_i=\alpha_j\cond \theta)\label{eq:12em}
\ee
where $y_i\in\{\alpha_1,...,\alpha_k\}$.

\section{Back to the Coins Example}

We will apply the EM scheme to our running example of mixture of Bernoulli distributions. We wish to compute
\beas
Q(\theta,\theta^{(t)}) &=& \sum_{\bfy} P(\bfy\cond D,\theta^{(t)})\log P(D,\bfy\cond \theta)\\
&=& \sum_{i=1}^n\sum_{j=1}^2 P(y_i=j\cond \bfx_i,\theta^{(t)})\log P(\bfx_i,y_i=j\cond \theta),
\eeas
and then maximize $Q()$ with respect to $p,q,\lambda$.
\beas
Q(\theta,\theta') &=& \sum_{i=1}^n \left[ P(y_i=1\cond \bfx_i,\theta')\log P(\bfx_i\cond y_i=1,\theta)P(y_i=1\cond\theta)\right]\\
&+&  \sum_{i=1}^n \left[ P(y_i=2\cond \bfx_i,\theta')\log P(\bfx_i\cond y_i=2,\theta)P(y_i=2\cond\theta)\right]\\
&=& \sum_i\left[\mu_i \log(\lambda p^{n_i}(1-p)^{(3-n_i)}) + (1-\mu_i) \log((1-\lambda) q^{n_i}(1-q)^{(3-n_i)})\right]
\eeas
where $\theta'$ stands for $\theta^{(t)}$ and $\mu_i = P(y_i=1\cond \bfx_i,\theta')$. The values of $\mu_i$ are known since $\theta'=(\lambda_o,p_o,q_o)$ are given from the previous iteration. The Bayes formula is used to compute $\mu_i$:
\beas 
\mu_i &=& P(y_i=1\cond \bfx_i,\theta') = \frac{P(\bfx_i\cond y_i=1,\theta')P(y_i=1\cond \theta')}{P(\bfx_i\cond\theta')}\\
&=& \frac{\lambda_o p_o^{n_i}(1-p_o)^{(3-n_i)}}{\lambda_o p_o^{n_i}(1-p_o)^{(3-n_i)} + (1-\lambda_o) q_o^{n_i}(1-q_o)^{(3-n_i)}}
\eeas
We wish to compute: $\max_{p,q,\lambda} Q(\theta,\theta')$. The partial derivative with respect to $\lambda$ is:
$$\frac{\partial Q}{\partial \lambda} = \sum_i \mu_i\frac{1}{\lambda} - \sum_i(1-\mu_i)\frac{1}{1-\lambda}=0,$$
from which we obtain the update formula of $\lambda$ given $\mu_i$:
$$\lambda= \frac{1}{k}\sum_{i=1}^n\mu_i.$$
The partial derivative with respect to $p$ is:
$$\frac{\partial Q}{\partial p} = \sum_i \frac{\mu_in_i}{p} - \sum_i \frac{\mu_i(3-n_i)}{1-p}=0,$$
from which we obtain the update formula:
$$p = \frac{1}{\sum_i\mu_i}\sum_i \frac{n_i}{3}\mu_i.$$
Likewise the update rule for $q$ is:
$$q = \frac{1}{\sum_i(1-\mu_i)}\sum_i \frac{n_i}{3}(1-\mu_i).$$
To conclude, we start with some initial "guess" of the values of $p,q,\lambda$, compute the values of $\mu_i$ and update iteratively the values of $p,q,\lambda$ where at the end of each iteration the  new values of $\mu_i$ are computed.

\section{Gaussian Mixture}

The Gaussian mixture model assumes that $P(\bfx)$ where $\bfx\in R^d$ is a linear combination of Gaussian distributions 
$$P(\bfx)=\sum_{j=1}^k P(\bfx\cond y=j)P(y=j)$$
where
$$P(\bfx\cond y=j) =\frac{1}{(2\pi)^{d/2}\sigma_j^d}\exp^{-\frac{\|\bfx-\bfc_j\|^2}{2\sigma_j^2}},$$
is Normally distributed with mean $\bfc_j$ and covariance matrix $\sigma_j^2I$. Let $D=\{\bfx_1,...,\bfx_m\}$ be the i.i.d sample data and we wish to solve for the mean and covariances of the individual Gaussians (the "factors") and the mixing coefficients $\lambda_j=P(y=j)$.   In order to make clear where the parameters are located we will write $P(\bfx\cond \phi_j)$ instead of $P(\bfx\cond y=j)$ where $\phi_j=(\bfc_j,\sigma^2_j)$ are the mean and variance of the $j$'th factor. We denote by $\theta$ the collection of mixing coefficients $\lambda_j$ and $\phi_j$, $j=1,...,k$. Let $w_i^j$ be auxiliary variables per point $x_i$ and per factor $y=j$ standing for:
$$w_i^j = P(y_i=j\cond \bfx_i,\theta).$$
The EM step (eqn.~\ref{eq:12em}) is:
\be
\theta^{(t+1)} = \argmax{\theta=\{\bflambda,\phi\}} \sum_{j=1}^k \sum_{i=1}^m  {w_i^j}^{(t)} \log\left(\lambda_j P(\bfx_i\cond\phi_j)\right)\ \ \ {\rm s.t.\ } \sum_j \lambda_j=1.\label{eq:14-1}
\ee
Note the constraint $\sum_j\lambda_j=1$. The update formula for ${w_i^j}$ is done through the use of Bayes formula:
$${w_i^j}^{(t)} =\frac{P(y_i=j\cond \theta^{(t)}) P(\bfx_i\cond y_i=j,\theta^{(t)})}{P(\bfx_i\cond\theta^{(t)})}=\frac{1}{Z_i} \lambda_j^{(t)} P(\bfx_i\cond\phi^{(t)}),$$
where $Z_i$ is a scaling factor so that $\sum_j w_i^j=1$. 

The update formula for $\lambda_j,\bfc_j,\sigma_j$ follow by taking partial derivatives of  eqn.~(\ref{eq:14-1}) and setting them to zero.  Taking partial derivatives with respect to $\lambda_j,\bfc_j$ and $\sigma_j$ we obtain the update rules:
\beas
\lambda_j &=& \frac{1}{m}\sum_{i=1}^m w_i^j\\
\bfc_j &=& \frac{1}{\sum_iw_i^j}\sum_{i=1}^m w_i^j\bfx_i,\\
\sigma_j^2 &=& \frac{1}{d\sum_iw_i^j}\sum_{i=1}^m w_i^j\|\bfx_i -\bfc_j\|^2.
\eeas
In other words, the observations $\bfx_i$ are weighted by $w_i^j$ before a Gaussian is fitted ($k$ times, one for each factor).

\section{Application Examples}

\subsection{Gaussian Mixture and Clustering}

The Gaussian mixture model is  classically used for clustering applications. In a clustering application one receives a sample of points $\bfx_1,...,\bfx_m$ where each point resides in $R^d$. The task of the learner (in this case "unsupervised" learning) is to group the $m$ points into $k$ sets. Let $y_i\in\{1,...,k\}$ where $i=1,...,m$ stands for the required labeling. The clustering solution is an assignment of values to $y_1,...,y_m$ according to some clustering criteria.

In the Gaussian mixture model points are clustered together if they arise from the same Gaussian distribution. The EM algorithm provides a probabilistic assignment $P(y_i=j\cond x_i)$ which we denoted above as $w_i^j$.

\subsection{Multinomial Mixture and "bag of words" Application}

The multinomial mixture (the coins example we toyed with) is typically used for representing "count" data, such as when representing text documents as high-dimensional vectors. A vector representation of a text document associates a word from a fixed vocabulary to a coordinate entry of the vector. The value of the entry  represents the number of times that particular word appeared in the document. If we ignore the order in which the words appeared and count only their frequency, a set of documents $d_1,...,d_m$ and a set of words $w_1,....,w_n$ could be jointly represented by a co-occurence $n\times m$ matrix $G$ where $G_{ij}$ contains the number of times word $w_i$ appeared in document $d_j$. If we scale $G$ such that $\sum_{ij}G_{ij}=1$ then we have a distribution $P(w,d)$. This kind of representation of a set of documents is called "bag of words".

For purposes of search and filtering it is desired to reveal additional information about words and documents such as to which "topic" a document belongs to or to which topics a word is associated with. This is similar to a clustering task where documents associated with the same topic are to be clustered together. This can be achieved by considering the topics as the value of a latent variable $y$:
$$P(w,d) = \sum_y P(w,d\cond y)P(y) = \sum_y P(w\cond y)P(d\cond y)P(y),$$
where we made the assumption that $w\bot d\cond y$ (i.e., words and documents are conditionally independent given the topic). The conditional independent assumption gives rise to the multinomial mixture model. To be more specific, ley $y\in\{1,...,k\}$ denote the $k$ possible topics and let $\lambda_j=P(y=j)$ (note that $\sum_j \lambda_j=1$), then the latent class model becomes:
$$P(w,d)=\sum_{j=1}^k \lambda_j P(w\cond y=j)P(d\cond y=j).$$
Note that $P(w\cond y=j)$ is a vector which we denote as $\bfu_j\in R^n$ and $P(d\cond y=j)$ is also a vector we denote by $\bfv_j\in R^m$. The term $P(w\cond y=j)P(d\cond y=j)$ stands for the outer-product $\bfu_j\bfv_j^\top$ of the two vectors, i.e., is a rank-1 $n\times m$ matrix. The Maximum-Likelihood estimation problem is therefore to find vectors $\bfu_1,...,\bfu_k$ and $\bfv_1,...,\bfv_k$ and scalars $\lambda_1,...,\lambda_k$ such that the empirical distribution represented by the unit scaled matrix $G$ is as close as possible (in relative-entropy measure) to the low-rank matrix $\sum_j \lambda_j\bfu_j\bfv_j^\top$ subject to the constraints of non-negativity and $\sum_j \lambda_j=1$, $\bfu_j$ and $\bfv_j$ are unit-scaled as well ($\bfone^\top\bfu_j=\bfone^\top\bfv_j=1$).

Let $\bfx_i=(w(i),d(i))$ stand for the $i$'th example $i=1,...,q$ where an example is a pair of word and document where $w(i)\in\{1,...,n\}$ is the index to the word alphabet and $d(i)\in\{1,...,m\}$ is the index to the document. The EM algorithm involves the following optimization step:
\begin{eqnarray*}
\theta^{(t+1)}&=&\argmax{\theta} \sum_{i=1}^q\sum_{j=1}^k P(y_i=j\cond \bfx_i,\theta^{(t)})\log P(\bfx_i,y_i=j\cond \theta)\\
&=&\argmax{\theta} \sum_{i=1}^q\sum_{j=1}^k w_{ij}^{(t)}\log \left[ \lambda_j u_{j,w(i)}v_{j,d(i)}\right]\ \ \ s.t. \ \ \ \bfone^\top\lambda=\bfone^\top\bfu_j=\bfone^\top\bfv_j=1
\end{eqnarray*}
An update rule for $u_{jr}$ (the $r$'th entry of $\bfu_j$) is derived below: the derivative of the Lagrangian is:
\begin{eqnarray*}
&&\frac{\partial}{\partial u_{jr}}\left[ \sum_{i=1}^q w_{ij}^{(t)}\log u_{j,w(i)}-\mu u_{jr}\right]\\
&& = \frac{\partial}{\partial u_{jr}}\left[ N(r)\log u_{jr}\sum_{w(i)=r} w_{ij}^{(t)}-\mu u_{jr}\right]\\
&&=\frac{N(r)\sum_{w(i)=r} w_{ij}^{(t)}}{u_{jr}}-\mu = 0
\end{eqnarray*}
where $N(r)$ stands for the frequency of the word $w_r$ in all the documents $d_1,...,d_m$. Note that $N(r)$ is the result of summing-up the $r$'th row of $G$ and that the vector $N(1),...,N(n)$ is the marginal $P(w)=\sum_d P(w,d)$. Given the constraint $\bfone^\top\bfu_j=1$ we obtain the update rule:
$$u_{jr}\leftarrow \frac{N(r)\sum_{w(i)=r} w_{ij}^{(t)}}{\sum_{s=1}^n N(s) \sum_{w(i)=s} w_{ij}^{(t)}}.$$
Update rules for the remaining unknowns are similarly derived. Once EM has converged, then
$\sum_{w(i)=r} w_{ij}^{*}$ is the probability of the word $w_r$ to belong to the $j$'th topic and $\sum_{d(i)=s} w_{ij}^{*}$ is the probability that the $s$'th document comes from the $j$'th topic.

\chapter{Support Vector Machines and Kernel Functions}
\label{chap:67}

In this lecture we begin the exploration of  the 2-class hyperplane separation problem. We
are given a training set of instances $\bfx_i\in R^n$, $i=1,...,m$,
and class labels $y_i=\pm 1$ (i.e., the training set is made up of
``positive'' and ``negative'' examples). We wish to find a hyperplane
direction $\bfw\in R^n$ and an offset scalar $b$ such that
$\bfw\cdot\bfx_i-b> 0$ for positive examples and $\bfw\cdot\bfx_i-b<
0$ for negative examples --- which together means that the margins $y_i(\bfw\cdot\bfx_i-b)>0$ are positive.

Assuming that such a hyperplane exists, clearly it is not unique. We therefore need to introduce another constraint so
that we could find the most ``sensible'' solution among all (infinitley
many) possible
hyperplanes which separate the training data. 
Another issue is that
the framework  is very limited in the sense that for most
real-world classification problems it is somewhat unlikely that there
would exist a linear separating function to begin with. We therefore
need to find a way to extend the framework to include non-linear
decision boundaries at a reasonable cost. These two issues will be the
focus of this lecture.

Regarding the first issue, since there is more than one separating
hyperplane (assuming the training data is linearly separable) then the
question we need to ask ourselves is among all those solutions which
of them has the best ``generalization'' properties? In other words,
our goal in constructing a learning machine is not necessarily to do
very well (or perfect) on the training data, because the training data
is merely a sample of the instance space, and not necessarily a
``representative'' sample --- it is simply a sample. 
Therefore, doing well on the sample (the training data) does
not necessarily guarantee (or even imply) that we will do well on the
entire instance space. The goal of constructing a learning machine is to maximize the
performance on the test data (the instances we haven't seen), which in
turn means that we wish to generalize ``good'' classification
performance on the training set onto the entire instance space.

A related issue to generalization is that the distribution used to generate the training data is unknown. Unlike the statistical inference material we had so far, this time we will not attempt to estimate the distribution. The reason one can derive optimal learning algorithms yet bypass the need for estimating distributions would be explained later in the course when PAC-learning will be introduced. For now we will focus only on the algorithmic aspect of the learning problem.

The idea is to consider a subset $C_\gamma$ of all hyperplanes which have a fixed margin $\gamma$ where the margin is defined as the distance of the closest training point to the hyperplane:
$$\gamma = \min_i \left\{ \frac{y_i (\bfw^\top\bfx_i - b)}{\|\bfw\|}\right\}.$$
The Support Vector Machine (SVM), first introduced by Vapnik and his colleagues in 1992,  seeks a separating hyperplane which simultaneously minimizes the empirical error {\it and\/} maximizes the margin. The idea of maximizing the margin is intuitively appealing 
because a decision boundary which
lies close to some of the training instances is less likely to
generalize well because the learning machine will be susceptible to
small perturbations of those instance vectors. A formal motivation for this approach is deferred to the PAC-learning material we will introduce later in the course.

\section{Large Margin Classifier as a Quadratic Linear Programming}

We would first like to set up the linear separating hyperplane as an optimization problem which is both consistent with the training data and maximizes the margin induce by the separating hyperplane over all possible consistent hyperplanes.

Formally speaking, the distance between a point $\bfx$ and the
hyperplane is defined by
$$\frac{\mid \bfw\cdot\bfx - b\mid}{\sqrt{\bfw\cdot\bfw}}.$$
Since we are allowed to scale the parameters $\bfw,b$ at will (note
that if $\bfw\cdot\bfx - b>0$ so is $(\lambda\bfw)\cdot\bfx - (\lambda b)>0$ for
all $\lambda>0$) we can set the distance between the boundary points
to the hyperplane to be $1/\sqrt{\bfw\cdot\bfw}$ by scaling $\bfw,b$ such the point(s) with smallest margin (closest to the hyperplane) will be normalized:  $\mid \bfw\cdot\bfx - b\mid = 1$, therefore the
margin is simply $2/\sqrt{\bfw\cdot\bfw}$ (see Fig.~\ref{fig:one}). 
Note that $\argmaxx_{\bfw}
2/\sqrt{\bfw\cdot\bfw}$ is equivalent to $\argmaxx_{\bfw}
2/(\bfw\cdot\bfw)$ which in turn is equivalent to $\argminn_{\bfw}
\frac{1}{2}\bfw\cdot\bfw$. Since all positive points and negative
points should be farther away from the boundary points we also have
the separability constraints $\bfw\cdot\bfx - b \ge 1$ when $\bfx$ is a
positive instance and $\bfw\cdot\bfx  - b \le -1$ when $\bfx$ is a
negative instance. Both separability constraints can
be combined: $y(\bfw\cdot\bfx  - b) \ge 1$. Taken together, we have
defined the following optimization problem:

\begin{eqnarray}
\min_{\bfw,b}&& \frac{1}{2}\bfw\cdot\bfw\\
&&subject\ \ to\nonumber\\
&&y_i(\bfw\cdot\bfx_i - b)-1\ge 0\ \ \ \ \ i=1,...,m
\end{eqnarray}

This type of optimization problem has a quadratic criteria function
and linear inequalities and is known in the literature as a {\it
Quadratic Linear Programming\/} (QP) type of problem. 

This particular
QP, however, requires that the training data are linearly separable --- a
condition which may be unrealistic. We can relax this condition by
introducing the concept of a ``soft margin'' in which the separability
holds approximately with some error:

\begin{eqnarray}
\min_{\bfw,b,\epsilon_i}&& \frac{1}{2}\bfw\cdot\bfw +
\nu\sum_{i=1}^l\epsilon_i\label{eq:svm-primal}\label{eq:svm}\\ 
&&subject\ \ to\nonumber\\
&&y_i(\bfw\cdot\bfx_i - b)\ge 1-\epsilon_i\ \ \ \ \ i=1,...,m\nonumber\\
&&\epsilon_i\ge 0\nonumber
\end{eqnarray}

Where $\nu$ is some pre-defined weighting factor. The (non-negative) variables $\epsilon_i$ allow data points to be {\it
miss-classified\/} thereby creating an approximate
separation. Specifically, if $\bfx_i$ is a positive instance ($y_i=1$)
then the ``soft'' constraint becomes: 
$$\bfw\cdot\bfx_i - b\ge 1-\epsilon_i,$$
where if $\epsilon_i=0$ we are back to the original constraint where
$\bfx_i$ is either a boundary point or laying further away in the half
space assigned to positive instances. When $\epsilon_i>0$ the point
$\bfx_i$ can reside inside the margin or even in the half space
assigned to negative instances. Likewise, if $\bfx_i$ is a negative
instance ($y_i=-1$) then the soft constraint becomes:
$$\bfw\cdot\bfx_i - b\le -1 +\epsilon_i.$$
The criterion function penalizes (the $L_1$-norm) for non-vanishing
$\epsilon_i$ thus the overall system will seek a solution with few as
possible ``margin errors'' (see Fig.~\ref{fig:one}). Typically, when possible, an $L_1$ norm is preferable as the $L_2$ norm overly weighs high magnitude outliers which in some cases can dominate the energy function. Another note to make here is that strictly speaking the "right thing" to do is to penalize the margin errors based on the $L_0$ norm $\|\epsilon\|_0^0 = | \{i: \epsilon_i > 0\}|$, i.e., the number of non-zero entries, and drop the balancing parameter $\nu$. This is because it does not matter how far away a point is from the hyperplane --- all what matters is whether a point is classified correctly or not (see the definition of empirical error in Lecture 4). The problem with that is that the optimization problem would no longer be {\it convex\/} and non-convex problems are notoriously difficult to solve. Moreover, the class of convex optimization problems (as the one described in Eqn.~\ref{eq:svm}) can be solved in polynomial time complexity.

\begin{figure}
\begin{center}
\psfig{figure=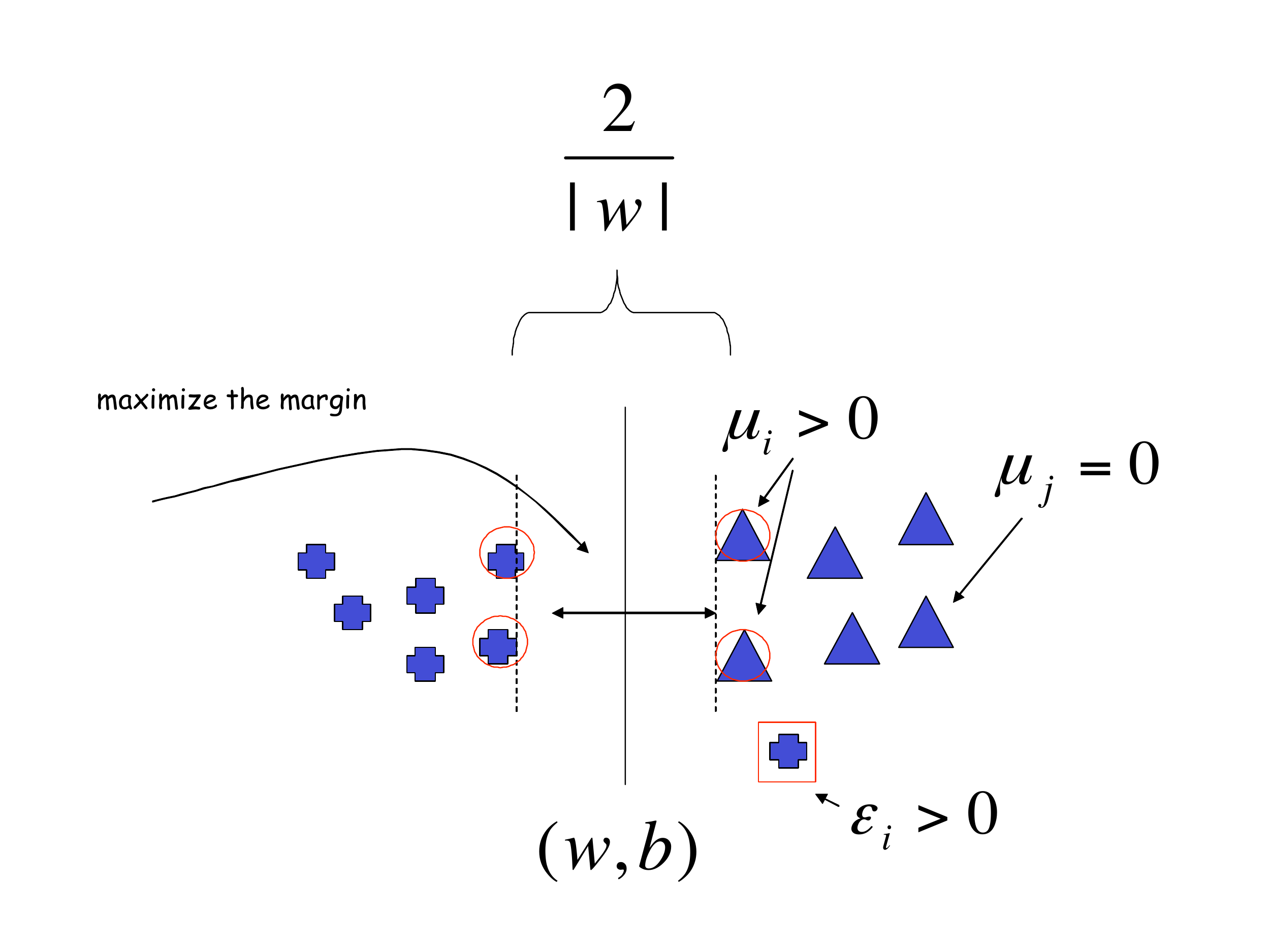,height={7cm}}
\end{center}
\caption{\protect\small Separating hyperplane $\bfw,b$ with maximal
margin. The boundary points are associated with non-vanishing Lagrange
multipliers $\mu_i>0$ and margin errors are associated with
$\epsilon_i>0$ where the criteria function encourages a small number
of margin errors.} 
\label{fig:one}
\end{figure}

So far we have described the problem formulation which {\it when
solved\/} would provide a solution with ``sensible'' generalization
properties. Although we can proceed using an off-the-shelf QLP solver, we will first pursue the "dual" problem. The dual form will highlight some key properties of the approach and will enable us to extend the framework to handle non-linear decision surfaces at a very little cost. In the appendix we take a brief tour on the basic principles associated
with constrained optimization, the Karush-Kuhn-Tucker (KKT) theorem
and the dual form. Those are recommended to read before moving to the next section.

\section{The Support Vector Machine}

We return now to the primal problem (eqn. 6.3) representing
the maximal margin 
separating hyperplane with margin errors: 

\begin{eqnarray}
\min_{\bfw,b,\epsilon_i}&& \frac{1}{2}\bfw\cdot\bfw +
\nu\sum_{i=1}^l\epsilon_i\nonumber\\
&&subject\ \ to\nonumber\\
&&y_i(\bfw\cdot\bfx_i - b)\ge 1-\epsilon_i\ \ \ \ \ i=1,...,m\nonumber\\
&&\epsilon_i\ge 0\nonumber
\end{eqnarray}

We will now derive the Lagrangian Dual of this problem. By doing so a
new key property will emerge facilitated by the fact that the criteria
function $\theta(\mu)$ (note there are no equality constraints thus
there is no need for $\lambda$) involves only inner-products of the
training instance vectors $\bfx_i$. This property will form the key of
mapping the original input space of dimension $n$ to a higher
dimensional space thereby allowing for non-linear decision surfaces
for separating the training data. 

Note that with this particular problem  the strong duality
conditions are satisfied because the criteria function and the
inequality constraints form a convex set.
The Lagrangian takes the following form:
$$L(\bfw,b,\epsilon_i,\mu)= \frac{1}{2}\bfw\cdot\bfw +
\nu\sum_{i=1}^m\epsilon_i - \sum_{i=1}^m \mu_i
\left[y_i(\bfw\cdot\bfx_i - b) - 1 + \epsilon_i\right]
- \sum_{i=1}^m \delta_{i}\epsilon_i
$$
Recall that 
$$\theta(\mu)=\min_{\bfw,b,\bfeps}L(\bfw,b,\bfeps,\bfmu,\bfdelta).$$
Since the minimum is obtained at the vanishing partial derivatives of
the Lagrangian with respect to $\bfw,b$, the next step would be to
evaluate those constraints and substitute them back into $L()$ to
obtain $\theta(\mu)$:
\begin{eqnarray}
\frac{\partial L}{\partial \bfw} &=& \bfw - \sum_i\mu_iy_i\bfx_i=0\label{eq:c1}\\
\frac{\partial L}{\partial b} &=& \sum_i \mu_iy_i=0\label{eq:c2}\\
\frac{\partial L}{\partial \epsilon_i} &=& \nu - \mu_i - \delta_{i}=0\label{eq:c3}
\end{eqnarray}  
From the first constraint (\ref{eq:c1}) we obtain
$\bfw=\sum_i\mu_iy_i\bfx_i$, that is, $\bfw$ is described by a linear
combination of a {\it subset\/} of the training instances. The reason
that not all instances participate in the linear superposition is due
to the KKT conditions: $\mu_i=0$ when $y_i(\bfw\cdot\bfx_i - b)> 1$,
i.e., the instance $\bfx_i$ is classified correctly and is not a
boundary point, and conversely, $\mu_i>0$ when $y_i(\bfw\cdot\bfx_i -
b) = 1 - \epsilon_i$, i.e., when $\bfx_i$ is a boundary point or when
$\bfx_i$ is a margin error ($\epsilon_i>0$) --- note that for a margin
error instance the value of $\epsilon_i$ would be the smallest
possible required to reach an equality in the constraint because the
criteria function penalizes large values of $\epsilon_i$. The boundary
points (and 
the margin errors) are
called {\it support vectors\/} thus $\bfw$ is defined by the
support vectors {\it only}. The third constraint (\ref{eq:c3}) is
equivalent to the constraint:
$$0\le \mu_i \le \nu\ \ \ \ \ i=1,...,l,$$
since $\delta_{i}\ge 0$. Also note that if $\epsilon_i > 0$, i.e., point $\bfx_i$ is a margin-error point, then by KKT conditions we must have $\delta_i=0$. As a result $\mu_i=\nu$. Therefore based on the values of $\mu_i$ alone we can make the following classifications:
\begin{itemize}
\item $0 < \mu_i < \nu$: point $\bfx_i$ is on the margin and is not a margin-error.
\item $\mu_i=\nu$: points $\bfx_i$ is a margin-error point.
\item $\mu_i=0$: point $\bfx_i$ is not on the margin.
\end{itemize}

Substituting these results/constraints back
into the Lagrangian $L()$ we obtain the {\it dual problem}:
\begin{eqnarray}
\max_{\mu_1,...,\mu_m}&& \theta(\bfmu)=\sum_{i=1}^m\mu_i - 
\frac{1}{2}\sum_{i,j}\mu_i\mu_jy_iy_j\bfx_i\cdot \bfx_j\label{eq:svm-dual}\\
&&subject\ \ to\nonumber\\
&&0\le \mu_i\le \nu\ \ \ \ \ i=1,...,m\nonumber\\
&&\sum_{i=1}^m y_i\mu_i=0\nonumber
\end{eqnarray}
The criterion function $\theta(\bfmu)$ can be written in a more compact
manner as follows: Let $M$ be a $l\times l$ matrix whose entries are
$M_{ij}=y_iy_j\bfx_i\cdot\bfx_j$ then $\theta(\bfmu)=\bfmu^\top{\bf 1} -
\frac{1}{2}\bfmu^\top M\bfmu$ where ${\bf 1}$ is the vector of $(1,...,1)$
and $\bfmu$ is the vector $(\mu_1,...,\mu_m)$ and $\mu^\top$ is the
transpose (row vector). Note that $M$ is {\it positive definite\/},
i.e., $\bfx^\top M\bfx> 0$ for all vectors $\bfx\not = 0$ --- a property
which will be important later.

The key feature of the dual problem is not so much that it is simpler
than the primal (in fact it isn't since the primal has no equality
constraints) or that it has a more ``elegant'' feel, the key feature
is that the problem is completely described by the inner products of
the training instances $\bfx_i$, $i=1,...,m$. This fact will be shown
to be a crucial ingredient in the so called ``kernel trick'' for the
computation of inner-products in high dimensional spaces using simple
functions defined on pairs of training instances.

\section{The Kernel Trick}

We ended with the dual formulation of the SVM problem and noticed that the input data vectors $\bfx_i$ are represented by the Gram matrix $M$. In other words, only inner-products of the input vectors play a role in the dual formulation --- there is no explicit use of $\bfx_i$ or any other function of $\bfx_i$ besides inner-products. This observation suggests the use of what is known as the "kernel trick" to replace the inner-products by non-linear functions.

The common principle of kernel methods is to construct nonlinear variants of linear algorithms by substituting inner-products by nonlinear kernel functions. Under certain conditions this process can be interpreted as mapping of the original measurement vectors (so called "input space") onto some higher dimensional space (possibly infinitely high) commonly referred to as the "feature space". 
Mathematically, the kernel approach is defined as follows: let $\bfx_1,...,\bfx_l$ be vectors in the input space, say $R^n$, and consider a mapping  $\phi(\bfx):R^n\rightarrow
{\cal F}$ where ${\cal F}$ is an inner-product space. The kernel-trick is to calculate the inner-product in ${\cal F}$ using a kernel function $k: R^n \times R^n \rightarrow R$, $k(\bfx_i,\bfx_j)=\phi(\bfx_i)^\top\phi(\bfx_j)$, while avoiding explicit mappings (evaluation of) $\phi()$.

Common choices of kernel selection include the d'th order polynomial kernels  $k(\bfx_i,\bfx_j)=(\bfx_i^\top \bfx_j + \theta)^d$ and the Gaussian RBF kernels $k(\bfx_i,\bfx_j) = \exp(-\frac{1}{2\sigma^2}\|\bfx_i-\bfx_j\|^2)$. If an algorithm can be restated such that the input vectors appear in terms of inner-products only, one can substitute the inner-products by such a kernel function. The resulting kernel algorithm can be interpreted as running the original algorithm on the space ${\cal F}$ of mapped objects $\phi(\bfx)$. 

We know that $M$ of the dual form is positive semi-definite because $M$ can be written is $M=Q^\top Q$ where $Q=[y_1\bfx_1,...,y_l\bfx_l]$. Therefore $\bfx^\top M\bfx = \|Q\bfx\|^2\ge 0$ for all choices of $\bfx$ (which means that the eigenvalues of $M$ are non-negative). If the entries of $M$ are to be replaced with $y_iy_jk(\bfx_i,\bfx_j)$ then the condition we must enforce on the function $k()$ is that it is a {\it positive definite kernel\/} function. A positive definite function is defined such that for any set of vectors $\bfx_1,...,\bfx_q$ and for any values of $q$ the matrix $K$ whose entries are $K_{ij}=k(\bfx_i,\bfx_j)$ is positive semi-definite. Formally, the conditions for admissible kernels $k()$ are known as Mercer's conditions summarized below:
\begin{theorem}[Mercer's Conditions]
Let $k(x,y)$ be symmetric and continuous. The following conditions are equivalent:
\begin{enumerate}
\item $k(x,y) = \sum_{i=1}^\infty \alpha_i \phi_i(x)\phi_i(y) = \phi(x)^\top \phi(y)$ for any uniformly converging series $\alpha_i>0$.
\item for all $\psi()$ satisfying $\int_x \psi^2(x)dx < \infty$, then $$\int_x\int_y k(x,y)\psi(x)\psi(y)dxdy \ge 0$$
\item for all $\{\bfx_i\}_{i=1}^q$ and for all $q$, the matrix $K_{ij} = k(x_i,x_j)$ is positive semi-definite.
\end{enumerate}
\end{theorem}
Perhaps the non-obvious condition is No. 1 which allows for the feature map $\phi()$ to have infinitely many coordinates (a vector in Hilbert space). For example, as we shall see below, the kernel $\exp(-\frac{1}{2\sigma^2}\|\bfx_i-\bfx_j\|^2)$ is an inner-product of two vectors with infinitely many coordinates.
We will consider next a number of popular kernels.

\subsection{The Homogeneous Polynomial Kernel}

Let $\bfx,\bfy\in R^k$ and define $k(\bfx,\bfy)=(\bfx^\top\bfy)^d$ where $d>0$ is a natural number. Then, the corresponding feature map $\phi(\bfx)$ has ${k+d-1\choose d} = O(k^d)$ coordinates which take the value:
$$\phi(\bfx) = \left( \sqrt{{d \choose {n_1,...,n_k}}} x_1^{n_1}\cdot\cdot\cdot x_k^{n_k}\right)_{n_i\ge0, \sum_i n_i=d}$$
where ${d \choose {n_1,...,n_k}}= d!/(n_1!\cdot\cdot\cdot n_k!)$ is the multinomial coefficient (number of ways to distribute $d$ balls into $k$ bins where the $j$'th bin hold exactly $n_j\ge 0$ balls):
$$(x_1+...+x_k)^d = \sum_{n_i\ge0, \sum_i n_i=d} {d \choose {n_1,...,n_k}} x_1^{n_1}\cdot\cdot\cdot x_k^{n_k}.$$
The dimension of the vector space $\phi(\bfx)$ where $\bfx\in R^k$ can be measured using the following combinatorial problem: how many arrangements of $k-1$ partitions to be placed among $d$ items? the answer is ${k+d-1\choose k-1} = {k+d-1\choose d} = O(k^d)$. For example, $k=d=2$ gives us :
$$(\bfx^\top\bfy)^2 = x_1^2y_1^2 + 2x_1x_2y_1y_2 + x_2^2y_2^2 = \phi(\bfx)^\top \phi(\bfy),$$
where $\phi(\bfx)=(x_1^2, x_2^2, \sqrt{2}x_1x_2)$.

\subsection{The non-homogeneous Polynomial Kernel}

The feature map $\phi(\bfx)$ contains all monomials whose power is lesser or equal to $d$, i.e., $\sum_i n_i \le d$. This can be acheived by increasing the dimension to $k+1$ where $n_{k+1}$ is used to fill the gap between $\sum_{i=1}^k n_i < d$ and $d$. Therefore the dimension of $\phi(\bfx)$ where $\bfx\in R^k$ would be ${k+d\choose d}$. We have:
\bea
(\bfx^\top\bfy + \theta)^d &=& (x_1y_1+...+x_ky_k+\sqrt{\theta}\sqrt{\theta})^d\nonumber\\
&=& \sum_{n_i\ge 0, \sum_{i=1}^{k+1}n_i=d}{d \choose {n_1,...,n_{k+1}}}x_1^{n_1}y_1^{n_1}\cdot\cdot\cdot x_1^{n_k}y_1^{n_k}\cdot\theta^{n_{k+1}/2}\theta^{n_{k+1}/2}\nonumber
\eea
Therefore, the entries of the vector $\phi(\bfx)$ take the values:
$$\phi(\bfx) = \left( \sqrt{{d \choose {n_1,...,n_{k+1}}}} x_1^{n_1}\cdot\cdot\cdot x_k^{n_k}\cdot\theta^{n_{k+1}/2}\right)_{n_i\ge0, \sum_{i=1}^{k+1} n_i=d}$$
For example, $k=d=2$ gives us :
$$(\bfx^\top\bfy+\theta)^2 = x_1^2y_1^2 + 2x_1x_2y_1y_2 + x_2^2y_2^2 + 2\theta x_1y_1 + 2\theta x_2y_2+ \theta = \phi(\bfx)^\top \phi(\bfy),$$
where $\phi(\bfx)=(x_1^2, x_2^2, \sqrt{2}x_1x_2, \sqrt{2\theta} x_1, \sqrt{2\theta}x_2, \sqrt{\theta})$.
In this example, $\phi()$ is a mapping from $R^2$ to $R^6$ and 
hyperplanes $\phi(\bfw)^\top\phi(\bfx) -b=0$ in $R^6$ correspond to {\it conics\/} in $R^2$:
$$(w_1^2)x_1^2 + (w_2^2)x_2 + (2w_1w_2)x_1x_2 + (2\theta w_1)x_1 + (2\theta w_2)x_2 + (\theta - b) = 0$$ 
Assume we would like to find a separating {\it conic\/} (Parabola, Hyperbola, Ellipse) function rather than a line in $R^2$. The discussion so far suggests we construct the Gram matrix $M$ in the dual form with the $d=2$ polynomial kernel $k(\bfx,\bfy)=(\bfx^\top\bfy+\theta)^2$ for some parameter $\theta$ of our choosing. The extra effort we will need to invest is negligible --- simply replace every occurrence $\bfx_i^\top\bfx_j$ with $(\bfx_i^\top\bfx_j+\theta)^2$.

\subsection{The RBF Kernel}

The function $k(\bfx,\bfy)=e^{- \|\bfx-\bfy\|^2/2\sigma^2}$ known as a {\it Radial Basis Function\/} (RBF) is a kernel function but with an infinite expansion. Without loss of generality let $\sigma=1$, then we have:
\bea
e^{- \|\bfx-\bfy\|^2/2}&=&e^{-\|\bfx\|^2/2}e^{-\|\bfy\|^2/2}e^{\bfx^\top\bfy}\nonumber\\
&=& \sum_{j=0}^\infty \frac{(\bfx^\top\bfy)^j}{j!}e^{-\|\bfx\|^2/2}e^{-\|\bfy\|^2/2}\nonumber\\
&=& \sum_{j=0}^\infty  \left( \frac{e^{-\frac{\|\bfx\|^2}{2j}}}
{\sqrt{j!}^{1/j}}  
\frac{e^{-\frac{\|\bfy\|^2}{2j}}}{\sqrt{j!}^{1/j}}
{\bfx^\top\bfy}\right)^j\nonumber\\
&=&\sum_{j=0}^\infty \sum_{\sum_i n_i=j} \frac{e^{-\frac{\|\bfx\|^2}{2j}}}
{\sqrt{j!}^{1/j}}  {j \choose {n_1,...,n_k}}^{1/2} x_1^{n_1}\cdot\cdot\cdot x_k^{n_k} 
\frac{e^{-\frac{\|\bfy\|^2}{2j}}}
{\sqrt{j!}^{1/j}}  {j \choose {n_1,...,n_k}}^{1/2} y_1^{n_1}\cdot\cdot\cdot y_k^{n_k}\nonumber
\eea
From which we can see that the entries of the feature map $\phi(\bfx)$ are:
$$\phi(\bfx) = \left( \frac{e^{-\frac{\|\bfx\|^2}{2j}}}
{\sqrt{j!}^{1/j}}  {j \choose {n_1,...,n_k}}^{1/2} x_1^{n_1}\cdot\cdot\cdot x_k^{n_k} 
 \right )_{j=0,..,\infty, \sum_{i=1}^k n_i = j}$$
 
 \subsection{Classifying New Instances}

By adopting some kernel $k()$ we are in fact mapping $\bfx \rightarrow \phi(\bfx)$, thus
we then proceed to solve for $\phi(\bfw)$ and $b$ using some QLP solver. The QLP solution of the dual form will yield the solution for the Lagrange multipliers $\mu_1,...,\mu_m$. We saw from eqn. (\ref{eq:c1}) that we can express $\phi(\bfw)$ as a function of the (mapped) examples:
$$\phi(\bfw)=\sum_i\mu_iy_i\phi(\bfx_i).$$
Rather than explicitly representing $\phi(\bfw)$ --- a task which may be prohibitly expensive since in general the dimension of the feature space of a polynomial mapping is $k+d\choose d$ --- we store all the support vectors (those input vectors with corresponding $\mu_i>0$) and use them for the evaluation of test examples:
\begin{eqnarray}
f(\bfx) &=& sign(\phi(\bfw)^\top\phi(\bfx) - b)=sign(\sum_i \mu_iy_i\phi(\bfx_i)^\top\phi(\bfx)-b)\nonumber\\
&=& sign(\sum_i \mu_iy_ik(\bfx_i,\bfx) - b).\nonumber
\end{eqnarray}
We see that the kernel trick enabled us to look for a non-linear separating surface by making an implicit mapping of the input space onto a higher dimensional feature space using the same dual form of the SVM formulation --- the only change required was in the way the Gram matrix was constructed. The price paid for this convenience is to carry {\it all\/} the support vectors at the time of  classification $f(\bfx)$. 

A couple of notes may be worthwhile at this point. The constant $b$ can be recovered from any of the support vectors. Say, $\bfx^+$ is a positive support vector (but not a margin error, i.e., $\mu_i < \nu$). Then $\phi(\bfw)^\top\phi(\bfx^+)-b=1$ from which $b$ can be recovered. The second note is that the number of support vectors is typically around 10\% of the number of training examples (empirically). Thus the computational load during evaluation of $f(\bfx)$ may be relatively high. Approximations have been proposed in the literature by looking for a reduced number of support vectors (not necessarily aligned with the training set) --- but this is beyond the scope of this course.

The kernel trick gained its popularity with the introduction of the SVM but since then has taken a life of its own and has been applied to principal component analysis (PCA), ridge regression, canonical correlation analysis (CCA), QR factorization and the list goes on. We will meet again with the kernel trick later on.

\chapter{Spectral Analysis I: PCA, LDA, CCA}
\label{chap:8}

In this lecture (and the following one) we will focus on {\it spectral methods\/} for learning. Today we will focus on dimensionality reduction using Principle Component Analysis (PCA), multi-class learning using Linear Discriminant Analysis (LDA) and Canonical Correlation Analysis (CCA). In the next lecture we will focus on spectral clustering methods.

Dimensionality reduction appears when the dimension of the input vector is very large (imagine pixels in an image, for example) while the coordinate measurements are highly inter-dependent (again, imagine the redundancy present among neighboring pixels in an image). High dimensional data impose computational efficiency challenges and often  translate to poor generalization abilities of the learning engine (see lectures on PAC). A dimensionality reduction can also be viewed as a {\it feature extraction\/} process where one takes as input a large feature set (the original measurements) and creates from them a much smaller number of new features which are then fed into the learning engine.

In this lecture we will focus on feature extraction from a very specific (and constrained) stanpoint. We would be looking for a mixing (linear combination) of the input coordinates such that we obtain a linear projection from $R^n$ to $R^q$ for some $q < n$. In doing so we wish to reduce the redundancy while preserving as much as possible the variance of the data. From a statistical standpoint this is achieved by transforming to a new set of variables, called principal components, which are uncorrelated so that the first few retain most of the variation present in all of the original coordinates. For example, in an image
processing application the input images are highly redundant where
neighboring pixel values are highly correlated. The purpose of feature
extraction would be to transform the input image into a vector of
output components with the least redundancy possible.
Form a geometric standpoint, this is achieved by finding the "closest" (in least squares sense) linear $q$-dimensional susbspace to the $m$ sample points $S$. The new subspace is a lower dimensional "best approximation" to the sample $S$. These two, equivalent, perspectives on data compression (dimensionality reduction) form the central idea of {\it principal component analysis\/} (PCA) which probably the oldest (going back to Pearson 1901) and best known of the techniques of multivariate analysis in statistics. The computation of PCA is very simple and the definition is straightforward, but has a wide variety of different applications, a number of different derivations, quite a number of different terminologies (especially outside the statistical literature) and is the basis for quite a number of variations on the basic technique.

We then extend the variance preserving approach for data representation for {\it labeled\/} data sets. We will describe the linear classifier approach (separating hyperplane) form the point of view of looking for a hyperplane such that when the data is projected onto it the separation is maximized (the distance between the class means is maximal) and the data within each class is compact (the variance/spread is minimized). The solution is also produced, just like PCA, by a spectral analysis of the data. This approach goes under the name of Fisher's Linear Discriminant Analysis (LDA).

What is common between PCA and LDA  is  (i) the use of spectral matrix analysis --- i.e., what can you do with eigenvalues and eigenvectors of matrices representing subspaces of the data?  (ii) these techniques produce optimal results for {\it normally distributed\/}  data and are very easy to implement. There is a large variety of uses of spectral analysis in statistical and learning literature including spectral clustering, Multi Dimensional Scaling (MDS) and data modeling in general. Another point to note is that this is the first time in the course where the type of data distribution plays a role in the analysis --- the two techniques are defined for any distribution but are optimal only under the Gaussian distribution.

We will also describe a non-linear extension of PCA known as
Kernel-PCA, but the focus would be mostly on  PCA itself and its
analysis from a couple of vantage points: (i) PCA as an optimal
reconstruction after a dimension reduction, i.e., data compression,
and (ii) PCA for redundancy reduction (decorrelation) of the output
components. 

\section{PCA: Statistical Perspective}

Let $\bfx_1,...,\bfx_m\in R^n$ be our sample data $S$ of vectors in
$R^n$, arranged as columns of a matrix $A$. It will be convenient to
assume that the data is centered, i.e., $\sum\bfx_i = 0$. If the data
is not centered we can always center it by computing the mean vector
$\mu=(1/m)\sum_i\bfx_i$ and replace the original data sample with the new
sample $\bfx_i - \mu$. In a statistical sense,
the coordinates of the vector $\bfx\in R^n$ are considered as random
variables, thus a row in the matrix $A$ is the sample of values of a
particular random variable, drawn from some unknown probability
distribution, associated with the row position. We wish 
to find vectors $\bfu_1,...,\bfu_q$ (arranged as columns
of a matrix $U$), where $q\le min(n,m)$, such that the new feature measurements $\bfy=U^\top\bfx$ (who are the result of linear combinations $\bfu_1^\top\bfx,...,\bfu_q^\top\bfx$ of the original feature measurements $\bfx$)  have certain desirable
properties. 

The idea property to seek from the new coordinates $\bfy$ is statistical independence, i.e., $P(y_1,..,y_q)=P(y_1)\cdot\cdot\cdot P(y_q)$ which would mean that we have removed the redundancy of the original data $\bfx$ in the best possible manner. This goal, however, is too much to ask from a linear transformation and instead we would ask for a weaker property to hold: that the pairwise covariance $cov(y_i,y_j)=0$ vanishes, i.e., that the covariance matrix on the new coordinates is diagonal. A diagonal covariance insures some redundancy removal, but not as good as statistical independence. However, when the data is Normally distributed $P(\bfx)\sim N(\mu, \Sigma)$ with mean $\mu$ and covariance $\Sigma$, then the transformation which diagonalizes the covariance matrix also guarantees statistical independence. Among all transformations that de-correlate the data we will seek the one that maximizes the {\it spread\/} (variance) of the sample data after being projected onto the new axes vectors.

\subsection{Maximizing the Variance of Output Coordinates}

The 
property we would like to maximize is that the projection of the sample
data on the new axes is as {\it spread\/} as possible.
To start this analysis, assume $q=1$, i.e., the $n$ components of the
input vector $\bfx$ are reduced to a single output component
$y=\bfu^\top\bfx$. We are looking for a
single vector $\bfu\in R^n$ whose direction {\it maximizes the variance\/} of
the output component $y$. 

Formally, we
are looking for a unit vector $\bfu$ which maximizes $\sum_i
(\bfu^\top \bfx_i)^2$ (see Appendix A for basic statistical definitions and note that $E[y]=0$ because $\sum_i\bfu^\top \bfx_i=\bfu_i^\top(\sum_i\bfx_i)=0$). In other words, the projected points onto the
axis represented by the vector $\bfu$ are as spread as possible (in a
least squares sense). In vector notation, the optimization problem
takes the following form:
$$
\max_{\bfu} \frac{1}{2}\|\bfu^\top A\|^2\ \ \ \ \ subject\ to\ \ \ \
\frac{1}{2}\bfu^\top\bfu=1
$$
The Lagrangian of the problem is:
$$L(\bfu,\lambda)=\frac{1}{2}\bfu^\top AA^\top\bfu
-\lambda(\frac{1}{2}\bfu^\top\bfu -1)$$
By taking the partial derivative $\partial L/\partial\bfu=0$ we obtain
the following necessary condition (see Appendix B):
$$AA^\top \bfu = \lambda\bfu,$$
which tells us that $\bfu$ is an eigenvector of the $n\times n$
(symmetric and positive definite) matrix
$AA^\top$. There are $n$ eigenvectors associated with $AA^\top$ and we
can easily convince ourselves that we are looking for the one
associated with the maximal eigenvalue: substitute $\lambda\bfu$
instead of $AA^\top\bfu$ in the criterion function $\bfu^\top
AA^\top\bfu$ to obtain $\lambda(\bfu^\top\bfu)=\lambda$ and since the
eigenvalues must be positive (since $AA^\top$ is positive definite),
then the optimum is obtained for the maximal eigenvalue. The leading
eigenvector $\bfu$ of $AA^\top$ is called the {\it first principal axis\/} of
the data sample represented by the columns of the matrix $A$, and $y=\bfu^\top\bfx$ is called the first {\it principal component\/} of the data sample.

For convenience, we denote $\bfu_1=\bfu$ and $\lambda_1=\lambda$ as the leading eigenvector and eigenvalue of $AA^\top$. Next, we look for $y_2=\bfu_2^\top\bfx$ which is {\it uncorrelated\/} with $y_1=\bfu_1^\top\bfx$ and which has maximum variance (and so on for $\bfu_3,...,\bfu_q$). Two random variables are uncorrelated if their covariance vanishes. By definition of covariance (see Appendix A) we obtain:
\begin{eqnarray*}
Cov(y_1y_2)&=&\sum_i (\bfu_1^\top\bfx_i)(\bfu_2^\top\bfx_i)=\bfu_1^\top(\sum_i \bfx_i\bfx_i^\top)\bfu_2\\
&=& \bfu_1^\top AA^\top\bfu_2= \bfu_2^\top AA^\top\bfu_1=\lambda_1\bfu_1^\top\bfu_2=0
\end{eqnarray*}
We can therefore use the condition $\bfu_1^\top\bfu_2=0$ to specify zero correlation between $y_1,y_2$. The functional to be optimized becomes:
$$
\max_{\bfu_2} \frac{1}{2}\|\bfu_2^\top A\|^2\ \ \ \ \ subject\ to\ \ \ \
\frac{1}{2}\bfu_2^\top\bfu_2=1,\ \ \  \bfu_1^\top\bfu_2=0,
$$
with the Lagrangian being:
$$L(\bfu_2,\lambda,\delta)=\frac{1}{2}\bfu_2^\top AA^\top\bfu_2
-\lambda(\frac{1}{2}\bfu_2^\top\bfu_2 -1) - \delta \bfu_1^\top\bfu_2.$$
By taking the partial derivative with respect to $\bfu_2$ we obtain the necessary condition:
$$AA^\top \bfu_2 - \lambda\bfu_2 - \delta\bfu_1=0.$$
Multiply the equation by $\bfu_1$ from the left:
$$\bfu_1^\top AA^\top \bfu_2 - \lambda\bfu_1^\top\bfu_2 - \delta\bfu_1^\top\bfu_1=0,$$
and noting from above that $\bfu_1^\top AA^\top \bfu_2=\bfu_1^\top\bfu_2=0$ we obtain $\delta=0$. As a result we obtain: 
$$AA^\top \bfu_2 = \lambda\bfu_2,$$
so once more we have that $\lambda,\bfu_2$ form an eigenvalue/eigenvector pair of $AA^\top$. As before, $\lambda$ should be as large as possible from the remaining spectral decomposition.  By induction, it can be shown that the remaining principal vectors $\bfu_3,...,\bfu_q$ are the decreasing order eigenvactors of $AA^\top$ and the variance of the $i$'th principal component $y_i=\bfu_i^\top\bfx$ is $\lambda_i$. 

Taken together, the PCA is the solution of the following optimization problem:
$$
\max_{\bfu_1,...,\bfu_q} \frac{1}{2}\sum_i\|\bfu_i^\top A\|^2\ \ \ \ \ subject\ to\ \ \ \
\bfu_i^\top\bfu_i=1,\ \ \  \bfu_i^\top\bfu_j=0,\ \ \ i\not = j=1,...,q.
$$
It will be useful for later to write the optimization function in a more concise manner as follows. Let
$U$ be the $n\times q$ matrix whose columns are $\bfu_i$ and $D=diag(\lambda_1,...,\lambda_q)$ is an $q\times q$ diagonal matrix
and $\lambda_1\ge \lambda_2\ge...\ge \lambda_q$. Then from above we have that $U^\top U=I$ and $AA^\top U=UD$. Using the fact that $trace(\bfx\bfy^\top)=\bfx^\top\bfy$, $trace(AB)=trace(BA)$ and $trace(A+B)=trace(A)+trace(B)$ we can convert $\sum_i\|\bfu_i^\top A\|^2$ to $trace(U^\top AA^\top U)$ as follows:
\begin{eqnarray*}
\sum_i \bfu_i^\top AA^\top \bfu_i &=& \sum_i trace(A^\top \bfu_i\bfu_i^\top A)=trace(A^\top(\sum_i \bfu_i\bfu_i^\top) A)\\
&=& trace(A^\top UU^\top A) = trace(U^\top AA^\top U)
\end{eqnarray*}
Thus, PCA becomes the solution of the following optimization function:
\be
\max_{U\in R^{n\times q}} trace(U^\top AA^\top U)\ \ \ \ \ subject\ to\ \ \ \ U^\top U=I.
\label{eq:1}
\ee
The solution, as saw above, is that $U=[\bfu_1,...,\bfu_q]$ consists of the decreasing order eigenvectors of $AA^\top$. At the optimum, $trace(U^\top AA^\top U)$ is equal to $trace(D)$ which is equal to the sum of eigenvalues $\lambda_1+...+\lambda_q$.

It is worthwhile noting that  
when $q=n$, $UU^\top=U^\top U=I$, and the PCA transform is a change of basis in $R^n$ known
as Karhunen-Loeve transform. 

To conclude, the PCA transform looks for $q$ orthogonal direction
vectors (called 
the principal axes) such that the projection of input sample
vectors onto the principal directions has the maximal spread, or
equivalently that the variance of the output coordinates $\bfy=U^\top\bfx$ is
maximal. The principal directions are the leading (with respect to
descending eigenvalues) $q$ eigenvectors of the matrix $AA^\top$. When
$q=n$, the principal directions form a basis of $R^n$ with the
property of de-correlating the data and maximizing the variance of the coordinates of the sample
input vectors.

\subsection{Decorrelation: Diagonalization of the Covariance Matrix}

In the previous section we saw that PCA generates a new coordinate system $\bfy=U^\top\bfx$ where the coordinates $y_1,...,y_q$ of $\bfx$ in the new system are {\it uncorrelated}. This means that the covariance matrix over the principle components should be diagonal. In this section we will explore this perspective in more detail.

The covariance matrix ${\bfsig}_x$ of the sample data $\bfx_1,...,\bfx_m$ with zero
mean is 
$$(1/m)\sum_i \bfx_i\bfx_i^\top=(1/m)AA^\top,$$
therefore the matrix $AA^\top$ we derived above is a scaled version of
the covariance of the sample data (see Appendix A). The scale factor $1/m$ was unimportant in
the process above because the eigenvectors are of unit norm, thus any
scale of $AA^\top$ would produce the same set of eigenvectors.

The off-diagonal entries of the covariance matrix ${\bfsig}_x$ represent
the correlation (a measure of statistical dependence) between the i'th
and j'th component vectors, i.e., the 
entries of the input vectors $\bfx$. The existence of correlations among the
components (features) of the input signal is a sign of redundancy,
therefore from the point of view of transforming the input
representation into one which is {\it less\/} redundant, we would like
to find a transformation $\bfy=U^\top\bfx$ with an output
representation $\bfy$ which is associated with a diagonal covariance matrix ${\bfsig}_y$,
i.e., the components of $\bfy$ are uncorrelated.

Formally, ${\bfsig}_y=(1/m)\sum_i \bfy_i\bfy_i^\top = (1/m)U^\top
AA^\top U$, therefore we wish to find an $n\times q$ matrix for which $U^\top
AA^\top U$ is diagonal. If in addition, we would require that the {\it
variance\/} of the output coordinates is maximized, i.e., $trace(U^\top
AA^\top U)$ is maximal (but then we need to constrain the length of
the column vectors of $U$, i.e., set $\|\bfu_i\|=1$) then we would get
a unique solution for $U$ where the columns are orthonormal and are
defined as the first $q$ eigenvectors of the covariance matrix
${\bfsig}_x$. This is exactly the optimization problem defined by eqn.~(\ref{eq:1}).

We see therefore that PCA  ``decorrelates'' the input
data. Decorrelation and statistical independence are not the same
thing. If the coordinates are statistically independent then the
covariance matrix is diagonal\footnote{$\sigma_{xy}=\sum_x\sum_y(x-\mu_x)(y-\mu_y)p(x,y)=\sum_x\sum_y(x-\mu_x)(y-\mu_y)p(x)(p(y)=(\sum_x(x-\mu_x)p(x))(\sum_y(y-\mu_y)p(y))=0$}, but it does not follow that
uncorrelated variables must be statistically independent ---
covariance is just one measure of dependence. In fact, the covariance
is a measure of pairwise dependency only. However, it is a fact that
uncorrelated variables are statistically independent if they have a
multivariate normal distribution (a Gaussian). In other words, if the
sample data $\bfx$ are drawn from a probability distribution $p(\bfx)$
which has Gaussian form, the PCA transforms the sample data into a
statistically independent set of variables $\bfy=U^\top\bfx$. The
details are explained below.

Recall that a multivariate normal distribution of the random variables
$\bfx=(x_1,...,x_n)^\top$ is defined as $p(\bfx)\approx N(\mu,\bfsig)$:
$$p(\bfx)=\frac{1}{(2\pi)^{n/2}|\bfsig|^{1/2}}e^{-\frac{1}{2}(\bfx -
\mu)^\top \bfsig^{-1}(\bfx - \mu)}.$$
Also recall that a linear combination of the variables produces also a
normal distribution $N(U^\top \mu, U^\top \bfsig U)$:
$$\bfsig_y = \sum_{\bfy}(\bfy - \mu_y)(\bfy - \mu_y)^\top = \sum_{\bfx}(U^\top\bfx - U^\top\mu_x)(U^\top\bfx - U^\top\mu_x)^\top = U^\top \bfsig_x U,$$
therefore 
choose $U$ such that $\bfsig_y=U^\top \bfsig U$ is a diagonal matrix
$\bfsig_y=diag(\sigma_1^2,...,\sigma_n^2)$. We have in that case:
$$p(\bfx)=\frac{1}{(2\pi)^{n/2}\prod_i \sigma_i}e^{-\frac{1}{2}
\sum_i\left(\frac{x_i-\mu_i}{\sigma_i}\right)^2}$$
which can be written as a product of univariate normal distributions $p_{x_i}(x_i)$:
$$p(\bfx)=\prod_{i=1}^n \frac{1}{(2\pi)^{1/2}\sigma_i}e^{-\frac{1}{2}
\left(\frac{x_i-\mu_i}{\sigma_i}\right)^2}=\prod_{i=1}^n p_{x_i}(x_i),$$
which proves the assertion that decorrelated normally distributed variables
are statistically independent.

\section{PCA: Optimal Reconstruction}

A different, yet equivalent, perspective on the PCA transformation is
as an optimal reconstruction (in a least squares sense) after a
dimension reduction. We are given a sample data as before
$\bfx_1,...,\bfx_m$ and we are looking for a {\it small\/} number of
orthonormal principal vectors $\bfu_1,...,\bfu_q$ where $q<min(n,k)$ 
which define a q-dimensional linear subspace of $R^n$ which 
{\it
best\/} approximate the original input vectors in a least squares
sense. In other words, the projection $\hat{\bfx_i}$ of the sample points $\bfx_i$
onto the q-dimensional subspace should minimize $\sum_i \|\bfx_i -
\hat{\bfx_i}\|^2$ over all possible q-dimensional subspaces of $R^n$. 

Let ${\cal U}$ be the subspace spanned by the principal vectors (columns of
$U$) and let $P$
be the $n\times n$ projection matrix mapping a point $\bfx\in 
R^n$ onto its projection $\hat{\bfx}\in {\cal U}$. From the definition of
projection, the vector $\bfx - \hat{\bfx}$ must be orthogonal to the
subspace ${\cal U}$. Let $\bfy=(y_1,...,y_q)$ be the coordinates of
$\hat{\bfx}$ with respect to the principal vectors, i.e., $U\bfy
=\hat{\bfx}$. Then, from orthogonality we have that $(\bfx -
U\bfy)^\top U\bfw = 0$ for all vectors $\bfw\in R^n$. Since this is
true for all $\bfw$ then $U^\top U\bfy- U^\top \bfx=0$. Therefore,
$\bfy=(U^\top U)^{-1}U^\top\bfx$ and as a result the projection matrix
$P$ becomes:
$$P= U(U^\top U)^{-1}U^\top,$$
satisfying $P\bfx= \hat{\bfx}$. In the case the columns of $U$ are
orthonormal, $U^\top U=I$, we have $P=UU^\top$. We are ready now to
describe the optimization problem on $U$: we wish to find an
orthonormal set of principal vectors, $U^\top U=I$, such that $\sum_i
\|\bfx_i - UU^\top\bfx_i\|^2$ is minimized.

Note that $\sum_i
\|\bfx_i - UU^\top\bfx_i\|^2= \|A - UU^\top A\|^2_F$ where $\|B\|^2_F
=\sum_{i,j} b^2_{ij}$ is the square {\it Frobenious\/} norm of a
matrix. The optimal reconstruction problem therefore becomes:
$$\min_U \|A - UU^\top A\|^2_F\ \ \ \ \ subject\ to\ \ \ \  U^\top U=I.$$
We will show now that:
$$\argmin{U} \|A - UU^\top A\|^2_F =\argmax{U} trace(U^\top AA^\top U),$$
which shows that the optimal reconstruction problem is solved by PCA
(recall Eqn.~\ref{eq:1}). 

From the identity $\|B\|^2_F=trace(BB^\top)$, we have:
$$ \|A - UU^\top A\|^2_F = trace((A-UU^\top A)(A-UU^\top A)^\top).$$
Expanding the right hand side gives us:
\begin{eqnarray*}
trace((A-UU^\top A)(A-UU^\top A)^\top) &=& trace (AA^\top) - trace(AA^\top UU^\top)\\
&-& trace(UU^\top AA^\top) + trace(UU^\top AA^\top UU^\top)
\end{eqnarray*}
The second and third term are equal (commutativity of trace) and is also equal to the 4th term due to commutativity of the trace and $U^\top U=I$. Taken together:
$$\|A - UU^\top A\|^2_F=trace (AA^\top) 
-trace(U^\top AA^\top U).$$
To conclude, we have proven that by taking the first $q$
eigenvectors of $AA^\top$ we obtain a linear subspace which is {\it as
close as possible\/} (in a least squares sense) to the original sample
data. Hence, PCA can be viewed as a vehicle for optimal reconstruction
after dimension reduction. The optimization problem whose solution is the leading $q$ eigenvectors of $AA^\top$ is described in eqn.~\ref{eq:1}:
$$\max_{U\in R^{n\times q}} trace(U^\top AA^\top U)\ \ \ \ \ subject\ to\ \ \ \ U^\top U=I.$$

\section{The Case $n>>m$}

Consider the situation where $n$, the dimension of the input vectors,
is relatively large compared to the number of sample vectors $m$. For
example, consider input vectors representing $50\times 50$ sized
images of faces, i.e., $n=2500$, where $m=100$. In other words, we are
looking for a small number of ``face templates'' (known as
``eigenfaces'') which approximate well the original set of 100 face
images. In this case, $AA^\top$ is very large, $2500\times 2500$, yet
the number of non-vanishing eigenvalues cannot be higher than
100. Given that the eigendecomposition process is $O(2500^3)$, the
computational burden would be very high. However, it is possible to
perform an eigendecomposition on $A^\top A$ (a $100\times 100$ matrix)
instead, as shown next.

Let the columns of $Q$ be the first $q< m$ eigenvectors of $A^\top
A$, i.e., $A^\top AQ=QD$ where $D$ is diagonal containing the
corresponding eigenvalues. After pre-multiplying both sides by $A$ we
obtain:
$$AA^\top(AQ)=(AQ)D,$$
from which we conclude that $AQ$ contains the first $q$ eigenvectors
(but un-normalized) of $AA^\top$. We have therefore that
$U=AQD^{-\frac{1}{2}}$ because:
$$U^\top U = D^{-\frac{1}{2}}Q^\top A^\top A Q D^{-\frac{1}{2}} =D^{-\frac{1}{2}}DD^{-\frac{1}{2}}=I,$$
where we used the fact that $Q^\top A^\top AQ=D$. Note that eigenvalues of $A^\top A$ and $AA^\top$ are the
same (because $AA^\top(AQD^{-\frac{1}{2}})=(AQD^{-\frac{1}{2}})D$).

\section{Kernel PCA}

We can take the case $n>>m$ described in the previous section one step
further and consider such large values of $n$ which are practically
uncomputable --- a situation which results when mapping the original
input vectors to a high dimensional space: $\phi(\bfx)$ where
$\phi:R^n \rightarrow {\cal F}$ for which $dim({\cal F})>>n$. For example, $\phi(\bfx)$
representing the d'th order monomials of the coordinates of $\bfx$,
i.e., $dim({\cal F})={{n+d-1} \choose d}$ which is exponential in $d$. The mappings
of interest are those which are paired with a non-linear kernel function:
$k(\bfx,\bfx')=\phi(\bfx)^\top\phi(\bfx')$.

Performing PCA on $A=[\phi(\bfx_1),...,\phi(\bfx_m)]$ is equivalent to
finding the non-linear surface in $R^n$ (the nature  of
the non-linearity depends on the choice of $\phi()$) which best
approximates the original sample data $\bfx_1,...,\bfx_m$. The problem
is that $AA^\top$ is not computable --- however $A^\top A$ is
computable because $(A^\top A)_{ij}= k(\bfx_i,\bfx_j)$. 

From the previous section, $U=AQD^{-\frac{1}{2}}=AV$ contains the
first $q$ eigenvectors of $AA^\top$(where $Q$ and $D$ are
computable). Since $A$ 
itself is not computable we cannot represent $U$ explicitly, but we
can project a new vector $\phi(\bfx)$ onto the principal directions
$\bfu_1,...,\bfu_q$ and obtain the principal components, i.e., the output vector $\bfy=U^\top
\phi(\bfx)$, as follows. 
$$\bfy=U^\top\phi(\bfx)=V^\top A^\top\phi(\bfx)=V^\top\left(\begin{array}{c} k(\bfx_1,\bfx)\\.\\.\\.\\ k(\bfx_m,\bfx)\end{array}\right).$$
Given the principal components (entries of $\bfy=U^\top\phi(\bfx)$ of $\phi(\bfx)$) we can measure, for example, the {\it distance\/} between $\phi(\bfx)$ and the projection $\hat{\phi(\bfx)}=UU^\top\phi(\bfx)=U\bfy$ onto the linear subspace spanned by $\bfu_1,...,\bfu_q$ (without the need to explicitly compute the principal axes $\bfu_i$), as follows.   
\begin{eqnarray*}
\|\phi(\bfx) - \hat{\phi(\bfx)}\|^2 &=& \phi(\bfx)^\top\phi(\bfx) +\hat{\phi(\bfx)}^\top \hat{\phi(\bfx)} - 2\phi(\bfx)^\top\hat{\phi(\bfx)}\\
&=& k(\bfx,\bfx) +\bfy^\top U^\top U\bfy - 2\phi(\bfx)^\top(UU^\top\phi(\bfx))\\
&=& k(\bfx,\bfx) -\bfy^\top\bfy -2\bfy^\top\bfy\\
&=& k(\bfx,\bfx) - \|\bfy\|^2
\end{eqnarray*}

\section{Fisher's LDA: Basic Idea}

We now extend the variance preserving approach for data representation for {\it labeled\/} data sets. We will focus on 2-class sets and look for a separating hyperplane:
$$f(\bfx) = \bfw^\top\bfx +b,$$
such that $\bfx$ belongs to the first class if $f(\bfx)>0$ and $\bfx$ belongs to the second class if $f(\bfx)<0$. In the statistical literature this type of function is called a {\it linear discriminant function}. The decision boundary is given by the set of points satisfying $f(\bfx)=0$ which is a hyperplane. Fisher's (1936) Linear Discriminant Analysis (LDA) is a variance preserving approach for finding a linear discriminant function. 

We will then introduce another popular statistical technique called Canonical Correlation Analysis (CCA) for learning the mapping between input and output vectors using the notion "angle" between subspaces.

What is common in the three techniques PCA, LDA and CCA is the use of spectral matrix analysis --- i.e., what can you do with eigenvalues and eigenvectors of matrices representing subspaces of the data? These techniques produce optimal results for normally distributed data and are very easy to implement. There is a large variety of uses of spectral analysis in statistical and learning literature including spectral clustering, Multi Dimensional Scaling (MDS) and data modeling in general.

\begin{figure}[t]
\centerline{\epsfig{figure=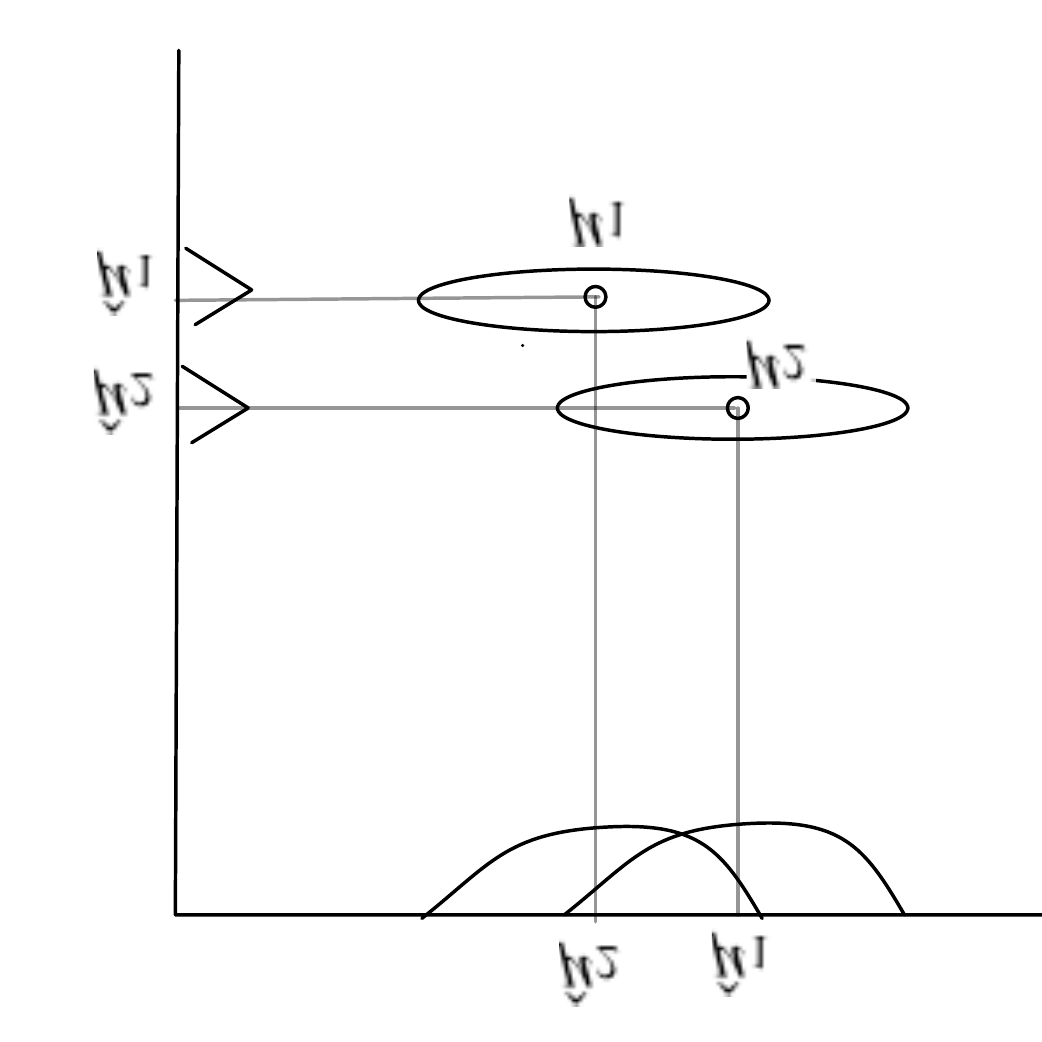,height=6cm}}
\caption{\protect\small Linear discriminant analysis based on class centers alone is not sufficient. Seeking a projection which maximizes the distance between the projected centers will prefer the horizontal axis over the vertical, yet the two classes overlap on the horizontal axis. The projected distance along the vertical axis is smaller yet the classes are better separated. The conclusion is that the sample variance of the two classes must be taken into consideration as well.}
\label{fig:one}
\end{figure}

To appreciate the general idea behind Fisher's LDA consider Fig.~\ref{fig:one}. Let the centers of classes one and two be denoted by $\mu_1$ and $\mu_2$ respectively. A linear discriminant function is a projection onto a 1D subspace such that the classes would be separated the most in the 1D subspace. The obvious first step in this kind of analysis is to make sure that the projected centers $\hat\mu_1,\hat\mu_2$ would be separated as much as possible. We can easily see that the direction of the 1D subspace should be proportional to $\mu_1-\mu_2$ as follows:
$$(\hat\mu_1-\hat\mu_2)^2 = \left(\frac{\bfw^\top\mu_1}{\|\bfw\|} - \frac{\bfw^\top\mu_2}{\|\bfw\|}\right)^2 = \left(\frac{\bfw^\top}{\|\bfw\|}(\mu_1-\mu_2)\right)^2.$$
The right-hand term is maximized when $\bfw\approx \mu_1-\mu_2$. As illustrated in Fig.~\ref{fig:one}, this type of consideration is not sufficient to capture separability in the projected subspace because the spread (variance) of the data points around their centers also play an important role. For example, the horizontal axis in the figure separates the centers better than the vertical axis but on the other hand does a worse job in separating the classes themselves because of the way the data points are spread around their centers. The argument in favor of separating the centers would work if the data points were living in a hyper-sphere around the centers, but will not be sufficient otherwise.

The basic idea behind Fisher's LDA is to consider the sample covariance matrix of the individual classes as well as their centers, in the following way. The optimal 1D projection would that which maximizes the variance of the projected centers while {\it minimizes\/} the variance of the projected data points of each class separately. Mathematically, this idea can be implemented by maximizes the following ratio:
$$\max_{\bfw} \frac{(\hat\mu_1-\hat\mu_2)^2}{s_1^2 + s_2^2},$$
where $s_1^2$ is the scaled variance of the projected points of the first class:
$$s_1^2 = \sum_{\bfx_i\in C_1} (\hat{\bfx_i} - \hat\mu_1)^2,$$
and likewise,
$$s_2^2 = \sum_{\bfx_i\in C_2} (\hat{\bfx_i} - \hat\mu_2)^2,$$
where $\hat{\bfx} = \frac{\bfw^\top}{\|\bfw\|}\bfx_i + b$.

We will now formalize this approach and derive its solution. We will begin with a general description of a multiclass problem where the sample data points belong to $q$ different classes, and later focus on the case of $q=2$.

\section{Fisher's LDA: General Derivation}

Let the sample data points $S$ be members of $q$ classes $C_1,...,C_q$ where the number of points belonging to class $C_i$ is denoted by $l_i$ and the total number of the training set is $l=\sum_i l_i$. Let $\mu_j$ denote the center of class $C_i$ and $\mu$ denote the center of the complete training set $S$:
\begin{eqnarray*}
\mu_j &=& \frac{1}{l_j}\sum_{bfx_i\in C_j}\bfx_i\\
\mu &=& \frac{1}{l}\sum_{\bfx_i\in S}\bfx_i
\end{eqnarray*}
Let $A_j$ be the matrix associated with class $C_j$ whose columns consists of the mean shifted data points:
$$A_j = [\bfx_1-\mu_j,...,\bfx_{l_j}-\mu_j]\ \ \ \ \bfx_i\in C_j.$$
Then, $\frac{1}{l_j}A_jA_j^\top$ is the covariance matrix  associated with class $C_j$. Let $S_w$ (where "w" stands for "within") be the sum of the class covariance matrices:
$$S_w = \sum_i^q \frac{1}{l_j}A_jA_j^\top.$$
From the discussion in the previous section, it is $\frac{1}{\|\bfw\|^2}\bfw^\top S_w\bfw$ which we wish to minimize. To see why this is so, note
$$\sum_{\bfx_i\in C_j}(\hat{\bfx_i} - \hat\mu_j)^2= \sum_{\bfx_i\in C_j}\frac{\bfw^\top(\bfx_i - \mu_j)^2}{\|\bfw\|^2}=\frac{1}{\|\bfw\|^2} \bfw^\top A_jA_j^\top\bfw.$$
Let $B$ be the matrix holding the class centers:
$$B=[\mu_1-\mu,...,\mu_q-\mu],$$
and let $S_b=\frac{1}{q}BB^\top$  (where "b" stands for "between"). From the discussion above it is $\frac{1}{\|\bfw\|^2}\bfw^\top S_b\bfw = \sum_i (\hat\mu_i - \hat\mu)^2$ which we wish to {\it maximize}. Taken together, we wish to maximize the ratio (called "Rayleigh's quotient"):
$$\max_{\bfw} J(\bfw)=\frac{\bfw^\top S_b\bfw}{\bfw^\top S_w\bfw}.$$
The necessary condition for optimality is:
$$\frac{\partial J}{\partial \bfw}=\frac{S_b\bfw (\bfw^\top S_w\bfw)-S_w\bfw (\bfw^\top S_b\bfw)}{ (\bfw^\top S_w\bfw)^2}=0,$$
From which we obtain the generalized eigensystem:
\be S_b\bfw = J(\bfw) S_w\bfw.\label{eq:w}\ee
That is, $\bfw$ is the leading eigenvector of $S_w^{-1}S_b$ (assuming $S_w$ is invertible). The general case of finding $q$ such axes involves finding the leading generalized eigenvectors of $(S_b,S_w)$ --- the derivation is out of scope of this lecture. Note that since $S_w^{-1}S_b$ is not symmetric there may be no real-value solution, which is a complication will not pursue further in this course. Instead we will focus now on the 2-class ($q=2$) setting below.

\section{Fisher's LDA: 2-class}

The general derivation is simplified when there are only two classes. The covariance matrix $BB^\top$ becomes a rank-1 matrix:
$$BB^\top = (\mu_1 - \mu)(\mu_1-\mu)^\top + (\mu_2 - \mu)(\mu_2-\mu)^\top = (\mu_1-\mu_2)(\mu_1-\mu_2)^\top.$$
As a result, $BB^\top\bfw$ is a vector in direction $\mu_1-\mu_2$. Therefore, the solution for $\bfw$ from eqn.~\ref{eq:w} is:
$$\bfw \cong S_w^{-1}(\mu_1-\mu_2).$$
The decision boundary $\bfw^\top(\bfx-\mu)=0$ becomes:
\be
\bfx^\top S_w^{-1}(\mu_1-\mu_2) - \frac{1}{2} (\mu_1+\mu_2)^\top S_w^{-1} (\mu_1-\mu_2) = 0.
\label{eq:fisher}
\ee
This decision boundary will surface again in the course when we consider Bayseian inference. It will be shown that this decision boundary is the Maximum Likelihood solution in the case where the two classes are normally distributed with means $\mu_1,\mu_2$ and with the same covariance matrix $S_w$.

\section{LDA versus SVM}

Both LDA and SVM search for a so called "optimal" linear discriminant function, what is the difference? The heart of the matter lies in the definition of what constitutes a sufficient compact representation of the data. In LDA the assumption is that each class can be represented by its mean vector and its spread (i.e., covariance matrix). This is true for normally distributed data --- but not true in general. This means that we should expect that LDA will produce the optimal discriminant linear function when each of the classes are normally distributed.

With SVM, on the other hand, there is no assumption on how the data is distributed. Instead, the emerging result is that the data is represented by the subset of data points which lie on the boundary between the two classes (the so called support vectors). Rather than making a parametric assumption on how the data can be captured (i.e., mean and covariance) the theory shows that the data can be captured by a special subset of points. The tools, as a result, are naturally more complex (quadratic linear programming versus spectral matrix analysis) --- but the advantage is that optimality is guaranteed without making assumptions on the distribution of the data (i.e., distribution free). It can be shown that SVM and LDA would produce the same result if the class data is normally distributed.  

\section{Canonical Correlation Analysis}

CCA is a technique for learning a mapping $f(\bfx)=\bfy$ where $\bfx\in R^k$ and $\bfy\in R^s$ using the notion of subspace similarity (an extension of the inner product between two vectors) from a training set of $(\bfx_i,\bfy_i)$, $i=1,...,n$. Such a mapping, where $\bfy$ can be any point in $R^k$ as opposed to a discrete set of labels, is often referred to as a "regression" (as opposed to "classification"). 

Like in PCA and LDA, the approach would be to look for projection axes such that the projection of the input and output vectors on those axes satisfy certain requirements --- and like PCA and LDA the tools we would be using is matrix spectral analysis. 

It will be convenient to stack our vectors as rows of an input matrix $A$ and output matrix $B$. Let $A$ be an $n\times k$ matrix whose rows are $\bfx_1^\top,...,\bfx_n^\top$ and $B$ is the $n\times s$ matrix whose rows are $\bfy_1^\top,...,\bfy_n^\top$. 
Consider vectors $\bfu\in R^k$ and $\bfv\in R^s$ and project the input and output data onto them producing $A\bfu = (\bfx_1^\top\bfu,...,\bfx_n^\top\bfu)$ and $B\bfv$. The requirement we would like to place on the projection axes is that $A\bfu \approx B\bfv$, or in other words that $(A\bfu)^\top (B\bfv)$ is maximal. The requirement therefore is that the projection of the input points onto the $\bfu$ axis is similar to the projection of the output points onto the $\bfv$ axis. If we extend this notion to multiple axes $\bfu_1,...,\bfu_q$ (not necessarily orthogonal) and $\bfv_1,...,\bfv_q$ where $q\le \min(k,s)$ our requirement becomes that the new coordinates of the input points projected onto the subspace spanned by the $\bfu$ vectors are {\it similar\/} to the new coordinates of the output points projected onto the subspace spanned by the $\bfv$ vectors. In other words, we wish to find two $q$-dimensional subspaces one of $R^k$ and the other of $R^s$ such that the two sets of projected points are as aligned as possible.

CCA goes a step further and makes the assumption that the input/output relationship is solely determined by the relation (angles) between the column spaces of $A,B$. In other words, the particular columns of $A$ are not really important, what is important is the space $U_A$ spanned by the columns. Since $\bfg=A\bfu$ is a point in $U_A$ (a linear combination of the columns of $A$) and $\bfh=B\bfv$ is a point in $U_B$, then $\bfg^\top\bfh$ is the cosine angle, $\cos(\phi)$ between the two axes provided that we normalize the vectors $\bfg$ and $\bfh$. If we continue this line of reasoning recursively, we obtain  a set of angles 
$0\le
\theta_1\le ...\le\theta_q\le(\pi/2)$, called "principal angles",  between the two subspaces
uniquely defined as:
\be
cos(\theta_j)=\max_{{\bfg}\in U_A}\max_{{\bfh}\in
U_B}\bfg^\top\bfh\label{eq:uv}
\ee
subject to:
$$
\bfg^\top\bfg=\bfh^\top\bfh=1,\ \ \
\bfh^\top\bfh_i=0,\bfg^\top\bfg_i=0,\ \ \ \ i=1,...,j-1
$$
As a result, we obtain the following optimization function over axes $\bfu,\bfv$:
$$\max_{\bfu,\bfv} \bfu^\top A^\top B\bfv\ \ \ {\rm s.t.}\ \ \ \|A\bfu\|^2=1,\ \ \|B\bfv\|^2=1.$$
To solve this problem we first perform a "QR" factorization of $A$ and $B$. A "QR" factorization of a matrix $A$ is a Grahm-Schmidt process resulting in an orthonormal set of vectors arranged as the columns of a matrix $Q_A$ whose column space is equal to the column space of $A$, and a matrix $R_A$ which contains the coefficients of the linear combination of the columns of $Q_A$ such that $A=Q_AR_A$. Since orthoganilzation is not unique, the Grahm-Schmidt process perfroms the orthogonalization such that $R_A$ is an upper-diagonal matrix. Likewise let $B=Q_BR_B$.
Because the column spaces of $A$ and $Q_A$ are the same, then for every $\bfu$ there exists a $\hat{\bfu}$ such that $A\bfu = Q_A\hat{\bfu}$. Our optimization problem now becomes:
$$\max_{\hat{\bfu},\hat{\bfv}} {\hat{\bfu}}^\top Q_A^\top Q_B{\hat{\bfv}}\ \ \ {\rm s.t.}\ \ \ \|\hat{\bfu}\|^2=1,\ \ \|\hat{\bfv}\|^2=1.$$
The solution of this problem is when $\hat{\bfu}$ and $\hat{\bfv}$ are the leading singular vectors of $Q_A^\top Q_B$. The singular value decomposition (SVD) of any matrix $E$ is a decomposition $E=UDV^\top$ where the columns of $U$ are the leading eigenvectors of $EE^\top$, the rows of $V^\top$ are the leading eigenvectors of $E^\top E$ and $D$ is a diagonal matrix whose entries are the corresponding square eigenvalues (note that the eigenvalues of $EE^\top$ and $E^\top E$ are the same). The SVD decomposition has the property that if we keep only the first $q$ leading eigenvectors then $UDV^\top$ is the closest (in least squares sense) rank $q$ matrix to $E$.

Therefore, let $\hat U D \hat V^\top$ be the SVD of $Q_A^\top Q_B$ using the first $q$ eigenvectors. Then, our sought after axes $U=[\bfu_1,...,\bfu_q]$ is simply $R_A^{-1}\hat U$ and likewise and the axes $V=[\bfv_1,...,\bfv_q]$ is equal to $R_B^{-1}\hat V$. The axes are called "canonical vectors", and the vectors $\bfg_i=A\bfu_i$ (mutually orthogonal) are called "variates". The concept of principal angles is due to Jordan in 1875, where
Hotelling in 1936 is the first to introduce the recursive definition
above.

Given a new vector $\bfx\in R^k$ the resulting vector $\bfy$ can be found by solving the linear system $U^\top\bfx = V^\top\bfy$ (since our assumption is that in the new basis the coordinates of $\bfx$ and $\bfy$ are similar).

To conclude, the relationship between $A$ and $B$ is captured by creating similar variates, i.e., creating subspaces of dimension $q$ such that the projections of the input vectors and the output vectors have similar coordinates. The process for obtaining the two $q$-dimensional subspaces is by performing a QR factorization of $A$ and $B$ followed by an SVD. Here again the spectral analysis of the input and output data matrices plays a pivoting role in the input/output association.

\chapter{Spectral Analysis II: Clustering}

In the previous lecture we ended up with the formulation:
\be
\max _{G_{m\times k}} trace(G^\top K G)\ \ \ {\rm s.t.\ } G^\top G=I
\label{eq:spectral}
\ee
and showed the solution $G$ is the leading eigenvectors of the symmetric positive semi definite matrix $K$. When $K=AA^\top$ (sample covariance matrix) with $A=[\bfx_1,...,\bfx_m]$, $\bfx_i\in R^n$, those eigenvectors form a basis to a $k$-dimensional subspace of $R^n$ which is the closest (in $L_2$ norm sense) to the sample points $\bfx_i$. The axes (called principal axes) $\bfg_1,...,\bfg_k$ preserve the variance of the original data in the sense that the projection of the data points on the $\bfg_1$ has maximum variance, projection on $\bfg_2$ has the maximum variance over all vectors orthogonal to $\bfg_1$, etc. The spectral decomposition of the sample covariance matrix is a way to "compress" the data by means of linear super-position of the original coordinates $\bfy = G^\top \bfx$.

We also ended with a ratio formulation:
$$\max_{\bfw} \frac{\bfw^\top S_1\bfw}{\bfw^\top S_2\bfw}$$
where $S_1,S_2$ where scatter matrices defined such that $\bfw^\top S_1\bfw$ is the variance of class centers (which we wish to maximize) and $\bfw^\top S_2\bfw$ is the sum of within class variance (which we want to minimize). The solution $\bfw$ is the generalized eigenvector $S_1\bfw = \lambda S_2\bfw$ with maximal $\lambda$.

In this lecture we will show additional applications where the search for leading eigenvectors plays a pivotal part of the solution. So far we have seen how spectral analysis relates to PCA and LDA and today we will focus on the classic Data Clustering problem of partitioning a set of points $\bfx_1,...,\bfx_m$ into $k\ge 2$ classes, i.e., generating as output indicator variables $y_1,...,y_m$ where $y_i\in\{1,...,k\}$. We will begin with "K-means" algorithm for clustering and then move on to show how the optimization criteria relates to grapth-theoretic approaches (like Min-Cut, Ratio-Cut, Normalized Cuts) and spectral decomposition. 

\section{K-means Algorithm for Clustering}

The K-means formulation (originally introduced by \cite{Linde80}) assumes that the clusters are defined by the distance of the points to their class centers only. 
In other words, the goal of clustering is to find those $k$ mean vectors $\bfc_1,...,\bfc_k$ and provide the cluster assignment $y_i\in \{1,...,k\}$ of each point $\bfx_i$ in the set. The K-means algorithm is based on an interleaving approach where the cluster assignments $y_i$ are established given the centers and the centers are computed given the assignments. The optimization criterion is as follows:
\be
\min_{y_1,...,y_m, \bfc_1,...,\bfc_k} \sum_{j=1}^k \sum_{y_i=j} \|\bfx_i-\bfc_j\|^2
\label{eq:kmeans}
\ee
Assume that $\bfc_1,...,\bfc_k$ are given from the previous iteration, then
$$y_i = \argmin{j} \|\bfx_i-\bfc_j\|^2,$$
and next assume that $y_1,..,y_m$ (cluster assignments) are given, then for any set $S\subseteq \{1,...,m\}$ we have that
$$\frac{1}{|S|}\sum_{j\in S} \bfx_j = \argmin{\bfc} \sum_{j\in S}\|\bfx_j-\bfc\|^2.$$
In other words, given the estimated centers in the current round, the new assignments are computed by the closest center to each point $\bfx_i$, and then given the updated assignments the new centers are estimated by taking the mean of each cluster. Since each step is guaranteed to reduce the optimization energy the process must converge --- to some local optimum.

The drawback of the K-means algorithm is that the quality of the local optimum strongly depends on the initial guess (either the centers or the assignments). If we start with a wild guess for the centers it would be fairly unlikely that the process would converge to a good local minimum (i.e. one that is close to the global optimum). An alternative approach would be to define an approximate but simpler problem which has a closed form solution (such as obtained by computing eigenvectors of some matrix). The global optimum of the K-means is an NP-Complete problem (mentioned briefly in the next section).

Next, we will rewrite the K-means optimization criterion in matrix form and see that it relates to the spectral formulation (eqn.~\ref{eq:spectral}).

\subsection{Matrix Formulation of K-means}

We rewrite eqn.~\ref{eq:kmeans} as follows \cite{cp-iccv05}. Instead of carrying the class variables $y_i$ we define class sets $\psi_1,...,\psi_k$ where $\psi_i\subset \{1,...,n\}$ with $\bigcup \psi_j = \{1,...,n\}$ and $\psi_i \bigcap \psi_j = \emptyset$. The K-means optimization criterion seeks for the centers and the class sets:
$$\min_{\psi_1,...,\psi_k, \bfc_1,...,\bfc_k} \sum_{j=1}^k \sum_{i\in\psi_j} \|\bfx_i-\bfc_j\|^2.$$
Let $l_j = |\psi_j|$ and following the expansion of the squared norm and dropping $\bfx_i^\top \bfx_i$ we end up with an equivalent problem:
$$\min_{\psi_1,...,\psi_k, \bfc_1,...,\bfc_k} \sum_{j=1}^k l_j\bfc_j^\top\bfc_j - 2 \sum_{j=1}^k\sum_{i\in\psi_j}\bfx_i^\top\bfc_j.$$
Next we substitute $\bfc_j$ with its definition: $(1/l_j)\sum_{i\in \psi_j}\bfx_j$ and obtain a new equivalent formulation where the centers $\bfc_j$ are eliminated form consideration:
$$\min_{\psi_1,...,\psi_k} -  \sum_{j=1}^k\frac{1}{l_j} \sum_{r,s\in\psi_j} \bfx_r^\top\bfx_s$$
which is more conveniently written as a maximization problem:
\be
\max_{\psi_1,...,\psi_k} \sum_{j=1}^k\frac{1}{l_j} \sum_{r,s\in\psi_j} \bfx_r^\top\bfx_s.
\label{eq:kmeans-max}
\ee
Since the resulting formulation involves only inner-products we could have replaced $\bfx_i$ with $\phi(\bfx_i)$ in eqn.~\ref{eq:kmeans} where the mapping $\phi(\cdot)$ is chosen such that $\phi(\bfx_i)^\top \phi(\bfx_j)$ can be replaced by some non-linear function $\kappa(\bfx_i,\bfx_j)$ --- known as the "kernel trick" (discussed in previous lectures). Having the ability to map the input vectors onto some high-dimensional space before K-means is applied provides more flexibility and increases our chances of getting out a "good" clustering from the global K-means solution (again, the local optimum depends on the initial conditions so it could be "bad"). The RBF kernel is quite popular in this context $\kappa(\bfx_i,\bfx_j)=e^{-\| \bfx_i-\bfx_j\|^2/\sigma^2}$ with $\sigma$ some pre-determined parameter. Note that $\kappa(\bfx_i,\bfx_j)\in (0,1]$ which can be interpreted loosely as the probability of $\bfx_i$ and $\bfx_j$ to be clustered together.

Let $K_{ij} = \kappa(\bfx_i,\bfx_j)$ making $K$ a $m\times m$ symmetric positive-semi-definite matrix often referred to as the "affinity" matrix. Let $F$ be an $n\times n$ matrix whose entries are $F_{ij}=1/l_r$ if $(i,j)\in \psi_r$ for some class $\psi_r$ and $F_{ij}=0$ otherwise. In other words, if we sort the points $\bfx_i$ according to cluster membership, then $F$ is a block diagonal matrix with blocks $F_1,...,F_k$ where $F_r = (1/l_r)\bfone\bfone^\top$ is an $l_r\times l_r$ block of 1's scaled by $1/l_r$. Then, 
Eqn.~\ref{eq:kmeans-max} can be written in terms of $K$ as follows:
\be
\max_{F} \sum_{i,j=1}^n K_{ij}F_{ij} = trace(KF)
\label{eq:tr1}
\ee
In order to form this as an optimization problem we need to represent the structure of $F$ in terms of constraints. Let $G$ be an $n\times k$ column-scaled indicator matrix: $G_{ij}=(1/\sqrt{l_j})$ if $i\in\psi_j$ (i.e., $\bfx_i$ belongs to the j'th class) and $G_{ij}=0$ otherwise. Let $\bfg_1,...,\bfg_k$ be the columns of $G$ and it can be easily verified that ${\bfg_r\bfg_r}^\top = diag(0,..,F_r,0,..,0)$ therefore $F=\sum_j \bfg_j\bfg_j^\top = GG^\top$. Since $trace(AB)=trace(BA)$ we can now write eqn.~\ref{eq:tr1} in terms of $G$:
$$\max_{G} trace(G^\top KG)$$
under conditions on $G$ which we need to further spell out. 

We will start with the necessary conditions. Clearly $G\ge 0$ (has non-negative entries). Because each point belongs to exactly one cluster we must have $G^\top G_{ij}=0$ when $i\not = j$ and $G^\top G_{ii} = (1/l_i)\bfone^\top \bfone = 1$, thus $G^\top G=I$. 
Furthermore we have that the rows and columns of $F=GG^\top$ sum up to 1, i.e., $F\bfone = \bfone, F^\top\bfone=\bfone$ which means that $F$ is {\it doubly stochastic} which translates to the constraint $GG^\top\bfone=\bfone$ on $G$. We have therefore three necessary conditions on $G$: (i) $G\ge 0$, (ii) $G^\top G=I$, and (iii) $GG^\top\bfone=\bfone$. The claim below asserts that these are also sufficient conditions:
\begin{claim}
The feasibility set of matrices $G$ which satisfy the three conditions $G\ge 0,\ GG^\top\bfone=\bfone$ and $G^\top G=I$  are of the form:
$$G_{ij}=\left\{\begin{array}{cc} \frac{1}{\sqrt{l_j}}& \bfx_i\in\psi_j\\ 0 & otherwise\end{array}\right\}$$
\end{claim}
{\bf Proof:\ }
From $G\ge 0$ and $\bfg_r^\top\bfg_s=0$ we have that $G_{ir}G_{is}=0$, i.e., $G$ has a single non-vanishing element in each row. It will be convenient to assume that the points are sorted according to the class membership, thus the columns of $G$ have the non-vanishing entries in consecutive order and let $l_j$ be the number of non-vanishing entries in column $\bfg_j$. Let $\bfu_j$ the vector of $l_j$ entries holding only the non-vanishing entries of $\bfg_j$. Then, 
the doubly stochastic constraint $GG^\top\bfone=\bfone$ results that $(\bfone^\top \bfu_j)\bfu_j = \bfone$ for $j=1,...,k$. Multiplying $\bfone$ from both sides yields $(\bfone^\top \bfu_j)^2=\bfone^\top \bfone=l_j$, therefore $\bfu_j = (1/\sqrt{l_j})\bfone$. \eop

This completes the {\it equivalence\/} between the matrix formulation:
\be
\max_{G\in R^{m\times k}} trace(G^\top K G)\ \ \ {\rm s.t.\ } G\ge 0, G^\top G=I, GG^\top\bfone =\bfone
\label{eq:tr2}
\ee
and the original K-means formulation of eqn.~\ref{eq:kmeans}.

We have obtained the same optimization criteria as eqn.~\ref{eq:spectral} with additional two constraints: $G$ should be non-negative and $GG^\top$ should be doubly stochastic. The constraint $G^\top G=I$ comes from the requirement that each point is assigned to one class only. The doubly stochastic constraint comes from a "class balancing" requirement which we will expand on below. 

\section{Min-Cut}

We will arrive to eqn.~\ref{eq:tr2} from a graph-theoretic perspective. We start with representing the graph Min-Cut problem in matrix form, as follows. A convenient way to represent the data to be clustered is by an undirected graph with edge-weights where $V=\{1,...,m\}$ is the vertex set, $E\subset V\times V$ is the edge set and $\kappa\ :E\ \rightarrow R_+$ is the positive weight function. Vertices of the graph correspond to data points $\bfx_i$, edges represent neighborhood relationships, and edge-weights represent the similarity (affinity) between pairs of linked vertices. The weight adjacency matrix $K$ holds the weights where $K_{ij} = \kappa(i,j)$ for $(i,j)\in E$ and $K_{ij}=0$ otherwise.

A {\it cut\/} in the graph is defined between two disjoint sets $A,B\subset V$, $A\cup B=V$, is the sum of edge-weights connecting the two sets: $cut(A,B)=\sum_{i\in A, j\in B} K_{ij}$ which is a measure of dissimilarity between the two sets. The Min-Cut problem is to find a minimal weight cut in the graph (can be solved in polynomial time through Max Network Flow solution). The following claim associates algebraic conditions on $G$ with an indicator matrix:
\begin{claim}
The feasibility set of matrices $G$ which satisfy the three conditions $G\ge 0,\ G\bfone=\bfone$ and $G^\top G=D$ for some diagonal matrix $D$ are of the form:
$$G_{ij}=\left\{\begin{array}{cc} 1 & x_i\in\psi_j\\ 0 & otherwise\end{array}\right\}$$
\end{claim}
{\bf Proof:\ }
Let $G=[\bfg_1,...,\bfg_k]$. From $G\ge 0$ and $\bfg_r^\top\bfg_s=0$ we have that $G_{ir}G_{is}=0$, i.e., $G$ has a single non-vanishing element in each row. From $G\bfone=\bfone$ the single non-vanishing entry of each row must have the value of $1$.
\eop

In the case of two classes ($k=2$), the function $tr(G^\top KG)$ is equal to $\sum_{(i,j)\in\psi_1} K_{ij} + \sum_{(i,j)\in\psi_2} K_{ij}$. Therefore $\max_G tr(G^\top KG)$ is equivalent to minimizing the cut: $\sum_{i\in\psi_1, j\in\psi_2} K_{ij}$. As a result, the Min-Cut problem is equivalent to solving the optimization problem:
\be
\max_{G\in R^{m\times 2}} tr(G^\top KG)\ s.t\ \ \ G\ge 0,\ G\bfone=\bfone,\ G^\top G=diag\label{eq:mincut}
\ee

We seem to be close to eqn.~\ref{eq:tr2} with the difference that $G$ is orthogonal (instead of orthonormal) and the doubly-stochasitc constraint is replaced by $G\bfone=\bfone$. The difference can be bridged by considering a "balancing" requirement. Min-Cut can produce an unbalanced partition where one set of vertices is very large and the other contains a spurious set of vertices having a small number of edges to the larger set. This is an undesirable outcome in the context of clustering. Consider a "balancing" constraint $G^\top\bfone = (m/k)\bfone$ which makes a strict requirement that all the $k$ clusters have an equal number of points. We can relax the balancing constraint slightly by combining the balancing constraint with $G\bfone=\bfone$ into one single constraint $GG^\top\bfone=(m/k)\bfone$, i.e., $GG^\top$ is scaled doubly stochastic. Note that the two conditions $GG^\top\bfone=(m/k)\bfone$ and $G^\top G=D$ result in $D=(m/k)I$. Thus we propose the relaxed-balanced hard clustering scheme:
$$\max_G tr(G^\top KG)\ s.t\ \ \ G\ge 0,\ GG^\top\bfone=\frac{m}{k}\bfone,\ G^\top G=\frac{m}{k}I$$
The scale $m/k$ is a global scale that can be dropped without affecting the resulting solution, thus the Min-Cut with a relaxed balancing requirement becomes eqn.~\ref{eq:tr2} which we saw is equivalent to K-means:
$$\max_G tr(G^\top KG)\ s.t\ \ \ G\ge 0,\ GG^\top\bfone=\bfone,\ G^\top G=I.$$

\section{Spectral Clustering: Ratio-Cuts and Normalized-Cuts}

We saw above that the doubly-stochastic constraint has to do with a "balancing" desire. A further relaxation of the balancing desire is to perform the optimization in two steps: (i) replace the affinity matrix $K$ with the closest (under some chosen error measure) doubly-stochastic matrix $K'$, (ii) find a solution to the problem:
\be
\max_{G\in R^{m\times k}} tr(G^\top K'G)\ s.t\ \ \ G\ge 0,\ G^\top G=I\label{eq:tr3}
\ee
because $GG^\top$ should come out close to $K'$ ($tr(G^\top K'G)=tr(K'GG^\top)$) and $K'$ is doubly-stochastic, then $GG^\top$ should come out close to satisfying a doubly-stochastic constraint --- this is the motivation behind the 2-step approach. Moreover, we drop the non-negativity constraint $G\ge 0$. Note that the non-negativity constraint is crucial for the physical interpretation of $G$; nevertheless, for $k=2$ clusters it is possible to make an interpretation, as we shall next. As a result we are left with a spectral decomposition problem of eqn.~\ref{eq:spectral}:
$$\max_{G\in R^{m\times k}} tr(G^\top K'G)\ s.t\ \ \ G^\top G=I,$$
where the columns of $G$ are the leading eigenvectors of $K'$. We will refer to the first step as a "normalization" process and there are two popular normalizations in the literature --- one leading to Ratio-Cuts and the other to Normalized-Cuts.

\subsection{Ratio-Cuts}

Let $D=diag(K\bfone)$ which is a diagonal matrix containing the row sums of $K$. The Ratio-Cuts normalization is to look for $K'$ as the closest doubly-stochastic matrix to $K$ by minimizing the $L_1$ norm --- this turns out to be $K'=K-D+I$.
\begin{claim}[ratio-cut]
Let $K$ be a symmetric positive-semi-definite whose values are in the range $[0,1]$. The closest doubly stochastic matrix $K'$ under the $L_1$ error norm is
$$K' = K-D+I$$
\end{claim}
{\bf Proof:\ }
Let $r = \min_F \| K-F\|_1\ \ s.t.\ \ F\bfone=\bfone,\ \ F=F^\top$. Since $\|K-F\|_1 \ge \|(K-F)\bfone\|_1$ for any matrix $F$, we must have:
$$r \ge \|(K-F)\bfone\|_1 = \|D\bfone - \bfone\|_1 = \|D - I\|_1.$$
Let $F=K-D+I$, then
$$\|K - (K-D+I)\|_1 = \|D-I\|_1.$$
\eop

The Laplacian matrix of a graph is $D-K$. 
If $\bfv$ is an eigenvector of the Laplacian $D-K$ with eigenvalue $\lambda$, then $\bfv$ is also an eigenvector of $K'=K-D+I$ with eigenvalue $1-\lambda$ and since $(D-K)\bfone=0$ then the smallest eigenvector $\bfv=\bfone$ of the Laplacian is the largest of $K'$, and the second smallest eigenvector of the Laplacian (the ratio-cut result) corresponds to the second largest eigenvector of $K'$. Because the eigenvectors are orthogonal, the second eigenvector must have positive and negative entries (because the inner-product with $\bfone$ is zero) --- thus the sign of the entries of the second eigenvector determines the class membership.

Ratio-Cuts, the second smallest eigenvector of the Laplacian $D-K$, is an approximation due to Hall in the 70s \cite{Hall70} to the Min-Cut formulation. Let $\bfz\in R^m$ determine the class membership such that $\bfx_i$ and $\bfx_j$ would be clustered together if $z_i$ and $z_j$ have similar values. This leads to the following optimization problem:
$$
\min_{\bfz} \frac{1}{2} \sum_{i,j} (z_i-z_j)^2 K_{ij}\ \ \ s.t.\ \ \  \bfz^\top\bfz=1
$$
The criterion function is equal to $(1/2)\bfz^\top(D-K)\bfz$ and the derivative of the Lagrangian $(1/2)\bfz^\top(D-K)\bfz - \lambda(\bfz^\top\bfz - 1)$ with respect to $\bfz$ gives rise to the necessary condition $(D-K)\bfz = \lambda\bfz$ and the Ratio-Cut scheme follows.

\subsection{Normalized-Cuts}

Normalized-Cuts looks for the closest doubly-stochastic matrix $K'$ in {\it relative entropy\/} error measure defined as: 
$$RE(\bfx\ ||\ \bfy) = \sum_i x_i \ln \frac{x_i}{y_i} + \sum_i y_i - \sum_i x_i.$$
We will encounter the relative entropy measure in more detail later in the course. We can show that $K'$ must have the form $\Lambda K \Lambda$ for some diagonal matrix $\Lambda$:
\begin{claim}
The closest doubly-stochastic matrix $F$ under the relative-entropy error measure to a given non-negative symmetric matrix $K$, i.e., which minimizes:
$$\min_F \; RE(F || K)\ \ s.t.\ \ F\ge 0,\ F=F^\top,\ F\bfone =1,\ F^\top\bfone=1$$
has the form $F=\Lambda K\Lambda$ for some (unique) diagonal matrix $\Lambda$.
\end{claim}
{\bf Proof:\ }The Lagrangian of the problem is:
$$L()=\sum_{ij} f_{ij}\ln\frac{f_{ij}}{k_{ij}} + \sum_{ij}k_{ij} - \sum_{ij}f_{ij} - \sum_i \lambda_i (\sum_j f_{ij}-1) - \sum_j \mu_j (\sum_i f_{ij}-1)$$
The derivative with respect to $f_{ij}$ is:
$$\frac{\partial L}{\partial f_{ij}} = \ln f_{ij} + 1 -\ln k_{ij} -1 - \lambda_i - \mu_j = 0$$
from which we obtain:
$$f_{ij}=e^{\lambda_i}e^{\mu_j}k_{ij}$$
Let $D_1=diag(e^{\lambda_1},...,e^{\lambda_n})$ and $D_2=diag(e^{\mu_1},...,e^{\mu_n})$, then we have:
$$F = D_1KD_2$$
Since $F=F^\top$ and $K$ is symmetric we must have $D_1=D_2$. \eop

Next, we can show that the diagonal matrix $\Lambda$ can found by an iterative process where $K$ is replaced by $D^{-1/2}KD^{-1/2}$  where $D$ was defined above as $diag(K\bfone)$:
\begin{claim}
For any non-negative symmetric matrix $K^{(0)}$, iterating the process $K^{(t+1)}\leftarrow D^{-1/2}K^{(t)}D^{-1/2}$ with $D=diag(K^{(t)}\bfone)$ converges to a doubly stochastic matrix.
\end{claim}
The proof is based on showing that the permanent increases monotonically, i.e. $perm(K^{(t+1)})\ge perm(K^{(t)})$. Because the permanent is bounded the process must converge and if the permanent does not change (at the convergence point) the resulting matrix must be doubly stochastic. The resulting doubly stochastic matrix is the closest to $K$ in relative-entropy. 

Normalized-Cuts takes the result of the first iteration by replacing $K$ with $K' = D^{-1/2}KD^{-1/2}$ followed by the spectral decomposition (in case of $k=2$ classes the partitioning information is found in the second leading eigenvector of $K'$ --- just like Ratio-Cuts but with a different $K'$). Thus, $K'$ in this manner is not the closest doubly-stochastic matrix to $K$ but is fairly close (the first iteration is the dominant one in the process).

Normalized-Cuts, as the second leading eigenvector of $K'=D^{-1/2}KD^{-1/2}$, is an approximation to a "balanced" Min-Cut described first in \cite{Malik-Shi-pami00}. Deriving it from first principles proceeds as follows:

 Let $sum(V_1,V_2) = sum_{i\in V_1, j\in V_2} K_{ij}$ be defined for any two subsets (not necessarily disjoint) of vertices. The normalized-cuts measures the cut cost as a fraction of the total edge connections to all the nodes in the graph:
$$Ncuts (A,B) = \frac{cut(A,B)}{sum(A,V)} + \frac{cut(A,B)}{sum(B,V)}.$$ 
A minimal Ncut partition will no longer favor small isolated points since the cut value would most likely be a large percentage of the total connections from that small set to all the other vertices. A related measure $Nassoc(A,B)$ defined as:
$$Nassoc(A,B) = \frac{sum(A,A)}{sum(A,V)} + \frac{sum(B,B)}{sum(B,V)},$$
reflects how tightly on average nodes within the group are connected to each other. Given that $cut(A,B) = sum(A,V) - sum(A,A)$ one can easily verify that:
$$Ncuts(A,B) = 2 - Nassoc(A,B),$$
therefore the optimal bi-partition can be represented as maximizing $Nassoc(A,V-A)$. The $Nassoc$ naturally extends to $k>2$ classes (partitions) as follows: Let $\psi_1,...,\psi_k$ be disjoint sets $\cup_j \psi_j = V$, then:
$$Nassoc(\psi_1,...,\psi_k) = \sum_{j=1}^k \frac{sum(\psi_j,\psi_j)}{sum(\psi_j, V)}.$$
We will now rewrite $Nassoc$ in matrix form and establish equivalence to eqn.~\ref{eq:tr3}. Let $\bar G = [\bfg_1,...,\bfg_k]$ with $\bfg_j = 1/\sqrt{sum(\psi_j,V)}(0,...,0,1,...1,0.,,,0)$ with the 1s indicating membership to the j'th class.
Note that 
$$\bfg_j^\top K \bfg_j = \frac{sum(\psi_j,\psi_j)}{sum(\psi_j, V)},$$
therefore $trace(\bar G^\top K \bar G) = Nassoc(\psi_1,...,\psi_k)$. Note also that $\bfg_i^\top D \bfg_i = (1/sum(\psi_i,V))\sum_{r\in \psi_i} d_r = 1$, therefore $\bar G^\top D \bar G = I$. Let $G=D^{1/2}\bar G$ so we have that $G^\top G = I$ and $trace(G^\top D^{-1/2}KD^{-1/2} G) = Nassoc(\psi_1,...,\psi_k)$. Taken together we have that maximizing $Nassoc$ is equivalent to:
\be
\max_{G\in R^{m\times k}} trace(G^\top K' G)\ \ \ {\rm s.t.\ } G\ge 0, G^\top G=I,\label{eq:tr4}
\ee
where $K' = D^{-1/2}KD^{-1/2}$. Note that this is exactly the K-means matrix setup of eqn.~\ref{eq:tr2} where the doubly-stochastic constraint is relaxed into the replacement of $K$ by $K'$. The constraint $G\ge 0$ is then dropped and the resulting solution for $G$ is the $k$ leading eigenvectors of $K'$.

We have arrived via seemingly different paths to eqn.~\ref{eq:tr4} which after we drop the constraint $G\ge 0$ we end up with a closed form solution consisting of the $k$ leading eigenvectors of $K'$. When $k=2$ (two classes) one can easily verify that the partitioning information is fully contained in the {\it second\/} eigenvector. Let $\bfv_1,\bfv_2$ be the first leading eigenvectors of $K'$. Clearly $\bfv=D^{1/2}\bfone$ is an eigenvector with eigenvalue $\lambda=1$:
$$D^{-1/2}KD^{-1/2}(D^{1/2}\bfone) = D^{-1/2}K\bfone = D^{1/2}\bfone.$$
In fact $\lambda=1$ is the largest eigenvalue (left as an exercise) thus $\bfv_1 = D^{1/2}\bfone> 0$. Since $K'$ is symmetric the $\bfv_2^\top\bfv_1=0$ thus $\bfv_2$ contains positive and negative entries --- those are interpreted as indicating class membership (positive to one class and negative to the other).

The case $k> 2$ is treated as an {\it embedding\/} (also known as Multi-Dimensional Scaling) by re-coordinating the points $\bfx_i$ using the rows of $G$. In other words, the i'th row of $G$ is a representation of $\bfx_i$ in $R^k$. Under {\it ideal\/} conditions where $K$ is block diagonal (the distance between clusters is infinity) the rows associated with points clustered together are identical (i.e., the $n$ original points are mapped to $k$ points in $R^k$) \cite{Weiss-et-al-nips01}. In practice, one performs the iterative K-means in the embedded space.

\chapter{The Formal (PAC) Learning Model}
\label{chap:10}

We have see so far algorithms that explicitly estimate the underlying distribution of the data (Bayesian methods and EM) and algorithms that are in some sense optimal when the underlying distribution is Gaussian (PCA, LDA). We have also encountered an algorithm (SVM) that made no assumptions on the underlying distribution and instead tied the accuracy to the margin of the training data.

In this lecture and in the remainder of the course we will address the issue of "accuracy" and "generalization" in a more formal manner. Because the learner receives only a finite training sample, the learning function can do very well on the training set yet perform badly on new input instances. What we would like to establish are certain guarantees on the accuracy of the learner measured over all the instance space and not only on the training set. We will then use those guarantees to better understand what the large-margin principle of SVM is doing in the context of generalization.

In the remainder of this lecture we will refer to the following notations: the class of learning functions is denoted by $C$. A learning functions is often referred to as a "concept" or "hypothesis". A target function $c_t\in C$ is a function that has zero error on all input instances (such a function may not always exist).

\section{The Formal Model}

 In many learning situations of interest,  we would like to assume that the learner receives $m$ examples sampled by some fixed (yet unknown) distribution $D$ and the learner must do its best with the training set in order to achieve the accuracy and confidence objectives. The Probably Approximate Correct (PAC) model, also known as the "formal model", first introduced by Valient in 1984, provides a probabilistic setting which formalizes the notions of accuracy and confidence.

The PAC model makes the following statistical assumption. We assume the learner receives a set $S$ of $m$ instances $\bfx_1,...,\bfx_m\in X$ which are sampled randomly and independently according to a distribution $D$ over $X$. In other words, a random training set $S$ of length $m$ is distributed according to the product probability distribution $D^m$. The distribution $D$ is unknown, but we will see that one can obtain useful results by simply assuming that $D$ is fixed --- there is no need to attempt to recover $D$ during the learning process. To recap, 
we make the following three assumptions: (i) $D$ is unkown, (ii) $D$ is fixed throughout the learning process, and (iii) the example instances are sampled independently of each other (are Identically and Independently Distributed --- i.i.d.).

We distinguish between the "realizable" case where a target concept $c_t(\bfx)$ is known to exist, and the unrealizable case, where there is no such guarantee. In the realizable case our training examples are $Z=\{(\bfx_i,c_t(\bfx_i)\}$, $i=1,...,m$ and $D$ is defined over $X$ (since $y_i\in Y$ are given by $c_t(\bfx_i)$). In the unrealizable case,  $Z=\{(\bfx_i,y_i)\}$ and $D$ is the distribution over $X\times Y$ (each element is a pair, one from $X$ and the other from $Y$).

We next define what is meant by the {\it error\/} induced by a concept function $h(\bfx)$. In the realizable case, given a function $h\in C$, the error of $h$ is defined with respect to the distribution $D$:
$$err(h)=prob_D[\bfx: c_t(\bfx)\not= h(\bfx)] = \int_{\bfx\in X} ind(c_t(\bfx)\not= h(\bfx))D(\bfx)d\bfx$$
where $ind(F)$ is an indication function which returns '1' if the proposition $F$ is {\it true\/} and '0' otherwise. The function $err(h)$  is the probability that an instance $\bfx$ sampled according to $D$ will be labeled incorrectly by $h(\bfx)$. Let $\epsilon > 0$ be a parameter given to the learner specifying the "accuracy" of the learning process, i.e. we would like to achieve $err(h) \le \epsilon$. Note that $err(c_t)=0$.

In addition, we define a "confidence" parameter $\delta > 0$, also given to the learner,  which defines the probability that $err(h) > \epsilon$, namely,
$$prob[err(h) > \epsilon] < \delta,$$
or equivalently:
$$prob[err(h) \le \epsilon]\ge 1-\delta.$$
In other words, the learner is supposed to meet some accuracy criteria but is allowed to deviate from it by some small probability. Finally, the learning algorithm is supposed to be "efficient" if the running time is polynomial in $1/\epsilon,\ln(1/\delta),n$ and the size of the concept target function $c_t()$ (measured by the number of bits necessary for describing it, for example).

We will say that an algorithm $L$ learns a concept family $C$ in the formal sense (PAC learnable) if for any $c_t\in C$ and for every distribution $D$ on the instance space $X$, the algorithm $L$ generates efficiently a concept function $h\in C$ such that the probability that $err(h)\le \epsilon$ is at least $1-\delta$.

The inclusion of the confidence value $\delta$ could seem at first unnatural. What we desire from the learner is to demonstrate a consistent performance regardless of the training sample $Z$. In other words, it is not enough that the learner produces a hypothesis $h$ whose accuracy is above threshold, i.e., $err(h)\le \epsilon$, for some training sample $Z$. We would like the accuracy performance to hold under {\it all\/} training samples (sampled from the distribution $D^m$) --- since this requirement could be too difficult to satisfy, the formal model allows for some "failures", i.e, situations where $err(h)> \epsilon$, for some training samples $Z$, as long as those failures are rare and the frequency of their occurrence is controlled (the parameter $\delta$) and can be as small as we like.

In the unrealizable case, there may be no function $h\in C$ for which $err(h)=0$, thus we need to define what we mean by the {\it best\/} a learning algorithm can achieve:  
$$Opt(C) = \min_{h\in C} err(h),$$
which is the best that can be done on the concept class $C$ using functions that map between $X$ and $Y$. Given the desired accuracy $\epsilon$ and confidence $\delta$ values the learner seeks a hypothesis $h\in C$ such that:
$$prob[err(h) \le Opt(C) + \epsilon]\ge 1-\delta.$$
We are ready now to formalize the discussion above and introduce the definition of the formal learning model (Anthony \& Bartlett \cite{Bartlett99}, pp. 16):
\begin{definition}[Formal Model]
Let $C$ be the concept class of functions that map from a set $X$ to $Y$. A learning algorithm $L$ is a function:
$$L: \bigcup_{m=1}^\infty \{ (\bfx_i,y_i)\}_{i=1}^m \rightarrow C$$
from the set of all training examples to $C$ with the following property: given any $\epsilon,\delta\in (0,1)$ there is an integer $m_0(\epsilon,\delta)$ such that if $m\ge m_0$ then, for any probability distribution $D$ on $X\times Y$, if $Z$ is a training set of length $m$ drawn randomly according to the product probability distribution $D^m$, then with probability of at least $1-\delta$ the hypothesis $h=L(Z)\in C$ output by $L$ is such that $err(h) \le Opt(C) + \epsilon$.
We say that $C$ is learnable (or PAC learnable) if there is a learning algorithm for $C$.
\end{definition}

There are few points to emphasize. The sample size $m_0(\epsilon,\delta)$ is a sufficient sample size for PAC learning $C$ by $L$ and is allowed to vary with $\epsilon,\delta$. Decreasing the value of either $\epsilon$ or $\delta$ makes the learning problem more difficult and in turn a larger sample size is required. 
Note however that $m_0(\epsilon,\delta)$ does {\it not depend\/} on the distribution $D$! that is, a sufficient sample size can be given that will work for {\it any\/} distribution $D$ --- provided that $D$ is fixed throughout the learning experience (both training and later for testing). This point is a crucial property of the formal model because if the sufficient sample size is allowed to vary with the distribution $D$ then not only we would need to have some information about the distribution in order to set the sample complexity bounds, but also an adversary (supplying the training set) could control the rate of convergence of $L$ to a solution (even if that solution can be proven to be optimal)  and make it arbitrarily slow by suitable choice of $D$. 

What makes the formal model work in a distribution-invariant manner is that it critically depends on the fact that in many interesting learning scenarios the concept class $C$ is not too complex. For example, we will show later in the lecture that any finite concept class $|C|<\infty$ is learnable, and the sample complexity (in the realizable case) is
$$m\ge \frac{1}{\epsilon}\ln \frac{|C|}{\delta}.$$ 
In the next lecture we will consider concept classes of infinite size and show that despite the fact that the class is infinite it still can be of low complexity!

Before we illustrate the concepts above with an example, there is another  
useful measure which is the {\it empirical\/} error (also known as the sample error) $\hat{err}(h)$ which is defined as the proportion of examples from $Z$ on which $h$ made a mistake:
$$\hat{err}(h) = \frac{1}{m}| \{ i: h(\bfx_i) \not= c_t(\bfx_i)\} |$$
(replace $c_t(\bfx_i)$ with $y_i$ for the unrealizable case). The situation of bounding the true error $err(h)$ by minimizing the sample error $\hat{err}(h)$ is very convenient --- we will get to that later.

\section{The Rectangle Learning Problem}

As an illustration of learnability we will consider the problem (introduced in Kearns \& Vazirani \cite{Kearns97}) of learning an axes-aligned rectangle from positive and negative examples. We will show that the problem is PAC-learnable and find out 
$m_0(\epsilon,\delta)$.

In the rectangle learning game we are given a training set consisting of points in the 2D plane with a positive '+' or negative '-' label. The positive examples are sampled inside the target rectangle (parallel to the main axes) $R$ and the negative examples are sampled outside of $R$. Given $m$ examples sampled i.i.d according to some distribution $D$ the learner is supposed to generate an approximate rectangle $R'$ which is consistent with the training set (we are assuming that R exists) and which satisfies the accuracy and confidence constraints. 

We first need to decide on a learning strategy.   Since the solution $R'$ is not uniquely defined given any training set $Z$, we need to add further constraints to guarantee a unique solution.  We will choose $R'$ as the axes-aligned concept which gives the {\it tightest fit\/} to the positive examples, i.e., the smallest area axes-aligned rectangle which contains the positive examples. If no positive examples are given then $R'=\emptyset$. We can also assume that $Z$ contains at least three non-collinear positive examples in order to avoid complications associated with infinitesimal area rectangles. Note that we could have chosen other strategies, such as the middle ground between the tightest fit to the positive examples and the tightest fit (from below) to the negative examples, and so forth. Defining a strategy is necessary for the analysis below --- the type of strategy is not critical though.

We next define the error $err(R')$ on the concept $R'$ generated by our learning strategy. We first note that with the strategy defined above we always have $R'\subset R$ since $R'$ is the tightest fit solution which is consistent with the sample data (there could be a positive example outside of $R'$ which is not in the training set). We will define the "weight" $w(E)$ of a region $E$ in the plane as 
$$w(E)=\int_{\bfx\in E} D(\bfx)d\bfx,$$
i.e., the probability that a random point sampled according to the distribution $D$ will fall into the region. Therefore, the error associated with the concept $R'$ is 
$$err(R')=w(R-R')$$
and we wish to bound the error $w(R-R')\le \epsilon$ with probability of at least $1-\delta$ after seeing $m$ examples.

\begin{figure}[t]
\centerline{\epsfig{figure=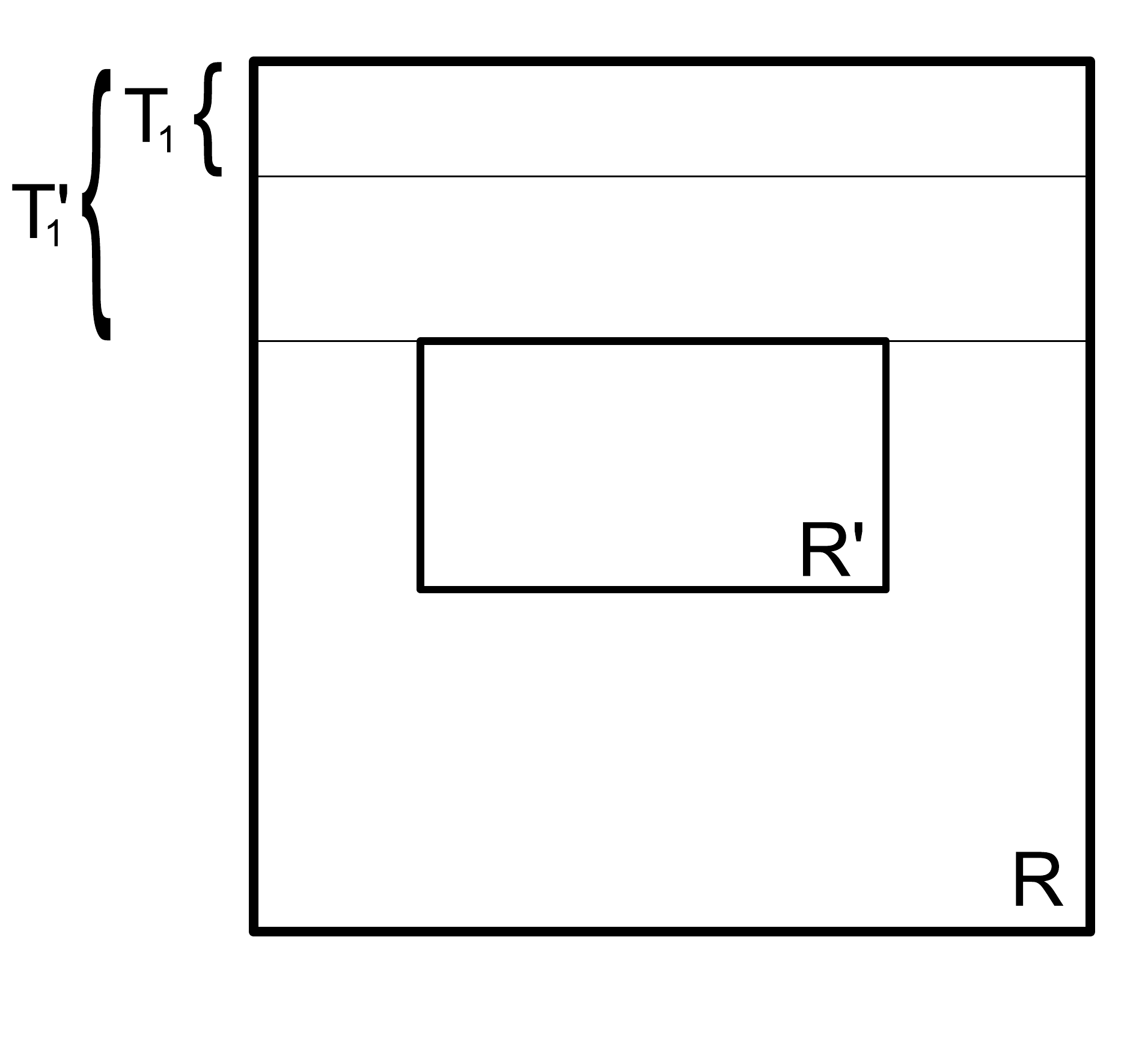,height=6cm}}
\caption{Given the tightest-fit to positive examples strategy we have that $R'\subset R$. The strip $T_1$ has weight $\epsilon/4$ and the strip $T'_1$ is defined as the upper strip covering the area between $R$ and $R'$.}
\label{fig:strips}
\end{figure}

We will divide the region $R-R'$ into four strips $T'_1,...,T'_4$ (see Fig.\ref{fig:strips}) which overlap at the corners. We will estimate $prob(w(T'_i)\ge \frac{\epsilon}{4})$ noting that the overlaps between the regions makes our estimates more pessimistic than they truly are (since we are counting the overlapping regions twice) thus making us lean towards the conservative side in our estimations. 

Consider the upper strip $T'_1$. If $w(T'_1\le \frac{\epsilon}{4})$ then we are done. We are however interested in quantifying the probability that this is not the case. Assume $w(T'_1)> \frac{\epsilon}{4}$ and define a strip $T_1$ which starts from the upper axis of $R$ and stretches to the extent such that $w(T_1)=\frac{\epsilon}{4}$. Clearly $T_1 \subset T'_1$. We have that $w(T'_1)> \frac{\epsilon}{4}$ iff $T_1 \subset T'_1$. Furthermore:
\begin{claim}
$T_1 \subset T'_1$ iff $\bfx_1,...,\bfx_m \not\in T_1$.
\end{claim}
{\bf Proof:\ }
If $\bfx_i\in T_1$ the the label must be positive since $T_1\subset R$. But if the label is positive then given our learning strategy of fitting the tightest rectangle over the positive examples, then $\bfx_i\in R'$. Since $T_1\not\subset R'$ it follows that $\bfx_i\not\in T_1$. \eop

We have therefore that $w(T'_1> \frac{\epsilon}{4})$ iff no point in $T_1$ appears in the sample $S=\{\bfx_1,...,\bfx_m\}$ (otherwise $T_1$ intersects with $R'$ and thus $T'_1 \subset T_1$). The probability that a point sampled according to the distribution $D$ will fall outside of $T_1$ is $1-\frac{\epsilon}{4}$. Given the independence assumption (examples are drawn i.i.d.), we have:
$$prob(\bfx_1,...,\bfx_m \not\in T_1)=prob(w(T'_1> \frac{\epsilon}{4}))=(1-\frac{\epsilon}{4})^m.$$ Repeating the same analysis to regions $T'_2,T'_3,T'_4$ and using the union bound $P(A\cup B)\le P(A) + P(B)$ we come to the conclusion that the probability that {\it any\/} of the four strips of $R-R'$ has weight greater that $\epsilon/4$ is at most $4(1-\frac{\epsilon}{4})^m$. In other words,
$$prob(err(L')\ge \epsilon) \le 4((1-\frac{\epsilon}{4})^m \le \delta.$$
We can make the expression more convenient for manipulation by using the inequality $e^{-x}\ge 1-x$ (recall that $1+(1/n))^n < e$ from which it follows that $(1+z)^{1/z} < e$ and by taking the power of $rz$ where $r\ge 0$ we obtain $(1+z)^r < e^{rz}$ then set $r=1,z=-x$):
$$4(1-\frac{\epsilon}{4})^m \le 4e^{-\frac{\epsilon m}{4}} \le \delta,$$
from which we obtain the bound:
$$m\ge \frac{4}{\epsilon}\ln \frac{4}{\delta}.$$
To conclude, assuming that the learner adopts the tightest-fit to positive examples strategy and is given at least $m_0=\frac{4}{\epsilon}\ln \frac{4}{\delta}$ training examples in order to find the axes-aligned rectangle $R'$, we can assert that with probability $1-\delta$ the error associated with $R'$ (i.e., the probability that an $(m+1)$'th point will be classified incorrectly) is at most $\epsilon$.

We can see form the analysis above that indeed it applies to any distribution $D$ where the only assumption we had to make is the independence of the draw. Also, the sample size $m$ behaves well in the sense that if one desires a higher level of accuracy (smaller $\epsilon$) or a higher level of confidence (smaller $\delta$) then the sample size grows accordingly. The growth of $m$ is linear in $1/\epsilon$ and linear in $\ln (1/\delta)$.

\section{Learnability of Finite Concept Classes}

In the previous section we illustrated the concept of learnability with a particular simple example. We will now focus on applying the learnability model to a more general family of learning examples. We will consider the family of all learning problems over finite concept classes $|C|< \infty$. For example, the conjunction learning problem (over boolean formulas)  with $n$ literals contains only $3^n$ hypotheses because each variable can appear in the conjunction or not and if appears it could be negated or not. We have shown that $n$ is the lower bound on the number of mistakes on the worst case analysis any on-line algorithm can achieve. With the definitions we have above on the formal model of learnability we can perform accuracy and sample complexity analysis that will apply to any learning problem over finite concept classes. This was first introduced by Valiant in 1984.

In the realizable case over $|C|< \infty$, we will show that any algorithm $L$ which returns a hypothesis $h\in C$ which is consistent with the training set $Z$ is a {\it learning algorithm\/} for $C$. In other words, any finite concept class is learnable and the learning algorithms simply need to generate consistent hypotheses. The sample complexity $m_0$ associated with the choice of $\epsilon$ and $\delta$ can be shown as equal to: $\frac{1}{\epsilon}\ln \frac{|C|}{\delta}$.

In the unrealizable case, any algorithm $L$ that generates a hypothesis $h\in C$ that {\it minimizes\/} the empirical error (the error obtained on $Z$) is a learning algorithm for $C$. The sample complexity can be shown as equal to: $\frac{2}{\epsilon^2}\ln \frac{2|C|}{\delta}$. We will derive these two cases below.

\subsection{The Realizable Case}

Let $h\in C$ be some consistent hypothesis with the training set $Z$ (we know that such a hypothesis exists, in particular $h=c_t$ the target concept used for generating $Z$) and suppose that 
$$err(h) = prob[\bfx\sim D: h(\bfx)\not = c_t(\bfx)] > \epsilon.$$
Then, the probability (with respect to the product distribution $D^m$) that $h$ agrees with $c_t$ on a random sample of length $m$ is at most $(1-\epsilon)^m$. Using the inequality we saw before $e^{-x}\ge 1-x$ we have:
$$prob[err(h)>\epsilon \ \&\&\ h(\bfx_i)=c_t(\bfx_i),\ \ i=1,...,m]\le (1-\epsilon)^m < e^{-\epsilon m}.$$
We wish to bound the error {\it uniformly\/}, i.e., that $err(h)\le \epsilon$ for all concepts $h\in C$. This requires the evaluation of:
$$prob[\max_{h\in C}\{ err(h)>\epsilon\}\  \&\&\ h(\bfx_i)=c_t(\bfx_i),\ \ i=1,...,m].$$

There at most $|C|$ such functions $h$, therefore using the Union-Bound the probability that {\it some\/} function in $C$ has error larger than $\epsilon$ and is consistent with $c_t$ on a random sample of length  $m$ is at most $|C|e^{-\epsilon m}$:
\begin{eqnarray*}
&&prob[\exists h: err(h)>\epsilon\  \&\&\ h(\bfx_i)=c_t(\bfx_i),\ \ i=1,...,m]\\
&&\le \sum_{h: err(h)>\epsilon} prob[h(\bfx_i)=c_t(\bfx_i),\ \ i=1,...,m]\\
&&\le |h: err(h)>\epsilon| e^{-\epsilon m}\\
&&\le |C|e^{-\epsilon m} 
\end{eqnarray*}
For any positive $\delta$, this probability is less than $\delta$ provided:  
$$m\ge \frac{1}{\epsilon}\ln \frac{|C|}{\delta}.$$
This derivation can be summarized in the following theorem (Anthony \& Bartlett \cite{Bartlett99}, pp. 25):
\begin{theorem}
Let $C$ be a finite set of functions from $X$ to $Y$. Let $L$ be an algorithm such that for any $m$ and for any $c_t\in C$, if $Z$ is a training sample $\{(\bfx_i,c_t(\bfx_i))\}$, $i=1,...,m$, then the hypothesis $h=L(Z)$ satisfies $h(\bfx_i)=c_t(\bfx_i)$. Then $L$ is a learning algorithm for $C$ in the realizable case with sample complexity
$$m_0 =  \frac{1}{\epsilon}\ln \frac{|C|}{\delta}.$$
\end{theorem}

\subsection{The Unrealizable Case}

In the realizable case an algorithm simply needs to generate a consistent hypothesize to be considered a learning algorithm in the formal sense. In the unrealizable situation (a target function $c_t$ might not exist) an algorithm which minimizes the empirical error, i.e., an algorithm $L$ generates $h=L(Z)$ having minimal sample error:
$$\hat{err}(L(Z)) = \min_{h\in C} \hat{err}(h)$$
is a learning algorithm for $C$ (assuming finite $|C|$). This is a particularly useful property given that the true errors of the functions in $C$ are unknown. It seems natural to use the sample errors $\hat{err}(h)$ as estimates to the performance of $L$. 

The fact that given a large enough sample (training set $Z$) then the sample error $\hat{err}(h)$ becomes close to the true error $err(h)$ is somewhat of a restatement of the "law of large numbers" of probability theory. For example, if we toss a coin many times then the relative frequency of 'heads' approaches the true probability of 'head' at a rate determined by the law of large numbers. We can bound the probability that the difference between the empirical error and the true error of some $h$ exceeds $\epsilon$ using Hoeffding's inequality:

\begin{claim}
Let $h$ be some function from $X$ to $Y=\{0,1\}$. Then
$$prob[| \hat{err}(h) - err(h)| \ge \epsilon]  \le 2e^{(-2\epsilon^2m)},$$
for any probability distribution $D$, any $\epsilon>0$ and any positive integer $m$.
\end{claim}
{\bf Proof:\ } This is a straightforward application of Hoeffding's inequality to Bernoulli variables. Hoeffding's inequality says: Let $X$ be a set, $D$ a probability distribution on $X$, and $f_1,...,f_m$ real-valued functions $f_i: X\rightarrow [a_i,b_i]$ from $X$ to an interval on the real line ($a_i < b_i$). Then,
\be
prob\left[| \frac{1}{m}\sum_{i=1}^m f_i(\bfx_i) - E_{\bfx\sim D}[f(\bfx)]| \ge \epsilon\right] \le 2e^{-\frac{2\epsilon^2m^2}{\sum_i (b_i-a_i)^2}}
\label{eq:H}
\ee
where 
$$E_{\bfx\sim D}[f(\bfx)] = \frac{1}{m}\sum_{i=1}^m \int f_i(\bfx)D(\bfx)d\bfx.$$
In our case $f_i(\bfx_i)=1$ iff $h(\bfx_i)\not= y_i$ and $a_i=0,b_i=1$. Therefore $(1/m)\sum_i f_i(\bfx_i)=\hat{err}(h)$ and $err(h)=E_{\bfx\sim D}[f(\bfx)]$. \eop

The Hoeffding bound almost does what we need, but not quite so. What we have is that for any {\it given\/} hypothesis $h\in C$, the empirical error is close to the true error with high probability. Recall that our goal is to minimize $err(h)$ over all possible $h\in C$ but we can access  only $\hat{err}(h)$. If we can guarantee that the two are close to each other {\it for every\/} $h\in C$, then minimizing $\hat{err}(h)$ over all $h\in C$ will approximately minimize $err(h)$. Put formally, in order to ensure that $L$ learns the class $C$, we must show that
$$prob\left[\max_{h\in C} | \hat{err}(h) - err(h)| < \epsilon\right ] > 1- \delta$$
In other words, we need to show that the empirical errors converge (at high probability) to the true errors {\it uniformly\/} over $C$ as $m\rightarrow \infty$. If that can be guaranteed, then with (high) probability $1-\delta$, for every $h\in C$,
$$err(h)-\epsilon < \hat{err}(h) < err(h) + \epsilon.$$
So, since the algorithm $L$ running on training set $Z$ returns $h=L(Z)$ which minimizes the empirical error, we have:
$$err(L(Z)) \le \hat{err}(L(Z)) + \epsilon = \min_h \hat{err}(h) + \epsilon \le Opt(C) + 2\epsilon,$$
which is what is needed in order that $L$ learns $C$. Thus, what is left is to prove the following claim:
\begin{claim}
$$prob\left[\max_{h\in C} | \hat{err}(h) - err(h)| \ge \epsilon\right ] \le 2|C|e^{-2\epsilon^2m}$$
\end{claim}
{\bf Proof:\ }
We will use the union bound. Finding the maximum over $C$ is equivalent to taking the union of all the events:
$$prob\left[\max_{h\in C} | \hat{err}(h) - err(h)| \ge \epsilon\right ] = prob\left[ \bigcup_{h\in C}  \{ Z: |\hat{err}(h) - err(h)| \ge \epsilon \}\right],$$
using the union-bound and Claim 2, we have:
$$\le \sum_{h\in C} prob\left[| \hat{err}(h) - err(h)| \ge \epsilon\right ] \le |C|2e^{(-2\epsilon^2m)}.$$
\eop

Finally, given that $2|C|e^{-2\epsilon^2m}\le \delta$ we obtain the sample complexity:
$$m_0 = \frac{2}{\epsilon^2}\ln \frac{2|C|}{\delta}.$$
This discussion is summarized with the following theorem (Anthony \& Bartlett \cite{Bartlett99}, pp. 21):
\begin{theorem}
Let $C$ be a finite set of functions from $X$ to $Y=\{0,1\}$. Let $L$ be an algorithm such that for any $m$ and for any training set $Z=\{(\bfx_i,y_i)\}$, $i=1,...,m$, then the hypothesis $L(Z)$ satisfies:
$$\hat{err}(L(Z)) = \min_{h\in C}\hat{err}(h).$$
Then $L$ is a learning algorithm for $C$ with sample complexity $m_0 = \frac{2}{\epsilon^2}\ln \frac{2|C|}{\delta}$.
\end{theorem}
Note that the main difference with the realizable case (Theorem 1) is the larger $1/\epsilon^2$ rather than $1/\epsilon$. The realizable case requires a smaller training set since we are estimating a random quantity so the smaller the variance the less data we need.

\chapter{The VC Dimension}
\label{chap:11}

The result of the PAC model (also known as the "formal" learning model) is that if the concept class $C$ is PAC-learnable then the learning strategy must simply consist of gathering a sufficiently large training sample $S$ of size $m>m_o(\epsilon,\delta)$, for given accuracy $\epsilon>0$ and confidence $0<\delta<1$ parameters, and finds a hypothesis $h\in C$ which is consistent with $S$. The learning algorithm is then guaranteed to have a bounded error $err(h)<\epsilon$ with probability $1-\delta$. The error measurement includes data {\it not seen\/} by the training phase.

This state of affair also holds (with some slight modifications on the sample complexity bounds) when there is no consistent hypothesis (the unrealizable case). In this case the learner simply needs to minimize the empirical error $\hat{err}(h)$ on the sample training data $S$, and if $m$ is sufficiently large then the learner is guaranteed to have $err(h) < Opt(C) + \epsilon$  with probability $1-\delta$. The measure $Opt(C)$ is defined as the minimal $err(g)$ over all $g\in C$. Note that in the realizable case $Opt(C)=0$. 

The property of bounding the true error $err(h)$ by minimizing the sample error $\hat{err}(h)$ is very convenient. The fundamental question is {\it under what conditions this type of generalization property applies?} We saw in the previous lecture that a satisfactorily answer can be provided when the cardinality of the concept space is bounded, i.e. $|C|<\infty$, which happens for Boolean concept space for example. In that lecture we have proven that:
$$m_o(\epsilon,\delta) = O(\frac{1}{\epsilon}\ln \frac{|C|}{\delta}),$$
is sufficient for guaranteeing a learning model in the formal sense, i.e., which has the generalization property described above.

In this lecture and the one that follows we have two goals in mind. First is to generalize the result of finite concept class cardinality to infinite cardinality --- note that the bound above is not meaningful when $|C|=\infty$. Can we learn in the formal sense any non-trivial infinite concept class? (we already saw an example of a PAC-learnable  infinite concept class which is the class of axes aligned rectangles). In order to answer this question we will need to a general measure of concept class complexity which will replace the cardinality term $|C|$ in the sample complexity bound $m_o(\epsilon,\delta)$. It is tempting to assume that the number of parameters which fully describe the concepts of $C$ can serve as such a measure, but we will show that in fact one needs a more powerful measure called the {\it Vapnik-Chervonenkis\/} (VC) dimension. Our second goal is to pave the way and provide the theoretical foundation for the {\it large margin principle\/} algorithm (SVM) we derived  in Lecture~\ref{chap:67}.


\section{The VC Dimension}

The basic principle behind the VC dimension measure is that although $C$ may have infinite cardinality, the restriction of the application of concepts in $C$ to a finite sample $S$ has a finite outcome. This outcome is typically governed by an exponential growth with the size $m$ of the sample $S$ --- but not always. The point at which the growth stops being exponential is when the "complexity" of the concept class $C$ has exhausted itself, in a manner of speaking.

We will assume $C$ is a concept class over the instance space $X$ --- both of which can be infinite. We also assume that the concept class maps instances in $X$ to $\{0,1\}$, i.e., the input instances are mapped to "positive" or "negative" labels. A training sample $S$ is drawn i.i.d according to some fixed but unknown distribution $D$ and $S$ consists of $m$ instances $\bfx_1,...,\bfx_m$. In our notations we will try to reserve $c\in C$ to denote the target concept and $h\in C$ to denote {\it some\/} concept. We begin with the following definition:

\begin{definition}
$$\Pi_C(S) = \{(h(\bfx_1),...,h(\bfx_m)\ :\ h\in C\}$$
which is a set of vectors in $\{0,1\}^m$.
\end{definition}
$\Pi_C(S)$ is set whose members are $m$-dimensional Boolean vectors induced by functions of $C$. These members are often called dichotomies or behaviors on $S$ induced or realized by $C$. If $C$ makes a full realization then $\Pi_C(S)$ will have $2^m$ members. An equivalent description is a collection of subsets of $S$:
$$\Pi_C(S) = \{ h \cap S\ :\ h\in C\}$$
where each $h\in C$ makes a partition of $S$ into two sets --- the positive and negative points. The set $\Pi_C(S)$ contains therefore subsets of $S$ (the positive points of $S$ under $h$). A full realization will provide $\sum_{i=0}^m {m\choose i}=2^m$. We will use both descriptions of $\Pi_C(S)$ as a collection of subsets of $S$ and as a set of vectors interchangeably.

\begin{definition}
If $|\Pi_C(S)|=2^m$ then $S$ is considered {\bf shattered} by $C$. In other words, $S$ is shattered by $C$ if $C$ realizes all possible dichotomies of $S$.
\end{definition}

Consider as an example a finite concept class $C=\{c_1,...,c_4\}$ applied to three instance vectors
with the results: 

\begin{center}
\begin{tabular}{c|ccc}
 & $\bfx_1$ & $\bfx_2$ & $\bfx_3$\\
\hline
$c_1$ & 1 & 1 & 1\\
$c_2$ & 0 & 1 & 1\\
$c_3$ & 1 & 0 & 0\\
$c_4$ & 0 & 0 & 0
\end{tabular}
\end{center}
Then, 

\begin{center}
\begin{tabular}{ll}
$\Pi_C(\{\bfx_1\})=\{(0),(1)\}$ & shattered\\
$\Pi_C(\{\bfx_1,\bfx_3\})=\{(0,0),(0,1),(1,0),(1,1)\}$ & shattered\\
$\Pi_C(\{\bfx_2,\bfx_3\})=\{(0,0),(1,1)\}$ & not shattered
\end{tabular}
\end{center}

\noindent With these definitions we are ready to describe the measure of concept class complexity. 

\begin{definition}[VC dimension]
The VC dimension of $C$, noted as $VCdim(C)$, is the cardinality $d$ of the largest set $S$ shattered by $C$. If all sets $S$ (arbitrarily large) can be shattered by $C$, then $VCdim(C)=\infty$.
$$VCdim(C)=\max\{d\ |\ \exists |S|=d,\ and\ |\Pi_C(S)|=2^d\}$$
\end{definition}

The VC dimension of a class of functions $C$ is the point $d$ at which {\it all\/} samples $S$ with cardinality  $|S|>d$ are {\it no longer shattered\/} by $C$. As long as $C$ shatters $S$ it manifests its full "richness" in the sense that one can obtain from $S$ all possible results (dichotomies). Once that ceases to hold, i.e., when $|S|>d$, it means that $C$ has "exhausted" its richness (complexity). An infinite VC dimension means that $C$ maintains full richness for all sample sizes. Therefore, the VC dimension is a combinatorial measure of a function class complexity.

Before we consider a number of examples of geometric concept classes and their VC dimension, it is important clarify the lower and upper bounds (existential and universal quantifiers) in the definition of VC dimension. The VC dimension is at least $d$ if there {\it exists\/} some sample $|S|=d$ which is shattered by $C$ --- this does not mean that all samples of size $d$ are shattered by $C$. Conversely, in order to show that the VC dimension is {\it at most\/} $d$, one must show that no sample of size $d+1$ is shattered. Naturally, proving an upper bound is more difficult than proving the lower bound on the VC dimension. The following examples are shown in a "hand waiving" style and are not meant to form rigorous proofs of the stated bounds --- they are shown for illustrative purposes only.

\vspace{0.5cm}

\noindent{\bf Intervals of the real line:}
The concept class $C$ is governed by two parameters $\alpha_1,\alpha_2$ in the closed interval $[0,1]$. A concept from this class will tag an input instance $0< x< 1$ as positive if $\alpha_1\le x \le \alpha_2$ and negative otherwise. The VC dimension is at least 2: select a sample of 2 points $x_1,x_2$ positioned in the open interval $(0,1)$. We need to show that there are values of $\alpha_1,\alpha_2$ which realize all the possible four dichotomies $(+,+),(-,-),(+,-),(-,+)$. This is clearly possible as one can place the interval $[\alpha_1,\alpha_2]$ such the intersection with the interval $[x_1,x_2]$ is null, (thus producing $(-,-)$), or to fully include $[x_1,x_2]$ (thus producing $(+,+)$) or to partially intersect $[x_1,x_2]$ such that $x_1$ or $x_2$ are excluded (thus producing the remaining two dichotomies). To show that the VC dimension is at most 2, we need to show that {\it any\/} sample of three points $x_1,x_2,x_3$ on the line $(0,1)$ cannot be shattered. It is sufficient to show that one of the dichotomies is not realizable: the labeling $(+,-,+)$ cannot be realizable by any interval $[\alpha_1,\alpha_2]$ --- this is because if $x_1,x_3$ are labeled positive then by definition the interval $[\alpha_1,\alpha_2]$ must fully include the interval $[x_1,x_3]$ and since $x_1< x_2<x_3$ then $x_2$ must be labeled positive as well. Thus $VCdim(C)=2$.

\vspace{0.5cm}

\noindent{\bf Axes-aligned rectangles in the plane:}
We have seen this concept class in the previous lecture --- a point in the plane is labeled positive if it lies in an axes-aligned rectangle. The concept class $C$ is thus governed by 4 parameters. The VC dimension is at least 4: consider a configuration of 4 input points arranged in a cross pattern (recall that we need only to show {\it some\/} sample $S$ that can be shattered). We can place the rectangles (concepts of the class $C$) such that all 16 dichotomies can be realized (for example, placing the rectangle to include the vertical pair of points and exclude the horizontal pair of points would induce the labeling $(+,-,+,-)$). It is important to note that in this case, not all configurations of 4 points can be shattered --- but to prove a lower bound it is sufficient to show the existence of a single shattered set of 4 points. To show that the VC dimension is at most 4, we need to prove that any set of 5 points cannot be shattered. For any set of 5 points there must be some point that is "internal", i.e., is neither the extreme left, right, top or bottom point of the five. If we label this internal point as negative and the remaining 4 points as positive then there is no axes-aligned rectangle (concept) which cold realize this labeling (because if the external 4 points are labeled positive then they must be fully within the concept rectangle, but then the internal point must also be included in the rectangle and thus labeled positive as well).
 
\vspace{0.5cm}

\noindent{\bf Separating hyperplanes:}
Consider first linear half spaces in the plane. The lower bound on the VC dimension is 3 since any three (non-collinear) points in $R^2$ can be shattered, i.e., all 8 possible labelings of the three points can be realized by placing a separating line appropriately. By having one of the points on one side of the line and the other two on the other side we can realize 3 dichotomies and by placing the line such that all three points are on the same side will realize the 4th. The remaining 4 dichotomies are realized by a sign flip of the four previous cases. To show that the upper bound is also 3, we need to show that no set of 4 points can be shattered. We consider two cases: (i) the four points form a convex region, i.e., lie on the convex hull defined by the 4 points, (ii) three of the 4 points define the convex hull and the 4th point is internal. In the first case, the labeling which is positive for one diagonal pair and negative to the other pair cannot be realized by a separating line. In the second case, a labeling which is positive for the three hull points and negative for the interior point cannot be realize. Thus, the VC dimension is 3 and in general the VC dimension for separating hyperplanes in $R^n$ is $n+1$.

\vspace{0.5cm}

\noindent{\bf Union of a finite number of intervals on the line:}
This is an example of a concept class with an infinite VC dimension. For any sample of points on the line, one can place a sufficient number of intervals to realize any labeling.

\vspace{0.5cm}

The examples so far were simple enough that one might get the wrong impression that there is a correlation between the number of parameters required to describe concepts of the class and the VC dimension. As a counter example, consider the two parameter concept class:
$$C=\{sign(\sin (\omega x+\theta): \omega\}$$
which has an infinite VC dimension as one can show that for every set of $m$ points on the line one can realize all possible labelings by choosing a sufficiently large value of $\omega$ (which serves as the frequency of the sync function) and appropriate phase.

We conclude this section with the following claim:
\begin{theorem}
The VC dimension of a finite concept class $|C|<\infty$ is bounded from above:
$$VCdim(C) \le \log_2|C|.$$
\end{theorem}
{\bf Proof:\ } if $VCdim(C)=d$ then there exists at least $2^d$ functions in $C$ because every function induces a labeling and there are at least $2^d$ labelings. Thus, from $|C|\ge 2^d$ follows that $d\le \log_2|C|$. \eop

\section{The Relation between VC dimension and PAC Learning}

We saw that the VC dimension is a combinatorial measure of concept class complexity and we would like to have it replace the cardinality term in the sample complexity bound. The first result of interest is to show that if the VC dimension of the concept class is infinite then the class is not PAC learnable. 

\begin{theorem}
Concept class $C$ with $VCdim(C)=\infty$ is not learnable in the formal sense.
\end{theorem}
{\bf Proof:\ }
Assume the contrary that $C$ is PAC learnable. Let $L$ be the learning algorithm and $m$ be the number of training examples required to learn the concept class with accuracy $\epsilon=0.1$ and $1-\delta=0.9$. That is, after seeing at least $m(\epsilon,\delta)$ training examples, the learner generates a concept $h$ which satisfies $p(err(h)\le 0.1)\ge 0.9$.

Since the VC dimension is infinite there exist a sample set $S$ with $2m$ instances which is shattered by $C$. Since the formal model (PAC) applies to any training sample we will use the set $S$ as follows. We will define a probability distribution on the instance space $X$ which is uniform on $S$ (with probability $\frac{1}{2m}$) and zero everywhere else.

Because $S$ is shattered, then any target concept is possible so we will choose our target concept $c$ in the following manner:
$$prob(c_t(\bfx_i)=0)=\frac{1}{2}\ \ \ \forall \bfx_i\in S,$$
in other words, the labels $c_t(\bfx_i)$ are determined by a coin flip.
The learner $L$ selects an i.i.d. sample of $m$ instances $\bar S$ --- which due to the structure of $D$ means that the $\bar S\subset S$ and outputs a consistent hypothesis $h\in C$. The probability of error for each $\bfx_i\not\in \bar S$ is:
$$prob(c_t(\bfx_i)\not = h(\bfx_i))=\frac{1}{2}.$$
The reason for that is because $S$ is shattered by $C$, i.e., we can select any target concept for any labeling of $S$ (the $2m$ examples) therefore we could select the labels of the $m$ points not seen by the learner arbitrarily (by flipping a coin). Regardless of $h$, the probability of mistake is $0.5$. The expectation on the error of $h$ is:
$$E[err(h)] = m\cdot 0\cdot\frac{1}{2m}+m\cdot\frac{1}{2}\cdot\frac{1}{2m}=\frac{1}{4}.$$
This is because we have $2m$ points to sample (according to $D$ as all other points have zero probability) from which the error on half of them is zero (as $h$ is consistent on the training set $\bar S$) and the error on the remaining half is $0.5$. Thus, the average error is $0.25$. Note that $E[err(h)]=0.25$ for any choice of $\epsilon,\delta$ as it is based on the sample size $m$. For any sample size $m$ we can follow the construction above and generate the learning problem such that if the learner produces a consistent hypothesis the expectation of the error will be $0.25$.

The result that $E[err(h)]= 0.25$ is not possible for the accuracy and confidence values we have set: with probability of at least $0.9$ we have that $err(h)\le 0.1$ and with probability $0.1$ then $err(h)=\beta$ where $0.1 < \beta \le 1$. Taking the worst case of $\beta=1$ we come up with the average error:
$$E[err(h)] \le 0.9\cdot 0.1 + 0.1\cdot 1 = 0.19 < 0.25.$$
We have therefore arrived to a contradiction that $C$ is PAC learnable. \eop

We next obtain a bound on the growth of $|\Pi_S(C)|$ when the sample size $|S|=m$ is much larger than the VC dimension $VCdim(C)=d$ of the concept class. We will need few more definitions:

\begin{definition}[Growth function]
$$\Pi_C(m) = \max\{|\Pi_S(C)|\ :\ |S|=m\}$$
\end{definition}
The measure $\Pi_C(m)$ is the maximum number of dichotomies induced by $C$ for samples of size $m$. As long as $m\le d$ then $\Pi_C(m)=2^m$. The question is what happens to the growth pattern of $\Pi_C(m)$ when $m>d$. We will see that the growth becomes polynomial --- a fact which is crucial for the learnability of $C$.

\begin{definition}
For any natural numbers $m,d$ we have the following definition:
\begin{eqnarray*}
\Phi_d(m)&=&\Phi_d(m-1) + \Phi_{d-1}(m-1)\\
\Phi_d(0)&=& \Phi_0(m)=1
\end{eqnarray*}
\end{definition}
By induction on $m,d$ it is possible to prove the following:
\begin{theorem}
$$\Phi_d(m) = \sum_{i=0}^d {m\choose i}$$
\end{theorem}
{\bf Proof:\ } by induction on $m,d$. For details see [\cite{Kearns97}, pp. 56].\eop

For $m\le d$ we have that $\Phi_d(m)=2^m$. For $m>d$ we can derive a polynomial upper bound as follows. 
$$\left(\frac{d}{m}\right)^d\sum_{i=0}^d {m\choose i} \le \sum_{i=0}^d \left(\frac{d}{m}\right)^i{m\choose i}\le \sum_{i=0}^m \left(\frac{d}{m}\right)^i{m\choose i}=(1+\frac{d}{m})^m\le e^d$$
From which we obtain:
$$\left(\frac{d}{m}\right)^d\Phi_d(m)\le e^d.$$
Dividing both sides by $\left(\frac{d}{m}\right)^d$ yields:
$$\Phi_d(m)\le e^d\left(\frac{m}{d}\right)^d=\left(\frac{em}{d}\right)^d = O(m^d).$$
We need one more result before we are ready to present the main result of this lecture:
\begin{theorem}[Sauer's lemma]
If $VCdim(C)=d$, then for any $m$, $\Pi_C(m)\le \Phi_d(m)$.
\end{theorem}
{\bf Proof:\ } By induction on both $d,m$. For details see [\cite{Kearns97}, pp. 55--56].\eop

Taken together, we have now a fairly interesting characterization on how the combinatorial measure of complexity of the concept class $C$ scales up with the sample size $m$. When the VC dimension of $C$ is infinite the growth is exponential, i.e., $\Pi_C(m)=2^m$ for all values of $m$. On the other hand, when the concept class has a bounded VC dimension $VCdim(C)=d<\infty$ then the growth pattern undergoes a discontinuity from an exponential to a polynomial growth:
$$\Pi_C(m)=\left\{\begin{array}{cl}
2^m & m\le d\\
\le \left(\frac{em}{d}\right)^d & m>d
\end{array}\right\}
$$
As a direct result of this observation, when $m>>d$ is much larger than $d$ the entropy becomes much smaller than $m$. Recall than from an information theoretic perspective, the entropy of a random variable $Z$ with discrete values $z_1,...,z_n$ with probabilities $p_i$, $i=1,...,n$ is defined as:
$$H(Z) = \sum_{i=0}^n p_i\log_2\frac{1}{p_i},$$
where $I(p_i)=\log_2\frac{1}{p_i}$ is a measure of "information", i.e., is large when $p_i$ is small (meaning that there is much information in the occurrence of an unlikely event) and vanishes when the event is certain $p_i=1$. The entropy is therefore the expectation of information. Entropy is maximal for a uniform distribution $H(Z)=\log_2n$. The entropy in information theory context can be viewed as the number of bits required for coding $z_1,...,z_n$. In coding theory it can be shown that the entropy of a distribution provides the lower bound on the average length of any possible encoding of a uniquely decodable code fro which one symbol goes into one symbol. When the distribution is uniform we will need the maximal number of bits, i.e., one cannot compress the data. In the case of concept class $C$ with VC dimension $d$, we see that one when $m\le d$ all possible dichotomies are realized and thus one will need $m$ bits (as there are $2^m$ dichotomies) for representing all the outcomes of the sample. However, when $m>>d$ only a small fraction of the $2^m$ dichotomies can be realized, therefore the distribution of outcomes is highly non-uniform and thus one would need much less bits for coding the outcomes of the sample. The technical results which follow are therefore a formal way of expressing in a rigorous manner this simple truth --- {\it If it is possible to compress, then it is possible to learn}. The crucial point is
that learnability is a direct consequence of the "phase transition" (from exponential to polynomial) in the growth of the number of dichotomies realized by the concept class.

In the next lecture we will continue to prove the "double sampling" theorem which derives the sample size complexity as a function of the VC dimension.

\chapter{The Double-Sampling Theorem}
\label{chap:12}

In this lecture will use the measure of VC dimension, which is a combinatorial measure of concept class complexity, to bound the sample size complexity. 

\section{A Polynomial Bound on the Sample Size $m$ for PAC Learning}

In this section we will follow the material presented in Kearns \& Vazirani \cite{Kearns97} pp. 57--61 and prove the following:
\begin{theorem}[Double Sampling]
Let $C$ be any concept class of VC dimension $d$. Let $L$ be any algorithm that when given a set $S$ of $m$ labeled examples $\{\bfx_i,c(\bfx_i)\}_i$, sampled i.i.d according to some fixed but unknown distribution $D$ over the instance space $X$, of some concept $c\in C$, produces as output a concept $h\in C$ that is consistent with $S$. Then $L$ is a learning algorithm in the formal sense provided that the sample size obeys:
$$m\ge c_0\left(\frac{1}{\epsilon}\log\frac{1}{\delta} + \frac{d}{\epsilon}\log\frac{1}{\epsilon}\right)$$
for some constant $c_0>0$.
\end{theorem}
The idea behind the proof is to build an "approximate" concept space which includes  concepts arranged such that the distance between the approximate concepts $h$ and the target concept $c$ is at least $\epsilon$ --- where distance is defined as the weight of the region in $X$ which is in conflict with the target concept. To formalize this story we will need few more definitions. Unless specified otherwise, $c\in C$ denotes the target concept and $h\in C$ denotes {\it some\/} concept.

\begin{definition}
$$c\Delta h = h\Delta c = \{\bfx\ :\ c(\bfx)\not=h(\bfx)\}$$
\end{definition}
$c\Delta h$ is the region in instance space where both concepts do not agree --- the error region. The probability that $\bfx\in c\Delta h$ is equal to (by definition) $err(h)$.
\begin{definition}
\begin{eqnarray*}
\Delta(c)&=&\{h\Delta c\ :\ h\in C\}\\
\Delta_\epsilon(c)&=&\{h\Delta c\ :\ h\in C\ and\ err(h)\ge\epsilon\}
\end{eqnarray*}
\end{definition}
$\Delta(c)$ is a set of error regions, one per concept $h\in C$ over all concepts. The error regions are with respect to the target concept. The set $\Delta_\epsilon(c)\subset \Delta(c)$ is the set of all error regions whose weight exceeds $\epsilon$. Recall that weight is defined as the probability that a point sampled according to $D$ will hit the region.

It will be important for later to evaluate the VC dimension of $\Delta(c)$. Unlike $C$, we are not looking for the VC dimension of a class of function but the VC dimension of a set of regions in space. Recall the definition of $\Pi_C(S)$ from the previous lecture: there were two equivalent definitions one based on a set of vectors each representing a labeling of the instances of $S$ induced by some concept. The second, yet equivalent, definition is based on a set of subsets of $S$ each induced by some concept (where the concept divides the sample points of $S$ into positive and negative labeled points). So far it was convenient to work with the first definition, but for evaluating the VC dimension of $\Delta(c)$ it will be useful to consider the second definition:
$$\Pi_{\Delta(c)}(S) = \{r \cap S\ :\ r\in \Delta(c)\},$$
that is, the collection of subsets of $S$ induced by intersections with regions of $\Delta(c)$. An intersection between $S$ and a region $r$ is defined as the subset of points from $S$ that fall into $r$. We can easily show that the VC dimensions of $C$ and $\Delta(c)$ are equal:
\begin{lemma}
$$VCdim(C)=VCdim(\Delta(c)).$$
\end{lemma}
{\bf Proof:\ } we have that the elements of $\Pi_C(S)$ and $\Pi_{\Delta(c)}(S)$ are susbsets of $S$, thus we need to show that for every $S$ the cardinality of both sets is equal $|\Pi_C(S)|=|\Pi_{\Delta(c)}(S)|$. To do that it is sufficient to show that for every element $s\in \Pi_C(S)$ there is a unique corresponding element in $\Pi_{\Delta(c)}(S)$. Let $c\cap S$ be the subset of $S$ induced by the target concept $c$. The set $s$ (a subset of $S$) is realized by some concept $h$ (those points in $S$ which were labeled positive by $h$). Therefore, the set $s\cap (c\cap S)$ is the subset of $S$ containing the points that hit the region $h\Delta c$ which is an element of $\Pi_{\Delta(c)}(S)$. Since this is a one-to-one mapping we have that $|\Pi_C(S)|=|\Pi_{\Delta(c)}(S)|$. \eop

\begin{definition}[$\epsilon$-net]
For every $\epsilon>0$, a sample set $S$ is an $\epsilon$-net for $\Delta(c)$ if every region in $\Delta_\epsilon(c)$ is hit by at least one point of $S$: 
$$\forall r\in \Delta_\epsilon(c),\ \ S\cap r \not=\emptyset.$$
\end{definition}
In other words, if $S$ hits all the error regions in $\Delta(c)$ whose weight exceeds $\epsilon$, then $S$ is an $\epsilon$-net. Consider as an example the concept class of intervals on the line $[0,1]$. A concept is defined by an interval $[\alpha_1,\alpha_2]$ such that all points inside the interval are positive and all those outside are negative. Given $c\in C$ is the target concept and $h\in C$ is some concept, then the error region $h\Delta c$ is the union of two intervals: $I_1$ consists of all points $x\in h$ which are not in $c$, and $I_2$ the interval of all points $x\in c$ but which are not in $h$. Assume that the distribution $D$ is uniform (just for the sake of this example) then, $prob(x\in I)=|I|$ which is the length of the interval $I$. As a result,
$err(h) > \epsilon$ if either $|I_1|> \epsilon/2$ or  $|I_2|> \epsilon/2$. The sample set 
$$S=\{x=\frac{k\epsilon}{2}\ :\ k=0,1,...,2/\epsilon\}$$
contains sample points from 0 to 1 with increments of $\epsilon/2$. Therefore, every interval larger than $\epsilon$ must be hit by at least one point from $S$ and by definition $S$ is an $\epsilon$-net.

It is important to note that if $S$ forms an $\epsilon$-net then we are guaranteed that $err(h)\le \epsilon$.  Let $h\in C$ be the consistent hypothesis with $S$ (returned by the learning algorithm $L$). Becuase $h$ is consistent, $h\Delta c \in \Delta(c)$ has not been hit by $S$ (recall that $h\Delta c$ is the error region with respect to the target concept $c$, thus if $h$ is consistent then it agrees with $c$ over $S$ and therefore $S$ does not hit $h\Delta c$). Since $S$ forms an $\epsilon$-net for $\Delta(c)$ we must have $h\Delta c \not\in \Delta_\epsilon(c)$ (recall that by definition $S$ hits all error regions with weight larger than $\epsilon$). As a result, the error region $h\Delta c$ must have a weight smaller than $\epsilon$ which means that $err(h)\le \epsilon$.

The conclusion is that if we can {\it bound\/} the probability that a random sample $S$ {\it does not\/} form an $\epsilon$-net for $\Delta(c)$, then we have bounded the probability that a concept $h$ consistent with $S$ has $err(h)>\epsilon$. This is the goal of the proof of the double-sampling theorem which we are about to prove below:

\noindent{\bf Proof (following Kearns \& Vazirani \cite{Kearns97} pp. 59--61):\ }
Let $S_1$ be a random sample of size $m$ (sampled i.i.d. according to the unknown distribution $D$) and let $A$ be the event that $S_1$ {\it does not\/} form an $\epsilon$-net for $\Delta(c)$. From the preceding discussion our goal is to upper bound the probability for $A$ to occur, i.e., $prob(A)\le \delta$. 

If $A$ occurs, i.e., $S_1$ is not an $\epsilon$-net, then by definition there must be some region $r\in\Delta_\epsilon(c)$ which is not hit by $S_1$, that is $S_1 \cap r=\emptyset$. Note that $r=h\Delta(c)$ for some concept $h$ which is consistent with $S_1$. At this point the space of possibilities is infinite, because the probability that we fail to hit $h\Delta(c)$ in $m$ random examples is at most $(1-\epsilon)^m$. Thus the probability that we fail to hit {\it some\/} $h\Delta c \in \Delta_\epsilon(c)$ is bounded from above by $|\Delta(c)|(1-\epsilon)^m$ --- which does not help us due to the fact that $|\Delta(c)|$ is infinite. 
The idea of the proof is to turn this into a finite space by using another sample, as follows. 

Let $S_2$ be another random sample of size $m$. We will select $m$ (for both $S_1$ and $S_2$) to guarantee a high probability that $S_2$ will hit $r$ many times. In fact we wish that $S_2$ will hit $r$ at least $\frac{\epsilon m}{2}$ with probability of at least $0.5$:
$$prob(|S_2\cap r| > \frac{\epsilon m}{2}) = 1 - prob(|S_2\cap r| \le \frac{\epsilon m}{2}).$$
We will use the Chernoff bound (lower tail) to obtain a bound on the right-hand side term. Recall that if we have $m$ Bernoulli trials (coin tosses) $Z_1,...,Z_m$ with expectation $E(Z_i)=p$ and we consider the random variable $Z=Z_1+...+Z_m$ with expectation $E(Z)=\mu$ (note that $\mu=pm$) then for all $0 < \psi<1$ we have:
$$prob(Z< (1-\psi)\mu)\le e^{-\frac{\mu\psi^2}{2}}.$$
Considering the sampling of $m$ examples that form $S_2$ as Bernoulli trials, we have that $\mu\ge\epsilon m$ (since the probability that an example will hit $r$ is at least $\epsilon$) and $\psi=0.5$. We obtain therefore:
$$prob(|S_2\cap r| \le (1-\frac{1}{2})\epsilon m)\le e^{-\frac{\epsilon m}{8}}=\frac{1}{2}$$
which happens when $m=\frac{8}{\epsilon}\ln 2=O(\frac{1}{\epsilon})$. To summarize what we have obtained so far, we have calculated the probability that $S_2$ will hit $r$ many times {\it given\/} that $r$ was fixed using the previous sampling, i.e., given that $S_1$ does not form an $\epsilon$-net. To formalize this, let $B$ denote the combined event that $S_1$ does not form an $\epsilon$-event and $S_2$ hits $r$ at least $\epsilon m/2$ times. Then, we have shown that for $m=O(1/\epsilon)$ we have:
$$prob(B/A) \ge \frac{1}{2}.$$
From this we can calculate $prob(B)$:
$$prob(B)=prob(B/A)prob(A) \ge \frac{1}{2}prob(A),$$
which means that our original goal of bounding $prob(A)$ is equivalent to finding a bound $prob(B)\le \delta/2$ because $prob(A) \le 2\cdot prob(B)\le \delta$. The crucial point with the new goal is that to analyze the probability of the event $B$, we need only to consider a finite number of possibilities, namely to consider the regions of 
$$\Pi_{\Delta_\epsilon(c)}(S_1\cup S_2)=\left\{ r\cap \{S_1\cup S_2\}\ :\ r\in\Delta_\epsilon(c)\right\}.$$
This is because the occurrence of the event $B$ is equivalent to saying that there is some $r\in \Pi_{\Delta_\epsilon(c)}(S_1\cup S_2)$ such that $|r|\ge \epsilon m/2$ (i.e., the region $r$ is hit at least $\epsilon m/2$ times) and $S_1\cap r=\emptyset$. This is because $\Pi_{\Delta_\epsilon(c)}(S_1\cup S_2)$ contains all the subsets of $S_1\cup S_2$ realized as intersections over all regions in $\Delta_\epsilon(c)$. Thus even though we have an infinite number of regions we still have a finite number of subsets. We wish therefore to analyze the following probability:
$$prob\left(r\in \Pi_{\Delta_\epsilon(c)}(S_1\cup S_2)\ :\ |r|\ge \epsilon m/2\ and\ S_1\cap r=\emptyset\right).$$
Let $S=S_1\cup S_2$ a random sample of $2m$ (note that since the sampling is i.i.d. it is equivalent to sampling $S_1$ and $S_2$ separately) and $r$ satisfying $|r|\ge \epsilon m/2$  being {\it fixed}. Consider some random partitioning of $S$ into $S_1$ and $S_2$ and consider then the problem of estimating the probability that $S_1\cap r=\emptyset$. This problem is equivalent to the following combinatorial question: we have $2m$ balls, each colored Red or Blue, with exaclty $l\ge \epsilon m/2$ Red balls. We divide the $2m$ balls into groups of equal size $S_1$ and $S_2$ and we are interested in bounding the probability that all of the $l$ balls fall in $S_2$ (that is, the probability that $S_1\cap r=\emptyset$). This in turn is equivalent to first dividing the $2m$ uncolored balls into $S_1$ and $S_2$ groups and then randomly choose $l$ of the balls to be colored Red and analyze the probability that all of the Red balls fall into $S_2$. This probability is exactly
$$\frac{{m\choose l}}{{2m\choose  l}}=\prod_{i=0}^{l-1}\frac{m-i}{2m-i} \le \prod_{i=0}^{l-1}\frac{1}{2}=\frac{1}{2^l}=2^{-\epsilon m/2}.$$
This probability was evaluated for a {\it fixed\/} $S$ and $r$. Thus, the probability that this occurs for {\it some\/} $r\in \Pi_{\Delta_\epsilon(c)}(S)$ satisfying $|r|\ge \epsilon m/2$ (which is $prob(B)$) can be calculated by summing over all possible fixed $r$ and applying the union bound $prob(\sum_i Z_i)\le \sum_iprob(Z_i)$:
\begin{eqnarray*}
prob(B)&\le& |  \Pi_{\Delta_\epsilon(c)}(S)|2^{-\epsilon m/2} \le  |  \Pi_{\Delta(c)}(S)|2^{-\epsilon m/2}\\
&=&  |  \Pi_C(S)|2^{-\epsilon m/2} \le \left(\frac{2\epsilon m}{d}\right)^d2^{-\epsilon m/2}\le \frac{\delta}{2},
\end{eqnarray*}
from which it follows that:
$$m = O\left(\frac{1}{\epsilon}\log\frac{1}{\delta} + \frac{d}{\epsilon}\log\frac{1}{\epsilon}\right).$$
\eop

\noindent Few comments are worthwhile at this point:
\begin{enumerate}
\item It is possible to show that the upper bound on the sample complexity $m$ is tight by showing that the lower bound on $m$ is $\Omega(d/\epsilon)$ (see [\cite{Kearns97}, pp. 62]).
\item The treatment above holds also for the unrealizable case (target concept $c\not\in C$) with slight modifications to the bound. In this context, the learning algorithm $L$ must simply minimize the sample (empirical) error $\hat{err}(h)$ defined:
$$\hat{err}(h)=\frac{1}{m}|\{i: h(\bfx_i)\not= y_i\}|\ \ \ \bfx_i\in S.$$
The generalization of the double-sampling theorem (Derroye'82) states that the empirical errors converge uniformly to the true errors:
$$prob\left(\max_{h\in C}|\hat{err(h)} - err(h)|\ge \epsilon\right)\le 4e^{(4\epsilon + 4\epsilon^2)}\left(\frac{\epsilon m^2}{d}\right)^d2^{-m\epsilon^2 /2}\le \delta,$$
from which it follows that
$$m = O\left(\frac{1}{\epsilon^2}\log\frac{1}{\delta} + \frac{d}{\epsilon^2}\log\frac{1}{\epsilon}\right).$$
\end{enumerate}
Taken together, we have arrived to a fairly remarkable result. Despite the fact that the distribution $D$ from which the training sample $S$ is drawn from is {\it unknown\/} (but is known to be fixed), the learner simply needs to minimize the empirical error. If the sample size $m$ is large enough the learner is guaranteed to have minimized the true errors for some accuracy and confidence parameters which define the sample size complexity. Equivalently,
$$|Opt(C)-\hat{err}(h)|\longrightarrow_{m\rightarrow \infty} 0.$$
Not only is the convergence is independent of $D$ but also the rate of convergence is independent (namely, it does not matter where the optimal $h^*$ is located). The latter is very important because without it one could arbitrarily slow down the convergence rate by maliciously choosing $D$. The beauty of the results above is that $D$ does not have an effect at all --- one simply needs to choose the sample size to be large enough for the accuracy, confidence and VC dimension of the concept class to be learned over.

\section{Optimality of SVM Revisited}

In Lecture~\ref{chap:67} we discussed the large margin principle for finding an optimal separating hyperplane. It is natural to ask how does the PAC theory presented so far explains why a maximal margin hyperplane is optimal with regard to the formal sense of learning (i.e. to generalization from empirical errors to true errors)? We saw in the previous section that the sample complexity $m(\epsilon,\delta,d)$ depends also on the VC dimension of the concept class --- which is $n+1$ for hyperplanes in $R^n$. Thus, another natural question that may certainly arise is what is the gain 
in employing the "kernel trick"? For a fixed $m$, mapping the input instance space $X$ of dimension $n$ to some higher (exponentially higher) feature space might simply mean that we are compromising the accuracy and confidence of the learner (since the VC dimension is equal to the instance space dimension plus 1). 

Given a fixed sample size $m$, the best the learner can do is to minimize the empirical error {\it and at the same time to try to minimize the VC dimension $d$ of the concept class}. The smaller $d$ is, for a fixed $m$, the higher the accuracy and confidence of the learning algorithm. Likewise, the smaller $d$ is, for a fixed accuracy and confidence values, the smaller sample size is required.

There are two possible ways to decrease $d$. First is to decrease the dimension $n$ of the instance space $X$. This amounts to "feature selection", namely find a subset of coordinates that are the most "relevant" to the learning task r perform a dimensionality reduction via PCA, for example. A second approach is to maximize the margin. Let the margin associated with the separating hyperplane $h$ (i.e. consistent with the sample $S$) be $\gamma$. Let the input vectors $\bfx\in X$ have a bounded norm, $|\bfx|\le R$. It can be shown that the VC dimension of the concept class $C_\gamma$ of hyperplanes with margin $\gamma$ is:
$$C_\gamma = \min\left\{\frac{R^2}{\gamma^2},n\right\}+1.$$
Thus, if the margin is very small then the VC dimension remains $n+1$. As the margin gets larger, there comes a point where $R^2/\gamma^2 < n$ and as a result the VC dimension decreases. Moreover, mapping the instance space $X$ to some higher dimension feature space will not change the VC dimension as long as the margin remains the same. It is expected that the margin will not scale down or will not scale down as rapidly as the scaling up of dimension from image space to feature space. 

To conclude, maximizing the margin (while minimizing the empirical error) is advantageous as it decreases the VC dimension of the concept class and causes the accuracy and confidence values of the learner to be largely immune to dimension scaling up while employing the kernel trick.

\chapter{Appendix}
\appendix

\section{Variance, Covariance, etc.}
\label{app:cov}

Let $X,Y$ be two random variables and let $f(x,y)$ be some function on $x\in X, y\in Y$, and let $p(x,y)$ be the probability of the event $x$ and $y$ occurring together. The expectation $E[f(x,y)]$ is defined:
$$E[f(x,y)]=\sum_{x\in X}\sum_{y\in Y} f(x,y)p(x,y)$$. The mean, variance and covariance are defined:
\begin{eqnarray*}
\mu_x &=& E[X] = \sum_x\sum_y xp(x,y)\\
\mu_y &=& E[Y] = \sum_x\sum_y yp(x,y)\\
\sigma_x^2 &=& Var[X] = E[(x-\mu_x)^2]= \sum_x\sum_y (x-\mu_x)^2p(x,y)\\
\sigma_y^2 &=& Var[Y] = E[(y-\mu_y)^2]= \sum_x\sum_y (y-\mu_y)^2p(x,y)\\
\sigma_{xy} &=& Cov(XY) = E[(x-\mu_x)(y-\mu_y)]=\sum_x\sum_y (x-\mu_x)(y-\mu_y)p(x,y)
\end{eqnarray*}
In vector-matrix notation, let $\bfx$ represent the $n$ random variables of $X_1,...,X_n$, i.e., $\bfx=(x_1,...,x_n)^\top$ is an instance vector and $p(\bfx)$ is the probability of the instance occurrence. Then the mean is a vector $\mu$ and the covariance matrix $E$ are defined:
\begin{eqnarray*}
\mu &=& \sum_{\bfx\in\{X_1,...,X_n\}}\bfx p(\bfx)\\
E &=& \sum_{\bfx} (\bfx - \mu)(\bfx - \mu)^\top p(\bfx)
\end{eqnarray*}
Note that the covariance matrix $E$ is the linear superposition of rank-1 matrices $(\bfx - \mu)(\bfx - \mu)^\top$ with coefficients $p(\bfx)$. The diagonal of $E$ containes the variances of the variables $x_1,...,x_n$. For a uniform distribution and a sample data $S$ consisting of $m$ points, let $A=[\bfx_1-\mu,...,\bfx_m-\mu]$ be the matrix whose columns consist of the points centered around the mean: $\mu = \frac{1}{m}\sum_i \bfx_i$. The (sample) covariance matrix is $E=\frac{1}{m}AA^\top$.

\section{Derivatives of Matrix Operations: Scalar Functions of a Vector}

The two most important examples of a scalar function of a vector $\bfx$ are the linear form $\bfa^\top\bfx$ and the quadratic form $\bfx^\top A\bfx$ for some square matrix $A$. 
\begin{eqnarray*}
d(\bfa^\top \bfx) &=& \bfa^\top d\bfx\\
d(\bfx^\top A\bfx) &=& (d\bfx)^\top A\bfx + \bfx^\top A(d\bfx)\\
&=& \left((d\bfx)^\top A\bfx\right)^\top + \bfx^\top A(d\bfx)\\
&=& \bfx^\top (A + A^\top)d\bfx
\end{eqnarray*}
where the derivative $d(\bfx^\top A\bfx)$ using the rule of products $d(f\cdot g) = (df)\cdot g + f\cdot(dg)$ where $g=A\bfx$ and $f=\bfx^\top$ and noting that $d(A\bfx)=Ad\bfx$. Thus, $\frac{d}{d\bfx}(\bfa^\top\bfx)=\bfa^\top$ and $\frac{d}{d\bfx}(\bfx^\top A\bfx))=\bfx^\top (A + A^\top)$. If $A$ is symmetric then $\frac{d}{d\bfx}(\bfx^\top A\bfx))=(2A\bfx)^\top$.

\section{Primer on Constrained Optimization}
\label{app:A}

\subsection{Equality Constraints and Lagrange Multipliers}

Consider first the general optimization with {\it equality\/}
constraints which gives rise to the notion of {\it Lagrange
multipliers}. 

\begin{eqnarray}
\min_{\bfx}&& f(\bfx)\label{eq:one}\\
&&subject\ \ to\nonumber\\
&&\bfh(\bfx)=0\nonumber
\end{eqnarray}
where $f: R^n\rightarrow R$ and $\bfh:R^n\rightarrow R^k$ where $\bfh$
is a vector function $(h_1,...,h_k)$ each from $R^n$ to $R$. We want to
derive a necessary and sufficient constraint for a point $\bfx_o$ to
be a local minimum subject to the $k$ equality constraints
$\bfh(\bfx)=0$. Assume that $\bfx_o$ is a {\it regular\/} point,
meaning that the gradient vectors $\nabla h_j(\bfx)$ are linearly
independent. Note that $\nabla\bfh(\bfx_o)$ is a $k\times n$ matrix
and the null space of this matrix:
$$null(\nabla\bfh(\bfx_o))=\{\bfy:\nabla\bfh(\bfx_o)\bfy=0\}$$
defines the {\it tangent\/} plane at the point $\bfx_o$. We have the
following fundamental theorem:
$$\nabla f(\bfx_o)\  \bot\ null(\nabla\bfh(\bfx_o))$$
in other words, all vectors $\bfy$ spanning the tangent plane at the
point $\bfx_o$ are also perpendicular to the gradient of $f$ at
$\bfx_o$.

The sketch of the proof is as follows. Let $\bfx(t)$, $-a \le t < a$,
be a smooth curve 
on the surface $\bfh(\bfx)=0$, i.e., 
$\bfh(\bfx(t))=0$. Let
$\bfx_o=\bfx(0)$ and $\bfy=\frac{d}{dt}\bfx(0)$ the tangent to the
curve at $\bfx_o$. From the 
definition of tangency, the vector $\bfy$ lives in
$null(\nabla\bfh(\bfx_o))$, i.e., $\bfy\cdot \nabla h_j(\bfx(0))=0$,
$j=1,...,k$. Since $\bfx_o=\bfx(0)$ is a local extremum of $f(\bfx)$,
then
$$0=\frac{d}{dt}f(\bfx(t))\vert_{t=0} = \sum \frac{\partial
f}{\partial x_i}\frac{dx_i}{dt}\vert_{t=0}=\nabla
f(\bfx_o)\cdot\bfy.$$ 

As a corollary of this basic theorem, the gradient vector $\nabla
f(\bfx_o)\in span\{\nabla h_1(\bfx_o),...,\nabla h_k(\bfx_o)\}$, i.e.,
$$\nabla f(\bfx_o) + \sum_{i=1}^k \lambda_i\nabla h_i(\bfx_o)=0,$$ 
where the coefficients $\lambda_i$ are called {\it Lagrange
Multipliers\/} and the expression:
$$f(\bfx) + \sum_i \lambda_i h_i(\bfx)$$
is called the {\it
Lagrangian\/} of the optimization problem (\ref{eq:one}).

\subsection{Inequality Constraints and KKT conditions}

Consider next the general constrained optimization with inequality
constraints (called ``non-linear programming''):
\begin{eqnarray}
\min_{\bfx}&& f(\bfx)\label{eq:ineq}\\
&&subject\ \ to\nonumber\\
&&\bfh(\bfx)=0\nonumber\\
&&\bfg(\bfx)\le 0\nonumber
\end{eqnarray}
where $\bfg:R^n\rightarrow R^s$. We will assume that the optimal
solution $\bfx_o$ is a regular point which has the following meaning:
Let $J$ be the set of indices $j$ such that $g_j(\bfx_o)=0$, then
$\bfx_o$ is a regular point if the gradient vectors $\nabla
h_i(\bfx_o),\nabla g_j(\bfx_o)$, $i=1,...,k$ and $j\in J$ are linearly
independent. A basic result (we will not prove here) is the Karush-Kuhn-Tucker (KKT)
theorem:

{\it Let $\bfx_o$ be a local minimum of the problem and suppose
$\bfx_o$ is a regular point. Then, there exist
$\lambda_1,...,\lambda_k$ and $\mu_1\ge 0,...,\mu_s\ge 0$ such that:}

\begin{eqnarray}
\nabla f(\bfx_o) + \sum_{i=1}^k \lambda_i\nabla h_i(\bfx_o) +
\sum_{j=1}^s \mu_j\nabla g_j(\bfx_o) &=&0,\label{eq:lag}\\
\sum_{j=1}^s \mu_jg_j(\bfx_o) &=&0.
\end{eqnarray}
Note that the condition $\sum \mu_jg_j(\bfx_o)=0$ is equivalent to the
condition that $\mu_jg_j(\bfx_o)=0$ (since $\mu\ge 0$ and
$\bfg(\bfx_o)\le 0$ thus there sum cannot vanish unless each term
vanishes) which in turn implies: $\mu_j=0$ when $g_j(\bfx_o)< 0$. The expression
$$L(\bfx,\lambda,\mu)=f(\bfx) + \sum_{i=1}^k \lambda_ih_i(\bfx) +
\sum_{j=1}^s \mu_jg_j(\bfx)$$
is the {\it
Lagrangian\/} of the problem (\ref{eq:ineq}) and the associated
condition $\mu_jg_j(\bfx_o)=0$ is called the KKT condition. 

The remaining concepts we need are the 
``duality'' and the ``Lagrangian Dual'' problem. 

\subsection{The Langrangian Dual Problem}

\begin{figure}
\begin{center}
\psfig{figure=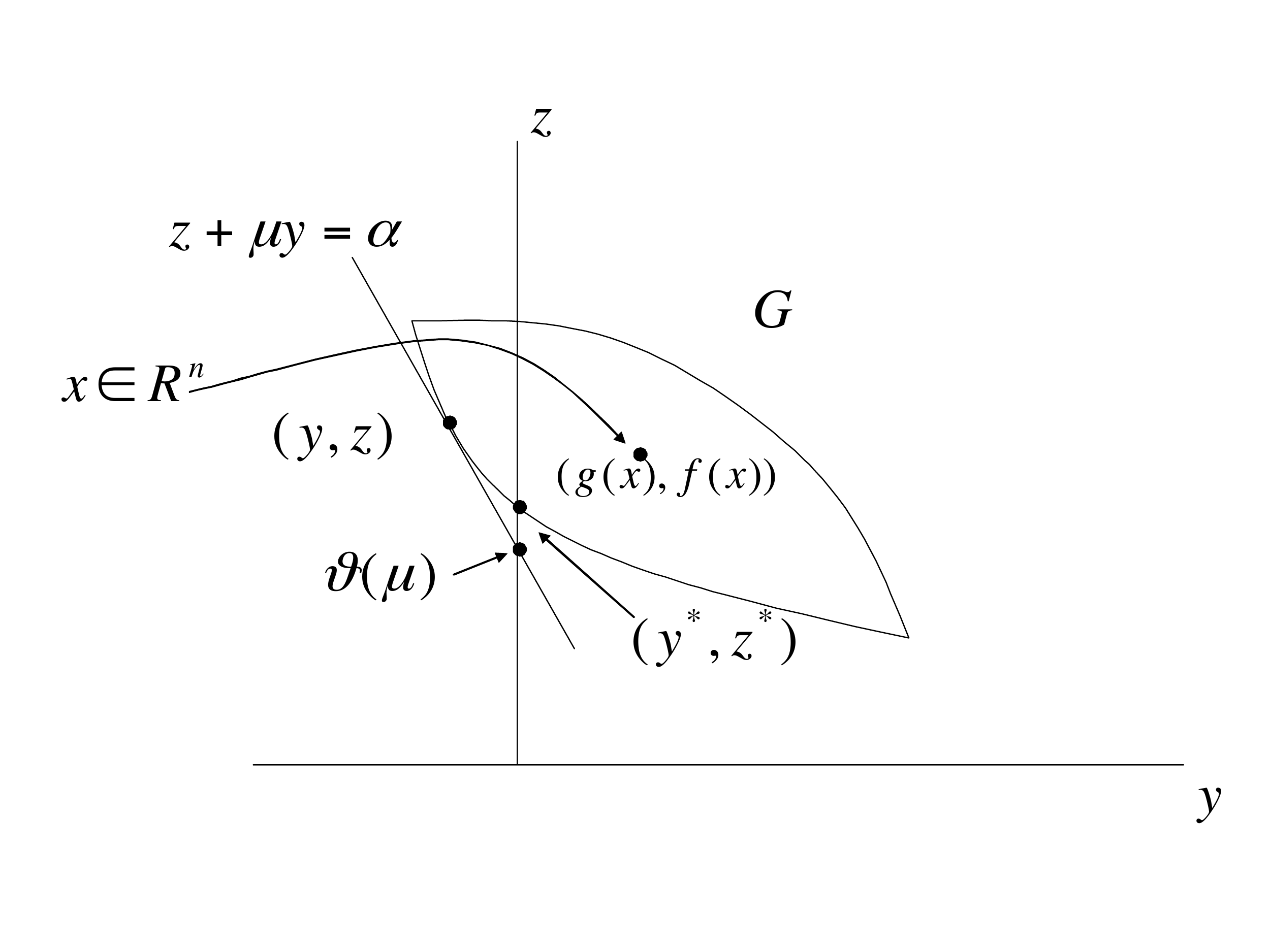,height={7cm}}
\end{center}
\caption{\protect\small Geometric interpreatation of Duality (see text). } 
\label{fig:geo1}
\end{figure} 

The optimization problem (\ref{eq:ineq}) is called the ``Primal''
problem. The Lagrangian Dual problem is defined as:
\begin{eqnarray}
\max_{\lambda,\mu}&& \theta(\lambda,\mu)\label{eq:dual}\\
&&subject\ \ to\nonumber\\
&&\mu\ge 0
\end{eqnarray}
where 
$$\theta(\lambda,\mu)=\min_{\bfx}\{f(\bfx)+\sum_i\lambda_ih_i(\bfx)+\sum_j\mu_jg_j(\bfx)\}.$$
Note that $\theta(\lambda,\mu)$ may assume the value $-\infty$ for some
values of $\lambda,\mu$ (thus to be rigorous we should have replaced
``min'' with ``inf''). The first basic result is the {\it weak
duality\/} theorem:

{\it Let $\bfx$ be a feasible solution to the primal (i.e.,
$\bfh(\bfx)=0,\bfg(\bfx)\le 0$) and let $(\lambda,\mu)$ be a feasible
solution to the dual problem (i.e., $\mu\ge 0$), then $f(\bfx)\ge
\theta(\lambda,\mu)$}

The proof is immediate:
\begin{eqnarray*}
\theta(\lambda,\mu)& =&
\min_{\bfy}\{f(\bfy)+\sum_i\lambda_ih_i(\bfy)+\sum_j\mu_jg_j(\bfy)\}\\
&\le& f(\bfx)+\sum_i\lambda_ih_i(\bfx)+\sum_j\mu_jg_j(\bfx)\\
&\le& f(\bfx)
\end{eqnarray*}
where the latter inequality follows from $\bfh(\bfx)=0$ and
$\sum_j\mu_jg_j(\bfx)\le 0$ because $\mu\ge0$ and $\bfg(\bfx)\le 0$.
As a corollary of this theorem we have:
\be
\min_{\bfx}\{f(\bfx):\bfh(\bfx)=0,\bfg(\bfx)\le 0\} \ge 
\max_{\lambda,\mu}\{\theta(\lambda,\mu):\mu\ge 0\}.\label{eq:weak}
\ee
The next basic result is the {\it strong duality\/} theorem which
specifies the conditions for when the inequality in (\ref{eq:weak})
becomes equality:

{\it Let $f(),\bfg()$ be convex functions and let $\bfh()$ be affine,
i.e., $\bfh(\bfx)=A\bfx - {\bf b}$ where $A$ is a $k\times n$ matrix,
then}
$$
\min_{\bfx}\{f(\bfx):\bfh(\bfx)=0,\bfg(\bfx)\le 0\} = 
\max_{\lambda,\mu}\{\theta(\lambda,\mu):\mu\ge 0\}.
$$
The strong duality theorem allows one to solve for the primal problem
by first dualizing it and solving for the dual problem instead (we
will see exactly how to do it when we return to solving the primal
problem (\ref{eq:svm-primal})). When the (convexity) conditions above
do not hold we obtain 
$$\min_{\bfx}\{f(\bfx):\bfh(\bfx)=0,\bfg(\bfx)\le 0\} > 
\max_{\lambda,\mu}\{\theta(\lambda,\mu):\mu\ge 0\}$$
which means that the optimal solution to the dual problem provides
only a lower bound to the primal problem --- this situation is called
a {\it duality gap}. Taken together, the "duality theorem" summarizes the discussion so far:
\begin{theorem}[Duality Theorem]
In order for $\bfx^*$ to be an optimal Primal solution and $(\lambda^*,\mu^*)$ to be an optimal Dual solution, it is necessary and sufficient that:
\begin{enumerate}
\item $\bfx^*$ is Primal feasible,
\item $\mu^* \ge 0$ and $\mu*_j=0$ for all $g_j(\bfx^*)<0$,
\item $\bfx^* \in \argminn_{\bfx} L(\bfx,\lambda^*,\mu^*)$.
\end{enumerate}
\end{theorem}
We will end this section with a geometric interpretation of
duality. 

\begin{figure}
\begin{center}
\psfig{figure=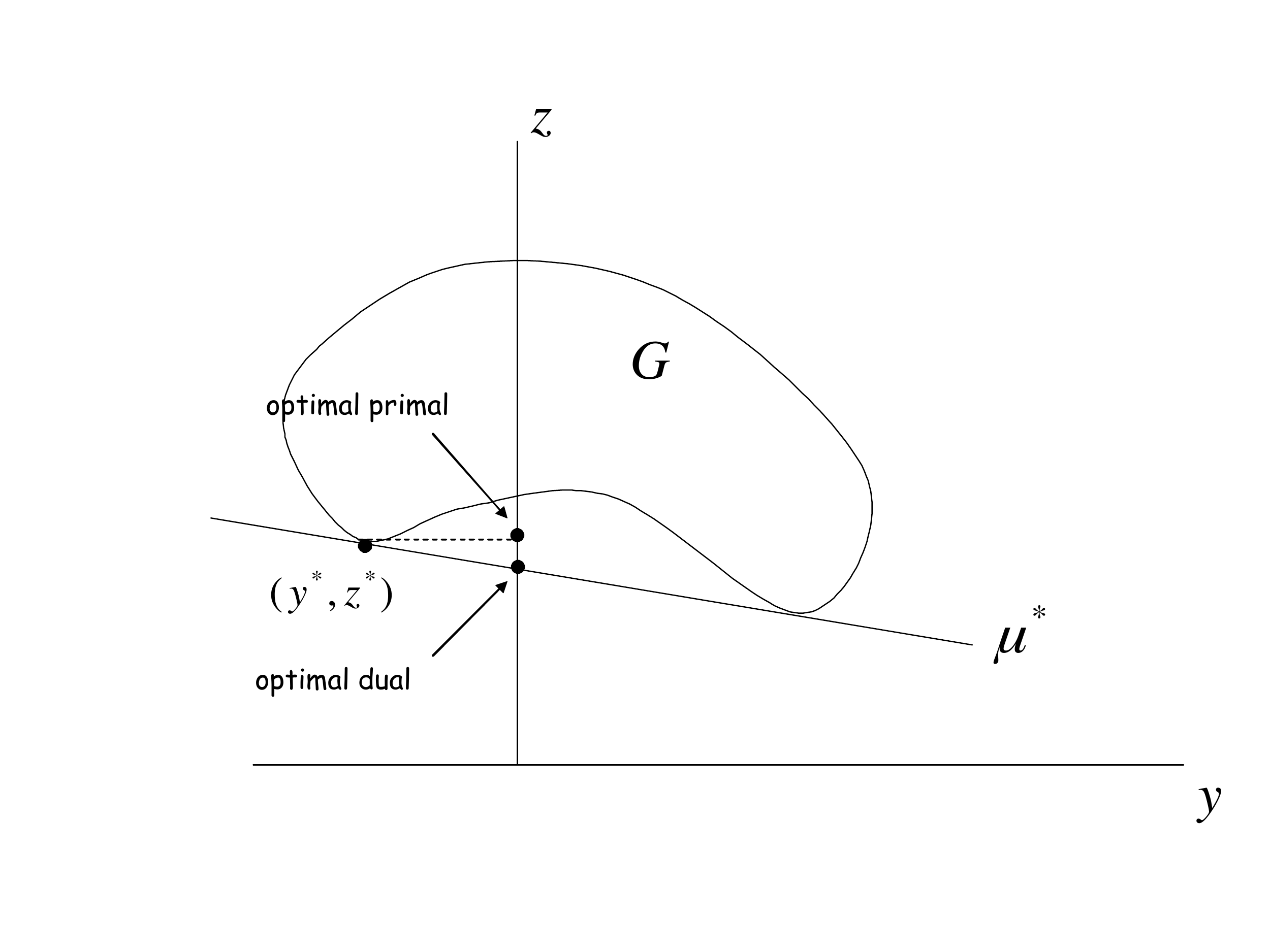,height={7cm}}
\end{center}
\caption{\protect\small An example of duality gap arising from
non-convexity (see text).} 
\label{fig:geo2}
\end{figure} 

\subsection{Geometric Interpretation of Duality}

For clarity we will consider a primal problem with a single
inequality constraint: $\min\{f(\bfx): g(\bfx)\le 0\}$ where
$g:R^n\rightarrow R$. 

Consider the set $G=\{(y,z):y=g(\bfx), z=f(\bfx)\}$ in the $(y,z)$
plane. The set $G$ is the image of $R^n$ under the $(g,f)$ map (see
Fig.~\ref{fig:geo1}). The primal problem is to find a point in $G$
that has a $y\le 0$ with the smallest $z$ value --- this is the point
$(y^*,z^*)$ in the figure.

In this case $\theta(\mu)=\min_{\bfx}\{f(\bfx)+\mu g(\bfx)\}$ which is
equivalent to minimize $z+\mu y$ over points in $G$. The equation
$z+\mu y=\alpha$ represents a straight line with slope $-\mu$ and
intercept (on $z$ axis) $\alpha$. For a given value $\mu$, to minimize
$z+\mu y$ over $G$ we need to move the line $z+\mu y=\alpha$ parallel
to itself as far down as possible while it remains in contact with
$G$ --- in other words $G$ is above the line and touches it. Then, the
intercept with the $z$ axis gives $\theta(\mu)$. {\it The dual problem
is therefore equivalent to finding the slope of the supporting
hyperplane such that its intercept on the $z$ axis is maximal}.

Consider the non-convex region $G$ in Fig.~\ref{fig:geo2} which
illustrates a duality gap condition. The optimal primal is the point
$(y^*,z^*)$ which is higher than the greatest intercept on the $z$
axis achieved by a line that supports $G$ from below. This is an
example of a duality gap caused by the non-convexity of the functions
$f(),g()$ (thereby making the set $G$ non-convex).

\bibliographystyle{plain}


\begin{thebibliography}{1}

\bibitem{Bartlett99}
M.~Anthony and P.L. Bartlett.
\newblock {\em Neural Neteowk Learning: Theoretical Foundations}.
\newblock Cambridge University Press, 1999.

\bibitem{Hall70}
K.M. Hall.
\newblock An r-dimensional quadratic placement algorithm.
\newblock {\em Manag. Sci.}, 17:219--229, 1970.

\bibitem{Kearns97}
M.J. Kearns and U.V. Vazirani.
\newblock {\em An Introduction to Computational Learning Theory}.
\newblock MIT Press, 1997.

\bibitem{Linde80}
Y.~Linde, A.~Buzo, and R.M. Gray.
\newblock An algorithm for vector quantizer design.
\newblock {\em IEEE Transactions on Communications}, 1:84--95, 1980.

\bibitem{Weiss-et-al-nips01}
A.Y. Ng, M.I. Jordan, and Y.~Weiss.
\newblock On spectral clustering: Analysis and an algorithm.
\newblock In {\em Proceedings of the conference on Neural Information
  Processing Systems (NIPS)}, 2001.

\bibitem{Malik-Shi-pami00}
J.~Shi and J.~Malik.
\newblock Normalized cuts and image segmentation.
\newblock {\em IEEE Transactions on Pattern Analysis and Machine Intelligence},
  22(8), 2000.

\bibitem{cp-iccv05}
R.~Zass and A.~Shashua.
\newblock A unifying approach to hard and probabilistic clustering.
\newblock In {\em Proceedings of the International Conference on Computer
  Vision}, Beijing, China, Oct. 2005.

\end{thebibliography}


\end{document}